\documentclass[11pt, a4paper, twocolumn]{article}
\usepackage[utf8]{inputenc}
\usepackage[pdftex]{graphicx}
\usepackage[lofdepth,lotdepth,labelformat=empty]{subfig}

\usepackage{pslatex}
\usepackage{colortbl}
\usepackage[margin=1in]{geometry}
\usepackage{setspace}
\usepackage{lscape}
\usepackage{multirow}
\usepackage{ctable}
\usepackage{amsmath}
\usepackage{natbib}
\usepackage{color}
\usepackage{eurosym}
\usepackage{booktabs}
\usepackage{dcolumn}
\newcolumntype{.}{D{.}{.}{-1}}
\usepackage[toc,page]{appendix}
\newcolumntype{C}[1]{>{\centering\let\\\tabularnewline}p{#1}}
\newcolumntype{R}[1]{>{\raggedleft\let\\\tabularnewline}p{#1}}
\newcolumntype{L}[1]{>{\raggedright\let\\\tabularnewline}p{#1}}

\usepackage{color}
\definecolor{darkred}{rgb}{0.5,0,0}
\definecolor{darkgreen}{rgb}{0,0.5,0}
\definecolor{darkblue}{rgb}{0,0,0.5}

\usepackage{hyperref}
\hypersetup{colorlinks,
 linkcolor=red,
 filecolor=darkgreen,
 urlcolor=darkblue,
 citecolor=darkblue,
 pdftitle={},
 pdfauthor={Stefano Gurciullo and Slava Mikhaylov},
}

\begin{document}

\title{Topology Analysis of International Networks Based on Debates in the United Nations\thanks{Authors' names are listed in alphabetical order. Authors have contributed equally to all work. UN General Debate Corpus is available on the Harvard Dataverse at \url{http://dx.doi.org/10.7910/DVN/0TJX8Y}.}}

\vspace{-0.3in}
\author{
Stefano Gurciullo\\
  University College London\\
  \href{mailto:stefano.gurciullo.11@ucl.ac.uk}{stefano.gurciullo.11@ucl.ac.uk}
\and
Slava Mikhaylov \\
 University of Essex\\
  \href{mailto:s.mikhaylov@essex.ac.uk}{s.mikhaylov@essex.ac.uk}\\  
 }

%\date{24 June 2015}

%%%%%%%%%%% TITLE PAGE %%%%%%%%%%%

\maketitle

\begin{abstract}
\footnotesize
\noindent
In complex, high dimensional and unstructured data it is often difficult to extract meaningful patterns. This is especially the case when dealing with textual data. Recent studies in machine learning, information theory and network science have developed several novel instruments to extract the semantics of unstructured data, and harness it to build a network of relations. Such approaches serve as an efficient tool for dimensionality reduction and pattern detection. This paper applies semantic network science to extract ideological proximity in the international arena, by focusing on the data from General Debates in the UN General Assembly on the topics of high salience to international community. UN General Debate corpus (UNGDC) covers all high-level debates in the UN General Assembly from 1970 to 2014, covering all UN member states. The research proceeds in three main steps. First, Latent Dirichlet Allocation (LDA) is used to extract the topics of the UN speeches, and therefore semantic information. Each country is then assigned a vector specifying the exposure to each of the topics identified. This intermediate output is then used in to construct a network of countries based on information theoretical metrics where the links capture similar vectorial patterns in the topic distributions. Topology of the networks is then analyzed through network properties like density, path length and clustering. Finally, we identify specific topological features of our networks using the map equation framework to detect communities in our networks of countries.

\vspace{1cm} \noindent \textbf{Key Words}: Topic modeling, network science, topology, information theory, map equation framework.

\vspace{1in}
\end{abstract}

\thispagestyle{empty}

%%%%%%%%%%% TEXT BODY %%%%%%%%%%%
\newpage

\section{Introduction}

A network or graph is a collection of entities, known as nodes, and a collection of relationships between nodes, known as edges or links \citep{wasserman1994social}. Analytically, a network can be represented as a set containing all nodes and edges, or by an adjacency square matrix with dimensions equal to the number of nodes. The matrix possesses nonzero values at the intersection of two nodes featuring a relationship. 

Networks concerning relationships across words and concepts are known as semantic networks \citep{fellbaum1998semantic}. Recently, advances in textual information processing have allowed the study of relationships among entities based on extracted semantics. For example, \citet{waumans2015topology} create social networks from the dialogues in the Harry Potter series. \citet{schultz2012strategic} provide interesting insights into public policy framing and crises by using the British Petroleum scandal as a case study. The approach has also been used to identify and analyze policy areas \citep[see e.g.][]{jung2015semantic}. At the same time, all such attempts have focused on the relationships between semantic concepts rather than on the actual actors who communicate them. Our approach aims at addressing this issue through a synergy between machine learning, information theory, and network science. 

The goal of this paper is to extract and identify communities of countries based on semantic information extracted from their speeches during the United Nations annual General Debate covering the period from 1970 to 2014 \citep{undebates}.\footnote{The UNGDC is publicly available on the Harvard Dataverse at \url{http://dx.doi.org/10.7910/DVN/0TJX8Y}} By doing so, we hope to set a novel methodological procedure that would help the scholarly community to better infer patterns and test hypotheses on the epistemic and ideological structures permeating political contexts. The study proceeds in three steps. First, Latent Dirichlet Allocation (LDA) is used to extract policy topics from the debates. LDA is an unsupervised machine learning method able to infer distributions of words co-occurring. Assuming that co-occurrent sets of words refer to similar conceptual spaces, we can view the resulting topics as having semantic value. Indeed, several applications of LDA show such conclusion to be valid \citep{dimaggio2013exploiting,wang2011collaborative}. For each year, we use LDA to extract eight topics with policy relevance, and construct a vector space featuring topic prevalences or probabilities for each country. Secondly, we harness information theory to build a metric of the similarity of the semantic information extracted from the countries' speeches. More specifically, we use the normalized mutual information coefficient to understand the extent to which a country's semantic space can be explained by the semantic space of another country. The mutual information scores are then used to build networks of countries, and their key structural and topological properties are studied over time. We observe significant shifts in such properties during the fall of the Soviet Union, suggesting that a phase transition has occurred across an epistemic and ideological context. In the final topological feature extraction step, we apply the map equation framework and its Infomap search algorithm \citep{rosvall2008maps} to identify communities from our semantic networks. As an initial and only partial validation of our results we show that the algorithm successfully identifies a separate community consisting of Soviet Bloc countries during the Cold War.

\section{Data}

The United Nations General Assembly annual regular session begins with the General Debate. During the General Debate (GD) the heads of state (or high ranking officials, such as ministers of foreign affairs) deliver formal speeches on most important issues of international and domestic politics from the perspective of their government.\footnote{For more information on the data see \citet{undebates}.} 

By tradition since 1947, the opening speech in GD is made by the representative of Brazil. The US representative is also typically scheduled to speak on the first day of the debate. Typically, the heads of states and governments are scheduled in the first days of GD, followed by vice-presidents, deputy prime ministers and foreign ministers, and concluding with the heads of delegation to the UN. While numbers vary session by session, on average heads of state or government comprise thirty-seven percent of speakers; vice-presidents, deputy prime ministers and foreign ministers are about fifty-six percent of speakers; and country representatives at the UN are about seven percent of all speakers.

Speeches are mostly made in native languages. However, according to the rules of the Assembly, all statements are then translated by the UN into the six official languages (Arabic, Chinese, English, French, Russian, and Spanish) and deposited at the United Nations Dag Hammarskjold Library. We collected all General Debate speeches using dedicated pages of individual UNGA General Debates and available in the UN Bibliographic Information System (UNBIS). Our text corpus contains all speeches in the same language (English). If a speech was delivered in a language other than English, we use the official English version of the speech provided by the UN. Overall, we collected 7,310 statements delivered in GD by heads of state or their representative for the period between 1970 and 2014. The number of countries participating in GD grows from 70 in 1970 to 193 in 2014 in line with the growth of UN membership. On average speeches contain 123 sentences and 945 unique words.

UN General Debate text corpus used in subsequent estimations was created from digitized UN library archives, which were converted into plain text format. Speeches made before 1992 are stored as image copies of typewritten documents, generally of very poor image quality, which requires additional preprocessing using optical character recognition software. We apply standard preprocessing of the data: stop word removal; turning all words to lower case; and removal of all numbers and non-Latin1 characters, and stemming. We also trimmed the corpus by removing all words that appear less than ten times in the corpus, and in less than five documents. This reduced the document-feature matrix from 7310 $\times$ 48968 to 7310 $\times$ 11733. We also experimented with various trimming settings ranging between 10 and 100 `words' and `documents' parameters with no noticeable effect on our outcomes. 

A typical country's speech during GD covers various issues of concern in international security, development, human rights, environment, amongst others, as well as issues of regional or national concern. In fact, such a speech --- often made by the heads of state and government  --- is not unlike the state-of-the-union legislative address that the heads of state deliver to their parliaments. In their address to the General Assembly during the GD, the heads of state or their representatives discuss the most important issues in international politics, review their national foreign policy, criticize or praise the work of the United Nations, and outline issues that need to be addressed by the international community. The most pressing domestic issues of relevance to international community are also raised during the speeches.  

The GD performs a number of important functions as scholars have pointed out \citep[see e.g.][]{luard1994,nicholas1959,smith2006,bailey1960}. The central function is that the statements made by member states ``act as a barometer of international opinion on important issues, even those not on the agenda for that particular session'' \citep[155]{smith2006}. This means that speeches in the GD provide governments with an opportunity to put forward their perspective and their position on a range of issues. Indeed, a central purpose of statements at the GD is for representatives to ensure that a point of view is on public record \citep{bailey1960}. Furthermore, statements made at the UN GD provide one of the few opportunities for smaller and less powerful states to make their positions on various issues known to the public \citep{nicholas1959,smith2006}. Therefore, speeches in the GD provide a key source of information on state preferences.

\section{Analysis}

\subsection{Networks}

Using the UN General Debate corpus we estimate a standard Latent Dirichlet Allocation (LDA) model \citep{blei2003latent} for each year separately. After some experimentation we specified eight topics parameter choice as that produced the most subjectively meaningful topics. From the LDA analysis we extract topic prevalences, which is a proportion of each country's speech devoted to a given topic. 

The vectors of topic prevalences that we created for each country can also be interpreted as probability distributions that describe the chances that we randomly pick a word from a country's speech that belongs to any of the topics. Given this perspective, we can assess the similarity of two probability distributions using the mutual information (MI) measure and create a network based on this metric.

Mutual information is a measure of mutual independence of two variables, where MI of two continuous random variables is defined as:

$$I(X;Y)= \int_{Y} \int_{X} p(x,y) log \left( \frac{p(x,y)}{p(x)p(y)} \right) dx dy,$$

where $p(x,y)$ is the joint probability density function of $X$ and $Y$, and $p(x)$ and $p(y)$ are the marginal probability density functions of $X$ and $Y$. 

Mutual information measures how much two random variable share the information. Mutual information is zero when two variables are independent and knowing one doesn't provide any information about the other. On the other hand, when knowing one variable determines the value of the other variable, then MI is the uncertainty contained in (or entropy of) each variable. 

\begin{figure}
\centering

\includegraphics[width=.5\textwidth]{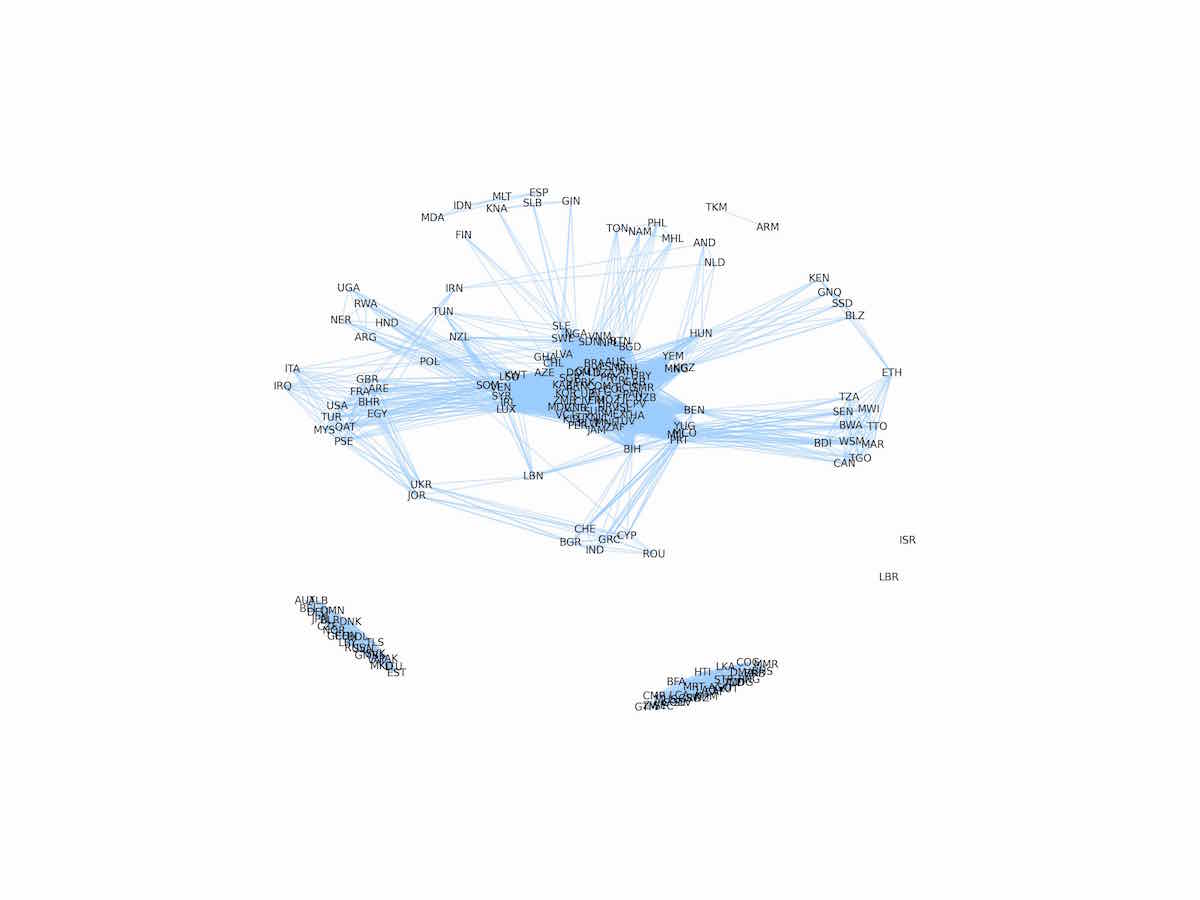}

\caption{\emph{Network built from topics covered in speeches by country leaders during the 2014 UNGA General Debate.} First, topics were identified using an LDA model, then utilizing the distribution of topic coverage across countries we built the network where the links between countries are identified using the mutual information measure. \label{fig:network2014}}
\end{figure}

Here we use \cite{strehl2003cluster}'s normalized variant of mutual information:

$$\frac{I(X;Y)}{\sqrt{H(X)H(Y)}},$$ 

where $H(X)$ and $H(Y)$ are marginal entropies. This variant of normalized mutual information ranges from 0 to 1. We build network based on important links that we define as having mutual information above a certain threshold. After some experimentation, we opted for having a threshold value of 0.8. Any link with a value lower than that has been cut. An example of the network based on the 2014 General Debate is presented in Figure \ref{fig:network2014}, with the full set of network visualizations presented in Figures \ref{fig:networks1}-\ref{fig:networks8} in supplementary materials.

%%%%%%%%%%%%%%%%%%%%%%%%%%%%%%%%%%%%%%%%
%FIGURE: 
%%%%%%%%%%%%%%%%%%%%%%%%%%%%%%%%%%%%%%%%

\subsection{Network properties}

In analyzing structural properties of our networks we focused on four quantities: density; average path length; global clustering coefficient; and diameter of the network. Density ($D$) is the actual number of links over the total possible number in the network: $$ D=\frac{2E}{N(N-1)},$$ where $E$ is the number of edges and denominator captures the number of possible edges. Average path length is calculated by finding the sum of shortest paths between all pairs of nodes divided by the total number of pairs. This captures the average number of steps it takes to travel from one member of the network to another. Global clustering coefficient is an average of clustering coefficients ($C$) for each node measured as the ratio of existing links connecting a node's neighbors to each other to the maximum possible number of such links. This is often viewed as a measure of ``small world'', i.e. how tightly connected are the nodes. The clustering coefficient of the $i$'th node is $$C_i=\frac{2e_i}{k_i(k_i-1)},$$ where $k_i$ is the number of neighbors of the $i$'s node, and $e_i$ is the number of connections between these neighbors, with the maximum number of connections between neighbors being $\binom{k}{2}=\frac{k(k-1)}{2}$. Diameter of a network is the longest of all the calculated shortest paths in a network and captures the linear size of a network. Table \ref{tab:properties} presents calculated structural properties of the networks. 

%%%%%%%%%%%%%%%%%%%%%%%%%%%%%%%%%%%%%%%
%TABLE:Network properties.
\begin{table}
\centering
\scriptsize
\begin{tabular}{l c c c c}
\toprule
Year	&	Density	&	Average path 	&	Global clustering 	&	Diameter	\\
	&		&	 length	&	coefficient	&		\\
\midrule
1970	&	0.18	&	2.09	&	0.76	&	6	\\
1971	&	0.14	&	2.37	&	0.60	&	6	\\
1972	&	0.17	&	2.06	&	0.79	&	6	\\
1973	&	0.11	&	2.83	&	0.55	&	7	\\
1974	&	0.20	&	2.09	&	0.67	&	6	\\
1975	&	0.19	&	2.05	&	0.76	&	6	\\
1976	&	0.17	&	2.38	&	0.70	&	7	\\
1977	&	0.11	&	2.68	&	0.63	&	7	\\
1978	&	0.21	&	2.02	&	0.75	&	6	\\
1979	&	0.18	&	2.39	&	0.68	&	7	\\
1980	&	0.25	&	1.97	&	0.78	&	6	\\
1981	&	0.21	&	2.20	&	0.63	&	6	\\
1982	&	0.10	&	2.72	&	0.58	&	6	\\
1983	&	0.18	&	2.20	&	0.82	&	6	\\
1984	&	0.25	&	1.90	&	0.77	&	6	\\
1985	&	0.23	&	1.83	&	0.94	&	5	\\
1986	&	0.17	&	2.18	&	0.72	&	6	\\
1987	&	0.17	&	2.29	&	0.66	&	6	\\
1988	&	0.28	&	1.37	&	0.97	&	5	\\
1989	&	0.15	&	2.45	&	0.67	&	8	\\
1990	&	0.16	&	2.23	&	0.74	&	6	\\
1991	&	0.15	&	2.26	&	0.73	&	6	\\
1992	&	0.15	&	2.29	&	0.63	&	6	\\
1993	&	0.20	&	1.84	&	0.84	&	5	\\
1994	&	0.13	&	2.46	&	0.70	&	6	\\
1995	&	0.15	&	2.21	&	0.70	&	6	\\
1996	&	0.22	&	1.87	&	0.78	&	6	\\
1997	&	0.16	&	2.30	&	0.73	&	8	\\
1998	&	0.15	&	2.25	&	0.90	&	6	\\
1999	&	0.16	&	2.14	&	0.80	&	6	\\
2000	&	0.10	&	2.70	&	0.56	&	6	\\
2001	&	0.12	&	2.46	&	0.79	&	8	\\
2002	&	0.13	&	2.50	&	0.73	&	7	\\
2003	&	0.15	&	2.12	&	0.84	&	6	\\
2004	&	0.09	&	2.68	&	0.57	&	7	\\
2005	&	0.13	&	2.41	&	0.77	&	6	\\
2006	&	0.19	&	1.93	&	0.87	&	5	\\
2007	&	0.14	&	2.38	&	0.59	&	6	\\
2008	&	0.16	&	2.40	&	0.87	&	6	\\
2009	&	0.21	&	1.99	&	0.81	&	6	\\
2010	&	0.13	&	2.22	&	0.89	&	6	\\
2011	&	0.14	&	2.55	&	0.65	&	7	\\
2012	&	0.10	&	2.65	&	0.77	&	8	\\
2013	&	0.12	&	2.56	&	0.60	&	6	\\
2014	&	0.21	&	1.96	&	0.85	&	6	\\
\bottomrule

\end{tabular}
\caption{Basic properties of the networks calculated for each year, 1970-2014. \label{tab:properties}}

\end{table}
%%%%%%%%%%%%%%%%%%%%%%%%%%%%%%%%%%%%%%%% 

\begin{figure}
\centering
\subfloat[Density]{\includegraphics[width=.45\textwidth]{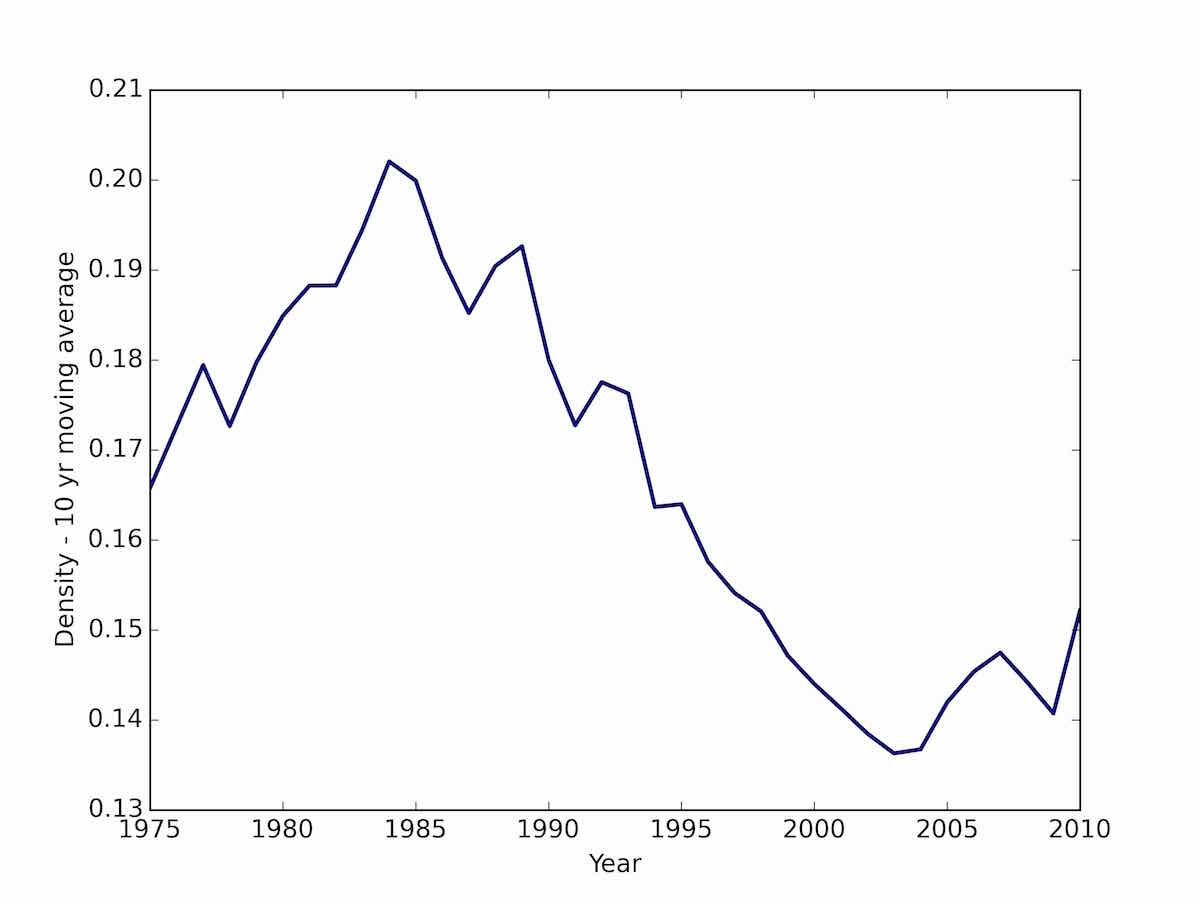}}\\
\subfloat[Average path length]{\includegraphics[width=.45\textwidth]{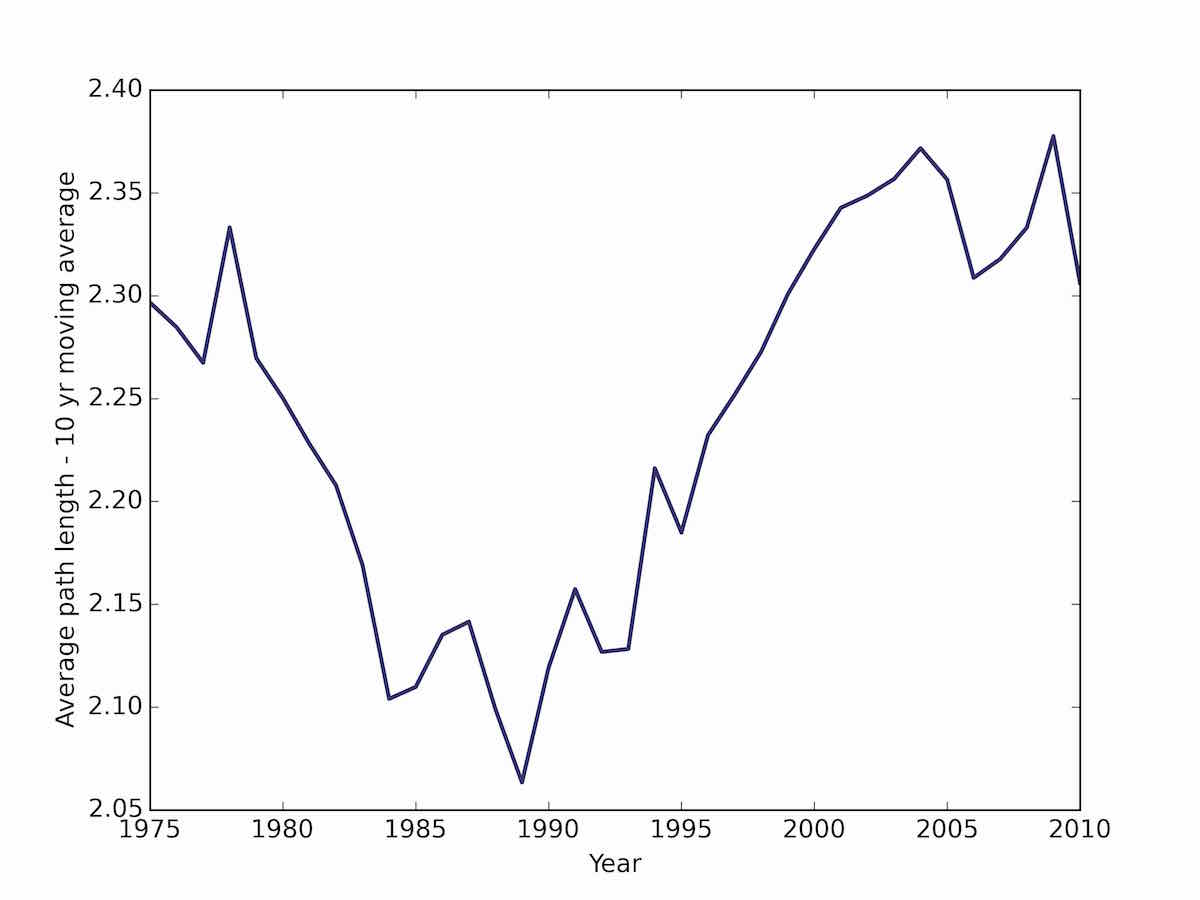}}\\
\subfloat[Global clustering coefficient]{\includegraphics[width=.45\textwidth]{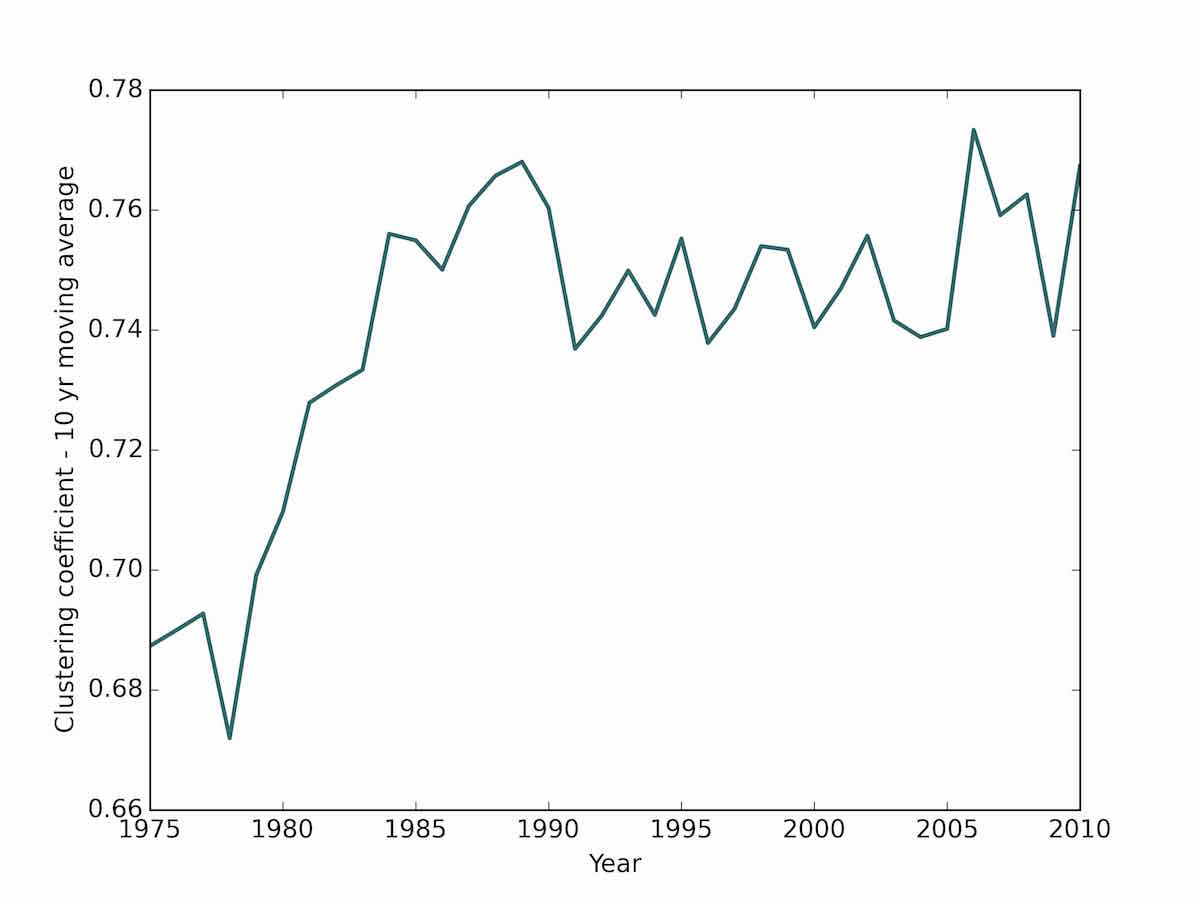}}

\caption{Network properties, 10-year moving averages. 
\label{fig:properties}}

\end{figure}

Annual fluctuations in network measures suggest some potential noise, so we created 10-year moving averages of each series. The results are plotted in Figure \ref{fig:properties}.

%%%%%%%%%%%%%%%%%%%%%%%%%%%%%%%%%%%%%%%%%
%FIGURE: Network properties
%%%%%%%%%%%%%%%%%%%%%%%%%%%%%%%%%%%%%%%%%

We can observe several interesting trends from network properties above. All measures undergo significant change at the time of the fall of the Soviet Union. Density peaks in 1980-90 and declines after that. This implies that, in fact, countries start to be less entangled by the topics they discuss. This is not something we would expect but it seems intuitive given the ideological constraints the Cold War imposed on the UN members and their rhetoric.  

The clustering coefficient peaks around 1990 and keeps fluctuating at high levels after that. It is possible that the structure of the network underwent a phase transition, where after 1990 various epistemic communities in the network started to be more tightly connected. Even if individual countries may on average be less entangled, the communities they belong to appear to have stronger connection. In other words, more countries connect the different epistemic policy communities. 

The average path length was at minimum at the fall of the Soviet Union, then it started to steadily increase, meaning that, on average, it takes more for a node to reach another node. This is in line with the decline in density. 

Overall, it seems that after the Cold War the way policies were discussed at the UN changed. Our results suggest that communities of countries talking about similar issues became more neatly connected, while the overall connection in the networks increased. These may be explained by several factors. One possibility is that during the Cold War era UN was focused on smaller number of issues that were of concern to most countries. In the post-Cold War period we observe a multipolarity of the policy topics discussed by countries and also the way countries discuss them becomes much more diverse compared to the previous period. Countries may have also undergone a process of `specialization' on policy topics, where in their speeches countries focus on some topics more than on other. For example, small island states have increasingly discussed the issue of global warming and catastrophic climate change. At the same time, some countries have developed into brokers across various epistemic policy communities.

\subsection{Community detection}

It is difficult to discern any meaningful patterns from simple network graphs like Figure \ref{fig:network2014}, particularly when we intend to do this systematically across different networks.  We approach this problem by using a community detection method. This approach allows identification of strongly related modules that correspond to functionally important units. Here we extracted communities from our network using the map equation framework and its Infomap search algorithm \citep{rosvall2008maps}. The algorithm is designed to reveal the community structure both in weighted and directional networks by tracing the information flow along the links of the system. The outcome of the analysis is a simplified map of the network that also emphasizes structural regularities and their relationships. An example of identified communities for 1982 is presented in Table \ref{tab:communities1982}. The full set of community detection results is presented in Tables \ref{tab:communities1}-\ref{tab:communities10} in supplementary materials.

The example in Table \ref{tab:communities1982} highlights communities of countries based on their rhetoric at the height of the Cold War. For example, community \#5 captures most of the Soviet Bloc countries. This pattern is sufficiently consistent over the years to view it as a partial validation of the results. 

The composition and interpretation of the communities is a substantively interesting question that we plan to explore in later work. At this stage it is important to note that we shouldn't expect exact overlap between Infomap communities and formal alliances. General Debate captures a mixture of sincere and strategic preferences. These are different from highly strategic voting behavior in the UN that has been shown to be associated with formal alliances \citep{voeten:2000,voeten2013data,voeten2004resisting}. Thus networks and communities are based on topics  discussed in the debates. In its current implementation we cannot judge how these topics are discussed, which would provide additional level of differentiation between countries, but rather we capture something akin to epistemic communities.

\section{Conclusion and next steps}

In the current draft we performed topology analysis of networks of countries built on the semantics of their UN General Debates. We build on standard social network analysis of textual data \citep[for a recent example see][]{waumans2015topology} and contribute to the literature by introducing information theoretic insights at the network building stage and map equation framework at the stage of identifying topological features of our networks. At the same time our current approach is not fully satisfactory as it doesn't take into account the temporal dimension of our data (as evident from a large number of annual network figures and community tables in supplementary materials). The aspect of time here is particularly important since it taps into the institutional focus of much of political science research.  

In the next stage of this project we plan to use Topological Data Analysis (TDA), which is a methodological framework for identifying topological structure in data \citep{carlsson2009topology}. TDA builds on recent advances in computational topology that makes it possible compute topological invariants from data. One approach is persistent homology \citep{edelsbrunner2010computational}, which captures how features in the space can become apparent under certain filtering solutions, and allows to study the homology at multiple scales at the same time. More importantly for our case, and relevant for a wider application in political science, persistent homology provides a framework to quantify the temporal development of the topology of international networks.

\clearpage \singlespacing
\bibliographystyle{apsr}
\bibliography{un}

%%%%%%%%%%%%%%%%%%%%%%%%%%%%%%%%%%%%%%%% 

\onecolumn
\begin{table}
\centering
\tiny
\begin{tabular}{c c p{12cm}}
\toprule
Year & Community & ISO Country Code \\
\midrule
1982	&	0	&	AUS, AUT, BGD, BTN, COD, CYP, DNK, EGY, FIN, FJI, GBR, GRC, IDN, IND, IRL, ITA, JPN, LKA, LSO, LUX, MDV, MMR, NLD, NOR, NPL, NZL, PAK, PHL, PNG, POL, SWE, TUR, TZA, YUG	\\
1982	&	1	&	AGO, BEL, BLZ, BOL, CHL, COG, COL, CRI, DZA, GAB, GIN, GNB, HND, MDG, PER, PRY, SGP, SUR, TGO, TUN, USA, VEN	\\
1982	&	2	&	AFG, BEN, BHR, DDR, DJI, ETH, GRD, GUY, IRN, JAM, LBY, MAR, MOZ, NGA, SLE, SOM, STP, VCT, VNM, YDY, ZWE	\\
1982	&	3	&	BDI, BFA, BHS, COM, CPV, HTI, IRQ, KHM, KWT, MLI, MRT, MUS, NER, SEN, TCD	\\
1982	&	4	&	ARE, ATG, BRB, CAN, GMB, ISR, KEN, LBR, RWA, SDN, TTO, UGA, ZMB	\\
1982	&	5	&	ALB, BGR, BLR, CSK, CUB, HUN, LAO, MNG, NIC, RUS, UKR	\\
1982	&	6	&	BRA, ESP, FRA, MEX, SLV	\\
1982	&	7	&	ARG, ECU, MLT, PAN	\\
1982	&	8	&	BWA, JOR, QAT, THA	\\
1982	&	9	&	CAF, DEU, DOM, GNQ, GTM, PRT, URY	\\
1982	&	10	&	ISL, ROU	\\
1982	&	11	&	CHN, YEM	\\
1982	&	12	&	CMR, GHA, OMN	\\
1982	&	13	&	SAU, SYR	\\
1982	&	14	&	MYS	\\
\bottomrule

\end{tabular}
\caption{\emph{Communities in the 1982 UNGA General Debate network.} Map equation framework and Infomap search algorithm were used to identify communities. \label{tab:communities1982}}

\end{table}
%%%%%%%%%%%%%%%%%%%%%%%%%%%%%%%%%%%%%%%% 

\clearpage
\appendix

\section{Supplementary materials}

\subsection{Networks}
%%%%%%%%%%%%%%%%%%%%%%%%%%%%%%%%%%%%%%%%%
%FIGURE: Networks based on MI measure for distance
\begin{figure}
\centering
\subfloat[1970]{\includegraphics[width=.5\textwidth]{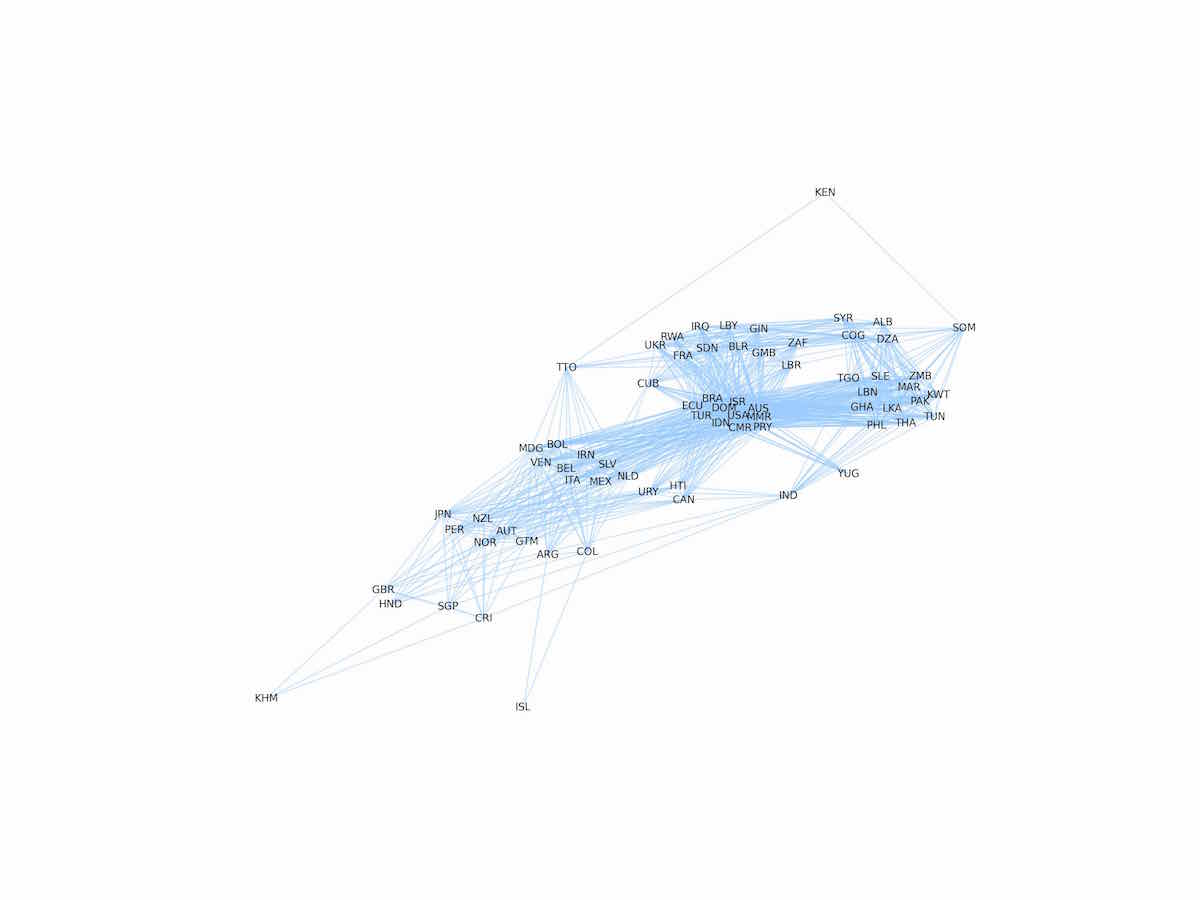}}
\subfloat[1971]{\includegraphics[width=.5\textwidth]{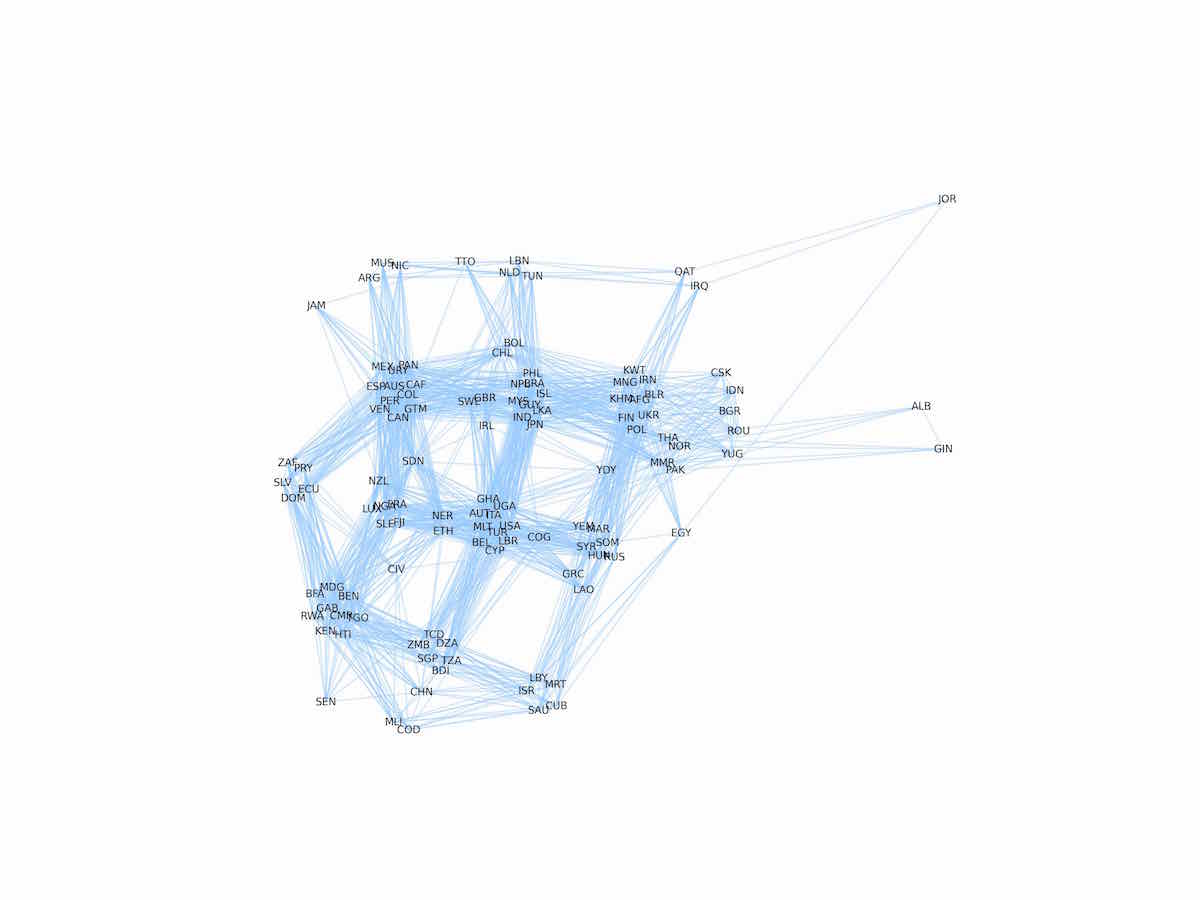}}\\
\subfloat[1972]{\includegraphics[width=.5\textwidth]{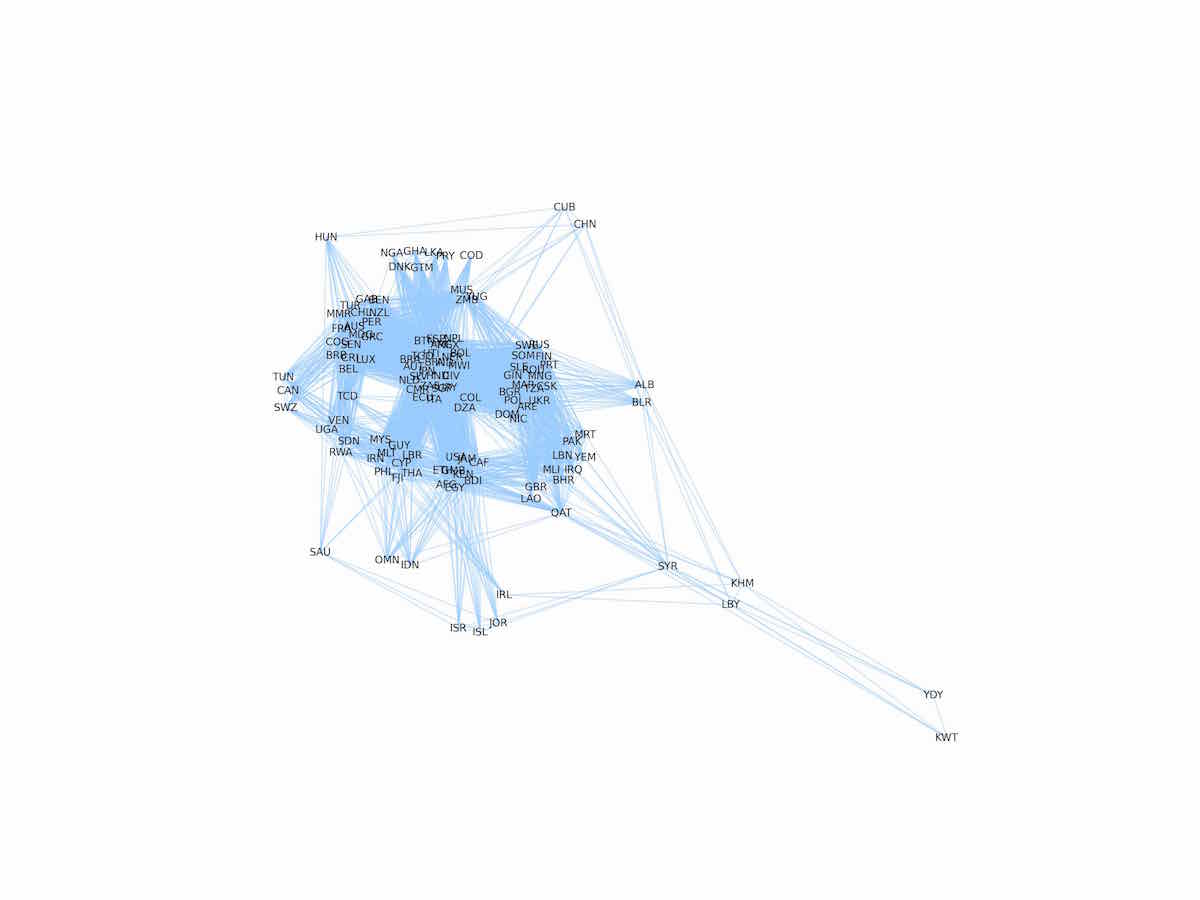}}
\subfloat[1973]{\includegraphics[width=.5\textwidth]{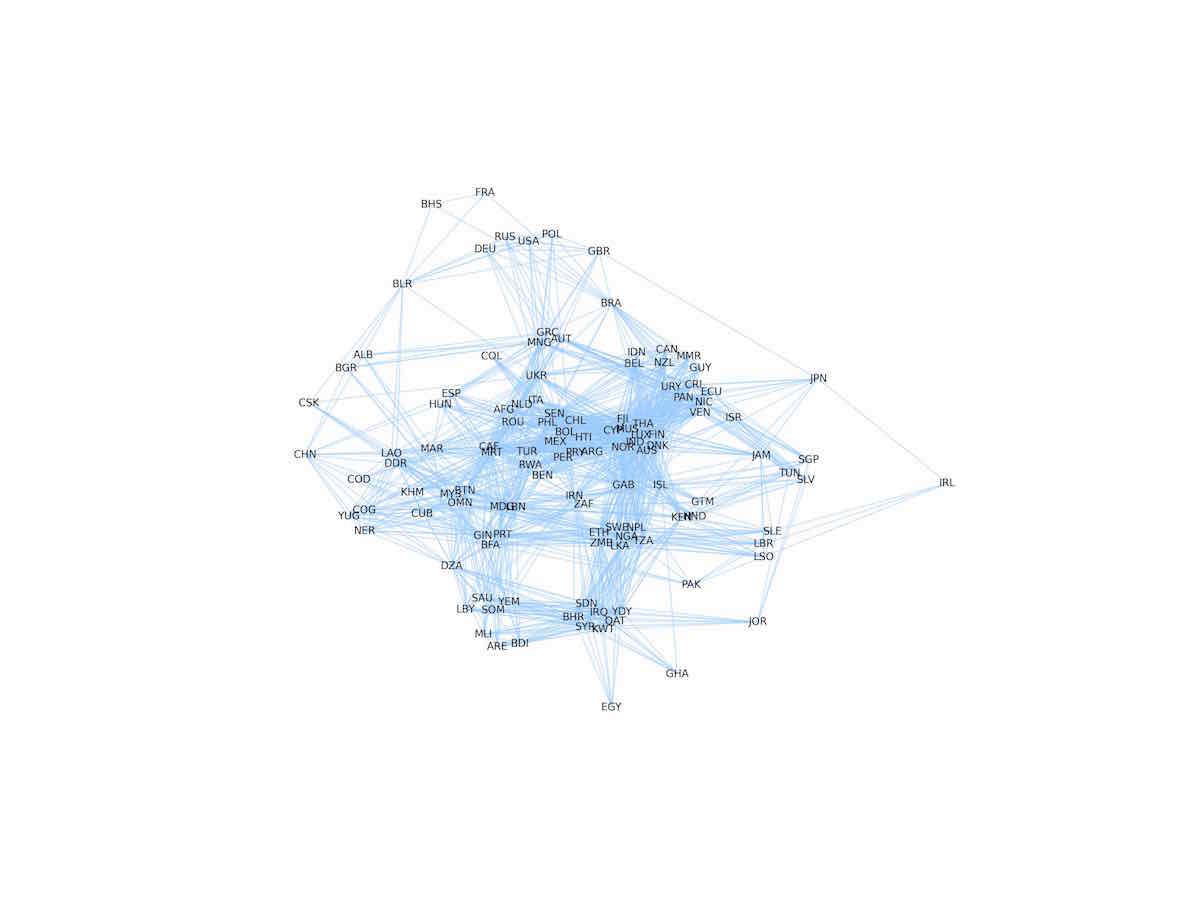}}\\
\subfloat[1974]{\includegraphics[width=.5\textwidth]{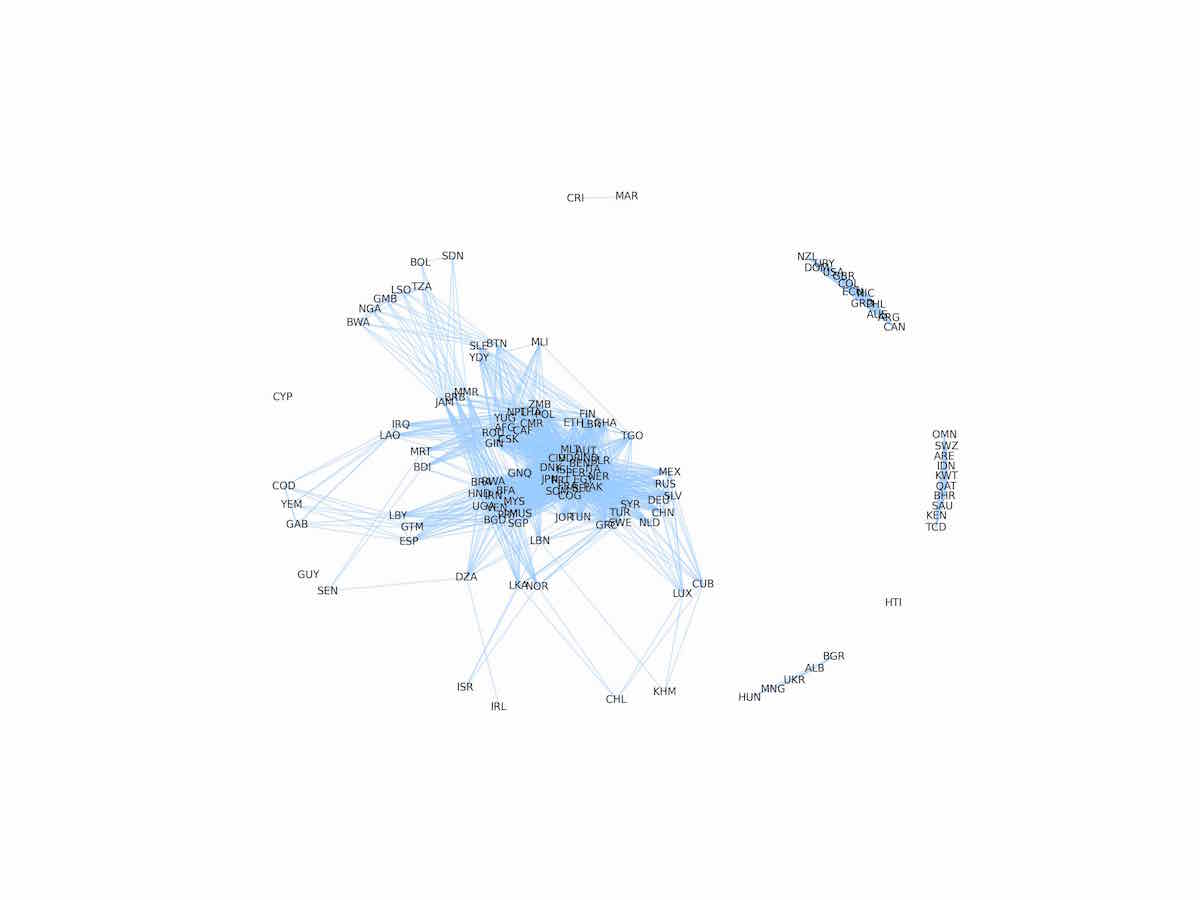}}
\subfloat[1975]{\includegraphics[width=.5\textwidth]{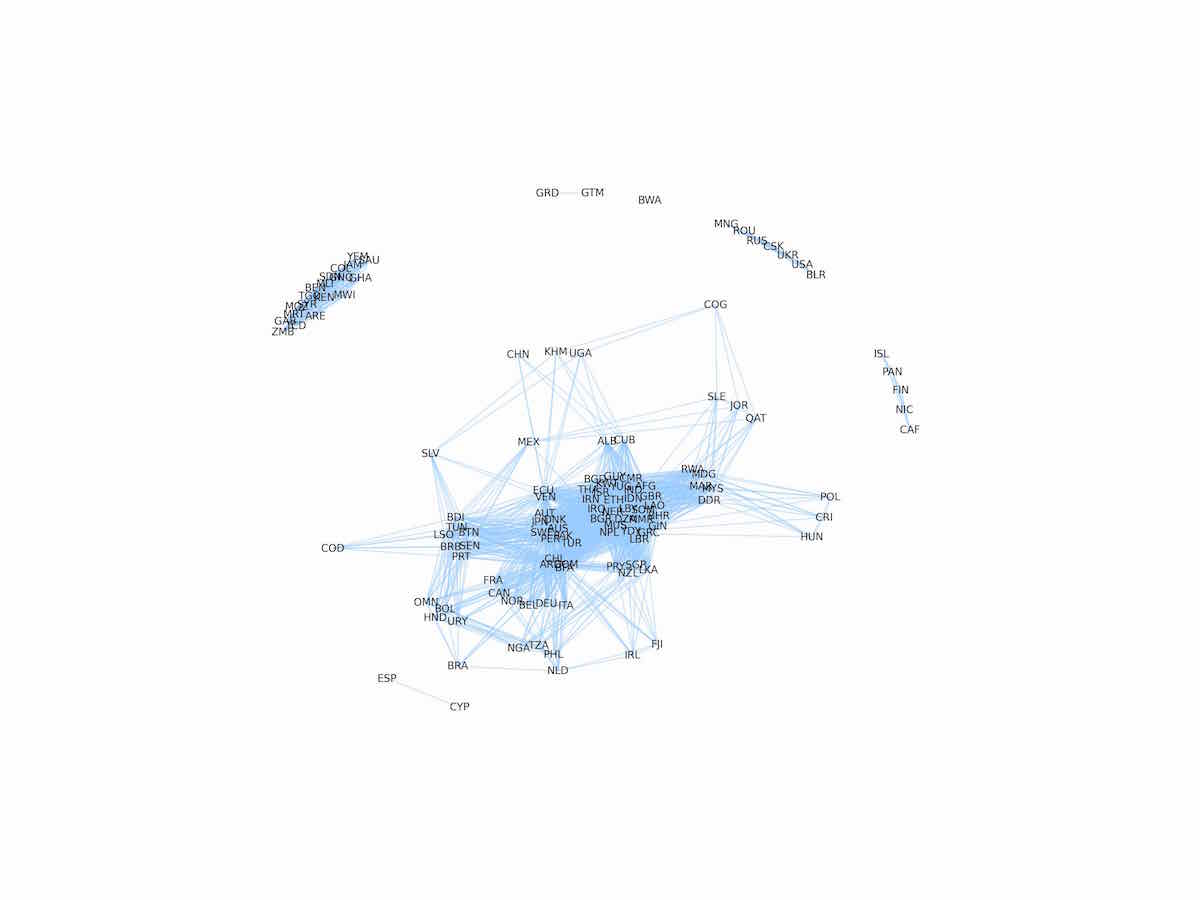}}

\caption{\emph{Networks from speeches, 1970-1975}. Networks built based on topics discussed in annual UNGA General Debate speeches. Links between countries is established using normalized mutual information criteria.
\label{fig:networks1}}

\end{figure}
%%%%%%%%%%%%%%%%%%%%%%%%%%%%%%%%%%%%%%%%%

%%%%%%%%%%%%%%%%%%%%%%%%%%%%%%%%%%%%%%%%%
%FIGURE: Networks based on MI measure for distance
\begin{figure}
\centering
\subfloat[1976]{\includegraphics[width=.5\textwidth]{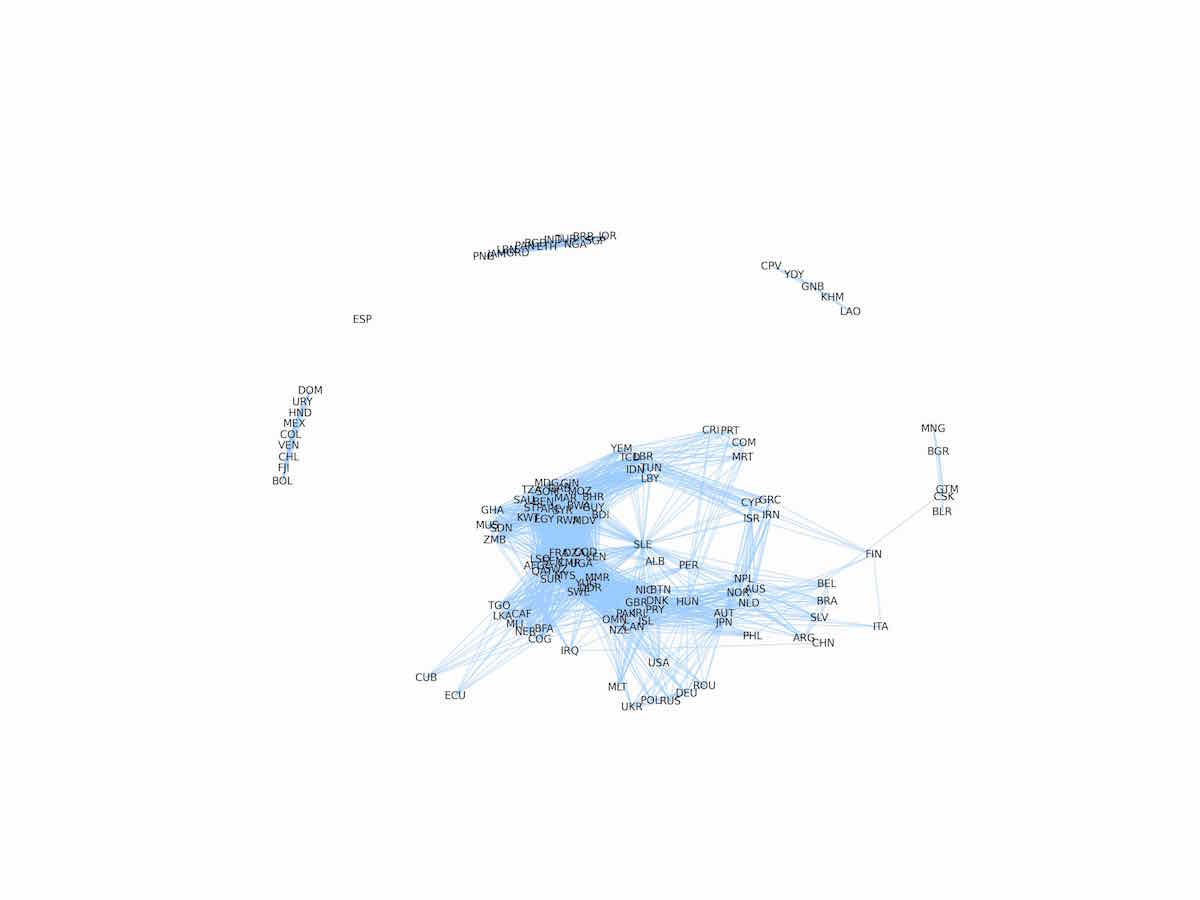}}
\subfloat[1977]{\includegraphics[width=.5\textwidth]{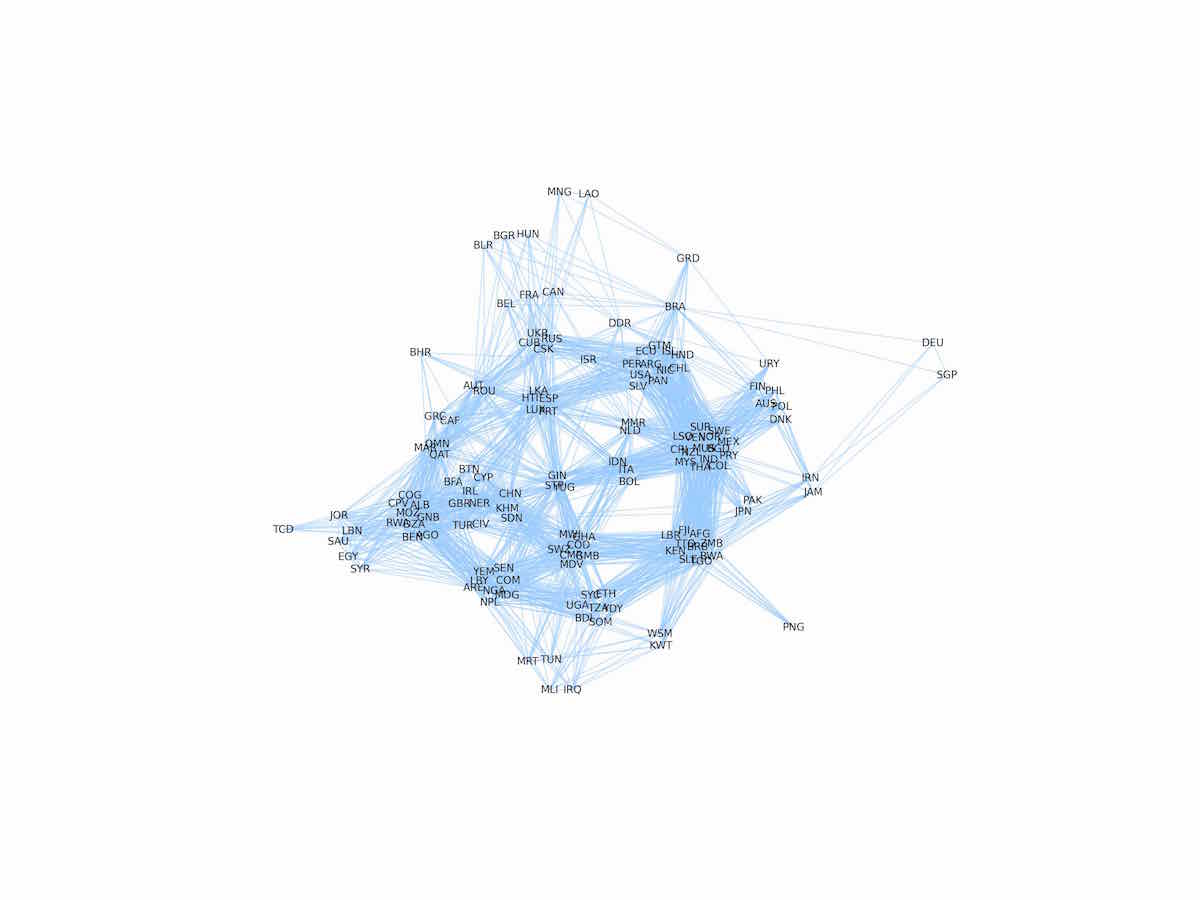}}\\
\subfloat[1978]{\includegraphics[width=.5\textwidth]{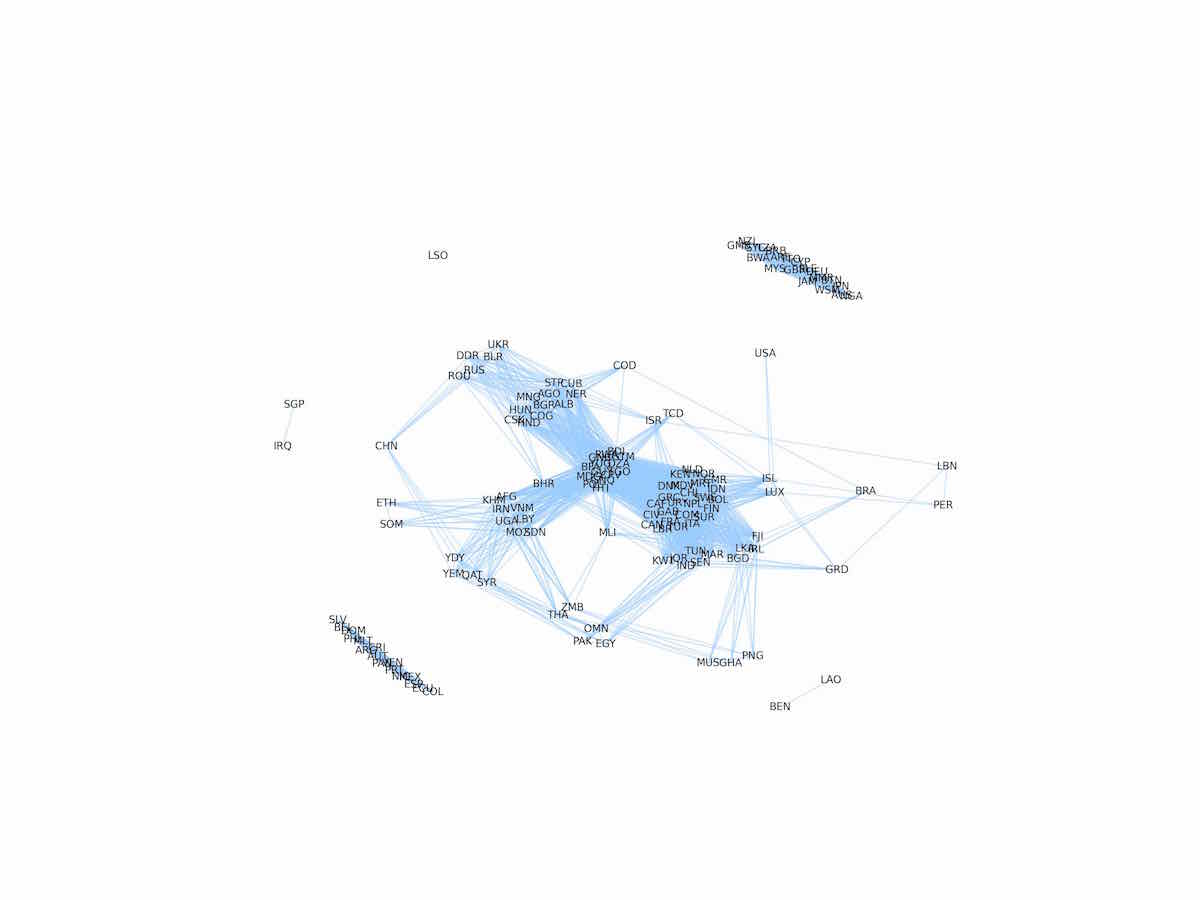}}
\subfloat[1979]{\includegraphics[width=.5\textwidth]{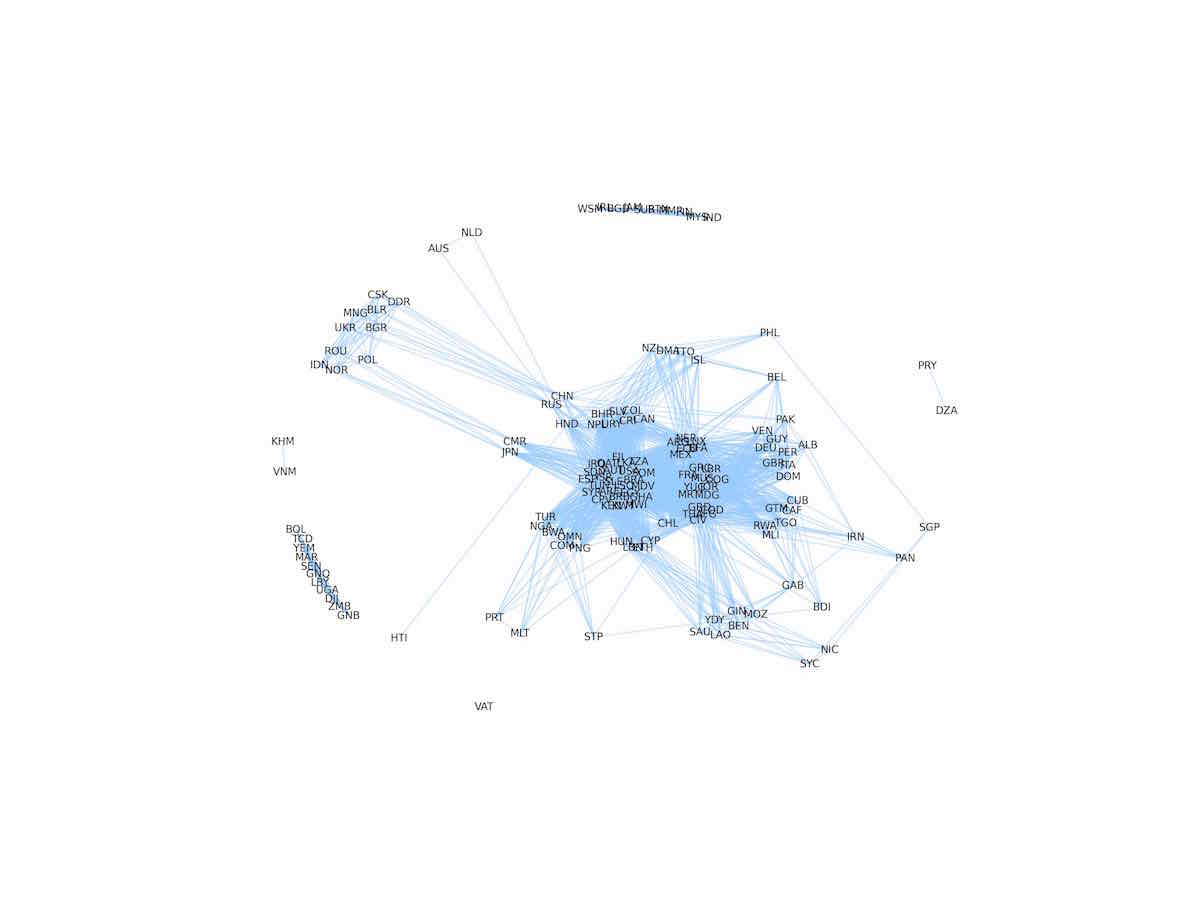}}\\
\subfloat[1980]{\includegraphics[width=.5\textwidth]{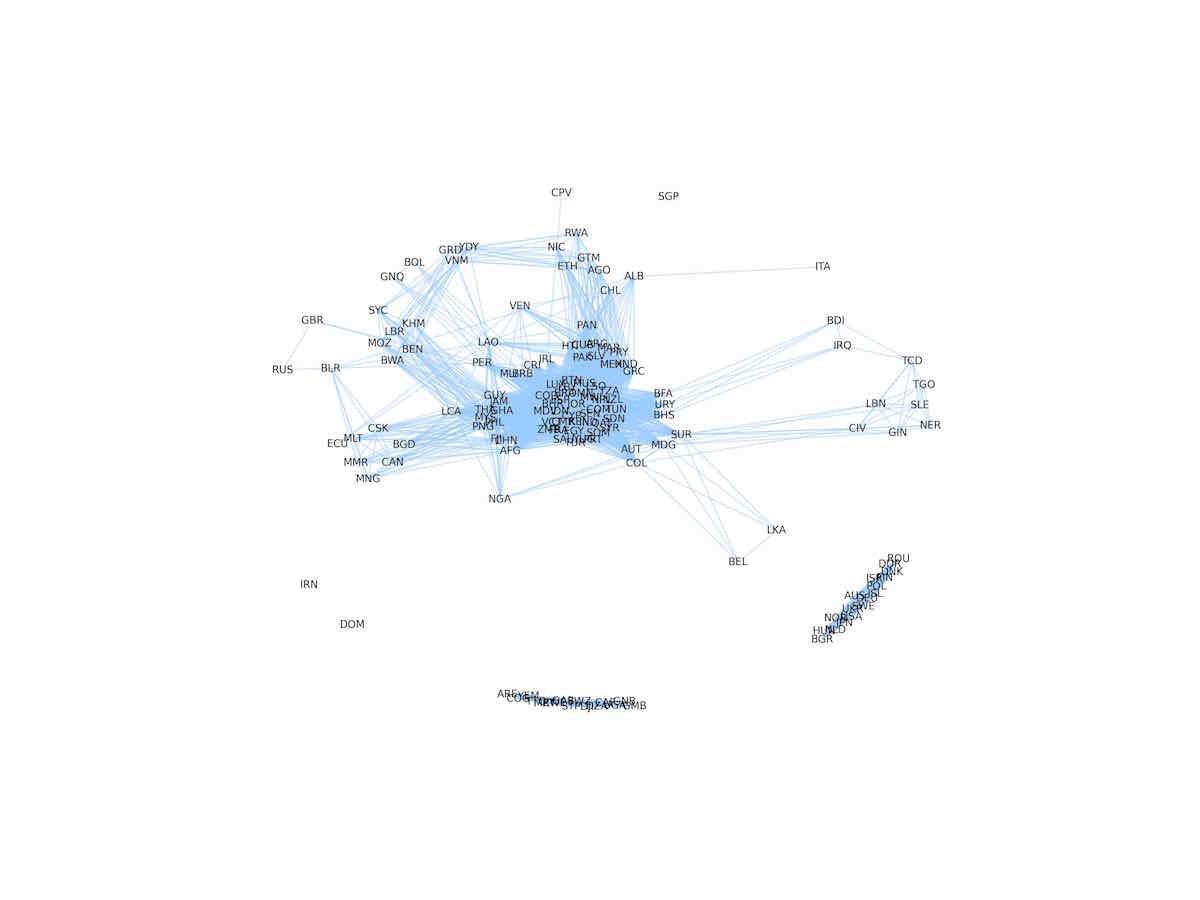}}
\subfloat[1981]{\includegraphics[width=.5\textwidth]{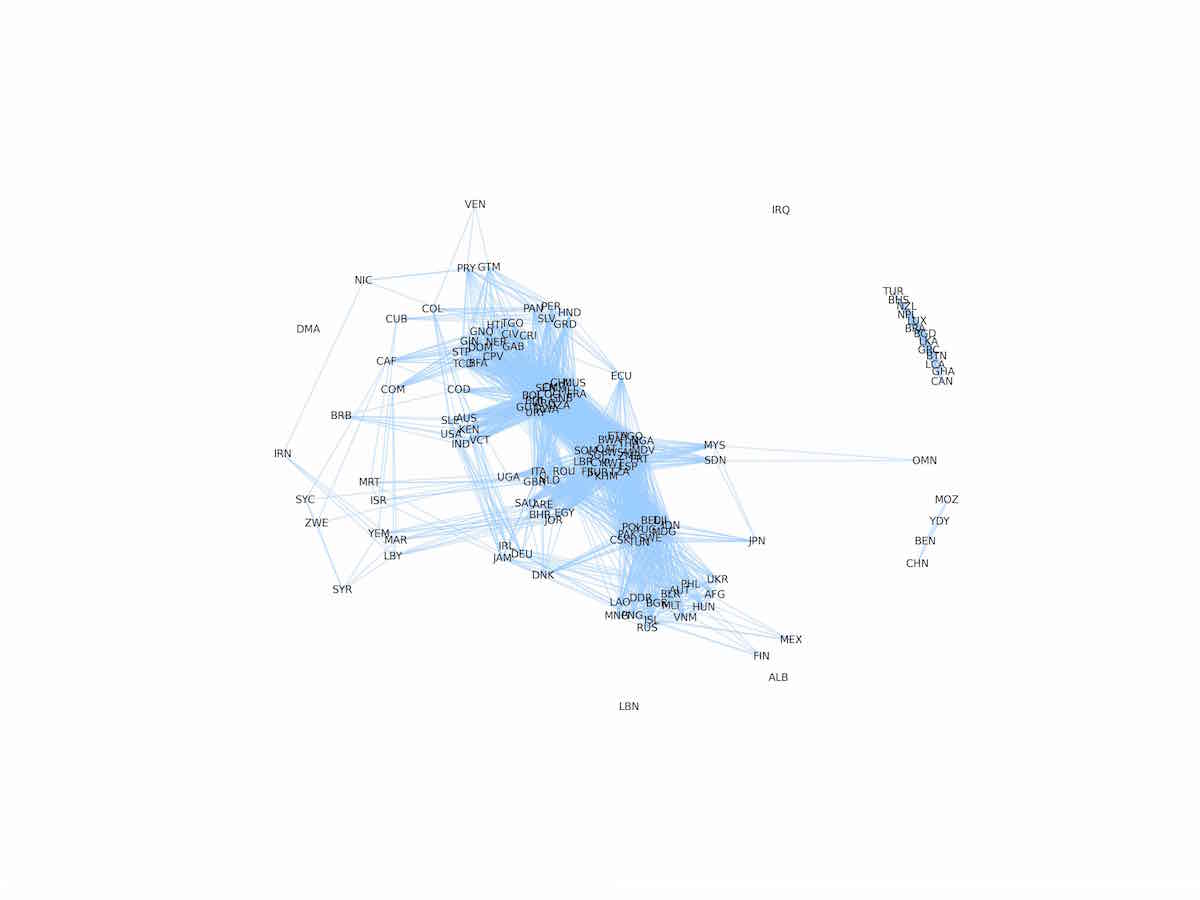}}

\caption{\emph{Networks from speeches, 1976-1981}. Networks built based on topics discussed in annual UNGA General Debate speeches. Links between countries is established using normalized mutual information criteria.
\label{fig:networks2}}

\end{figure}
%%%%%%%%%%%%%%%%%%%%%%%%%%%%%%%%%%%%%%%%%

%%%%%%%%%%%%%%%%%%%%%%%%%%%%%%%%%%%%%%%%%
%FIGURE: Networks based on MI measure for distance
\begin{figure}
\centering
\subfloat[1982]{\includegraphics[width=.5\textwidth]{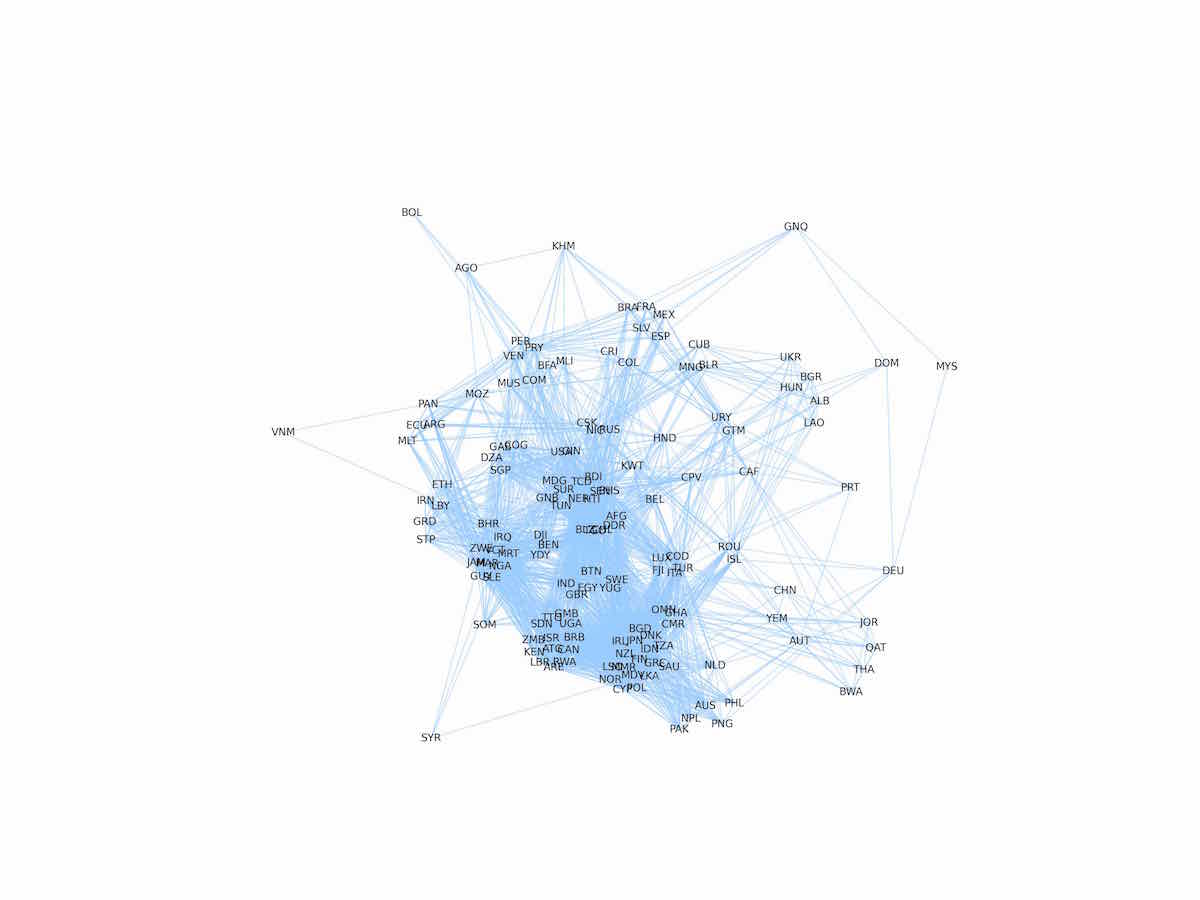}}
\subfloat[1983]{\includegraphics[width=.5\textwidth]{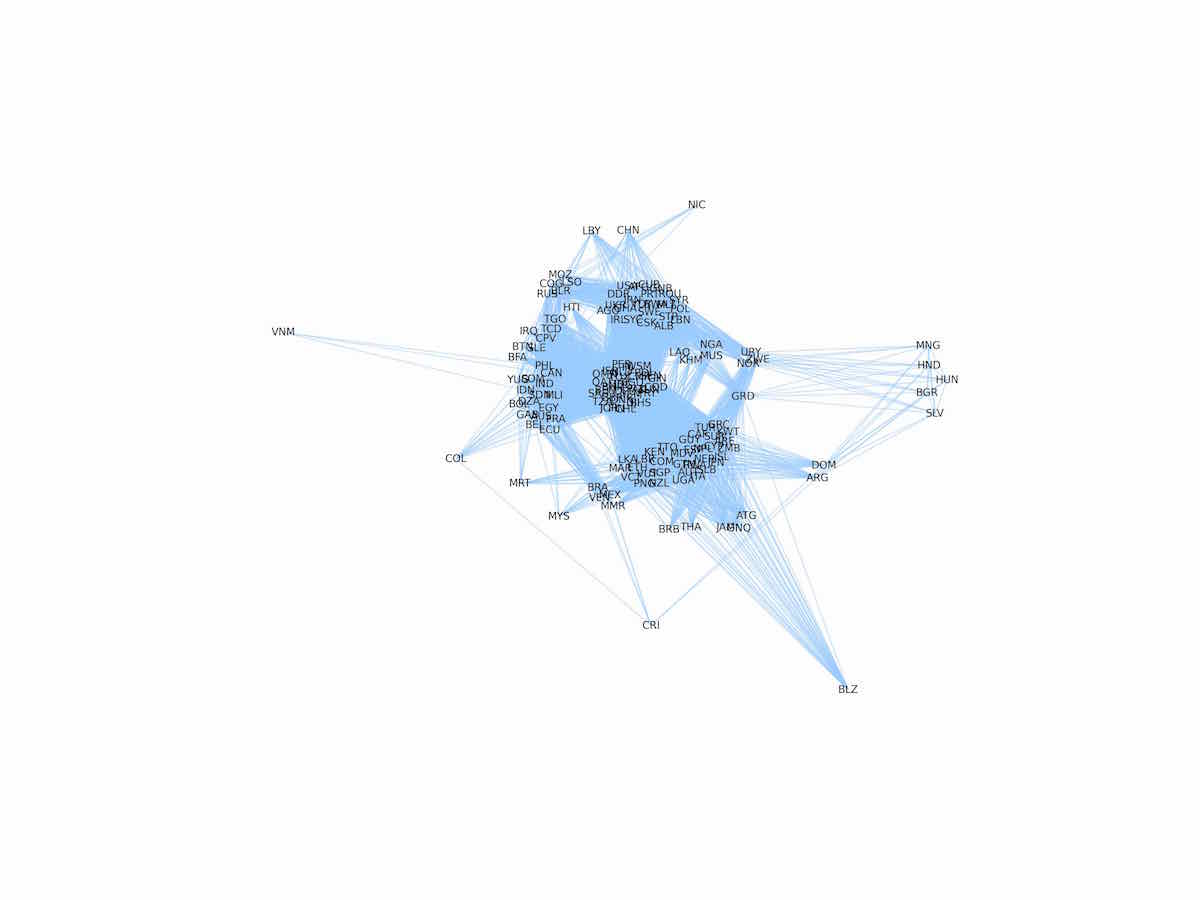}}\\
\subfloat[1984]{\includegraphics[width=.5\textwidth]{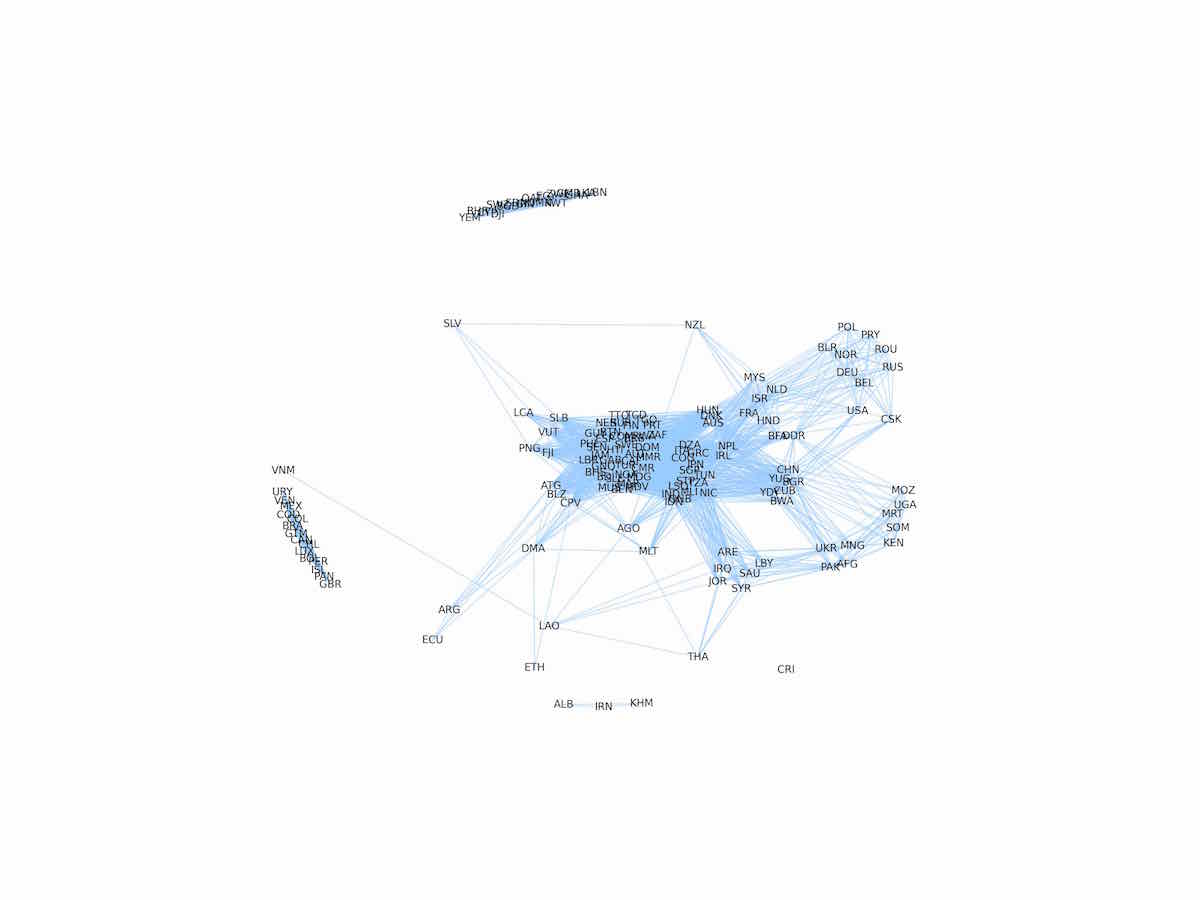}}
\subfloat[1985]{\includegraphics[width=.5\textwidth]{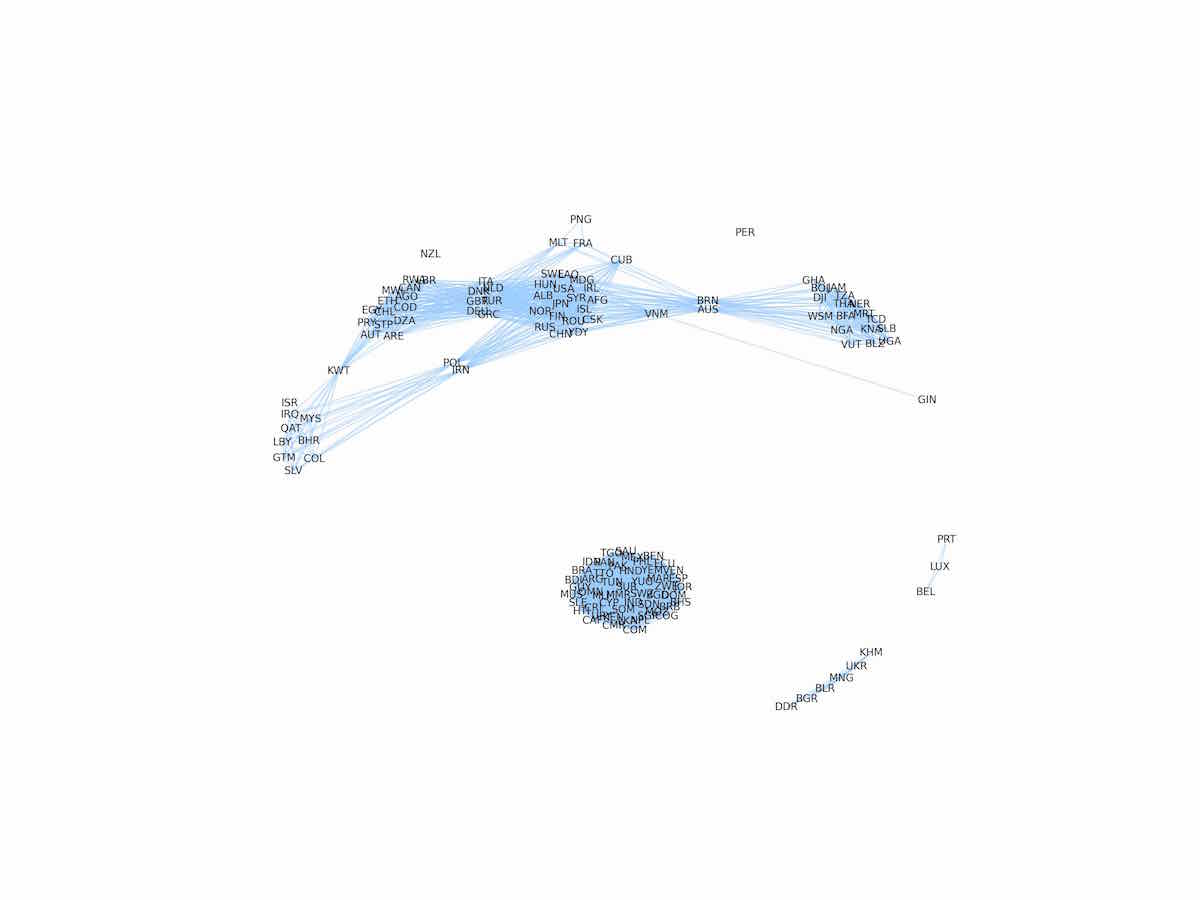}}\\
\subfloat[1986]{\includegraphics[width=.5\textwidth]{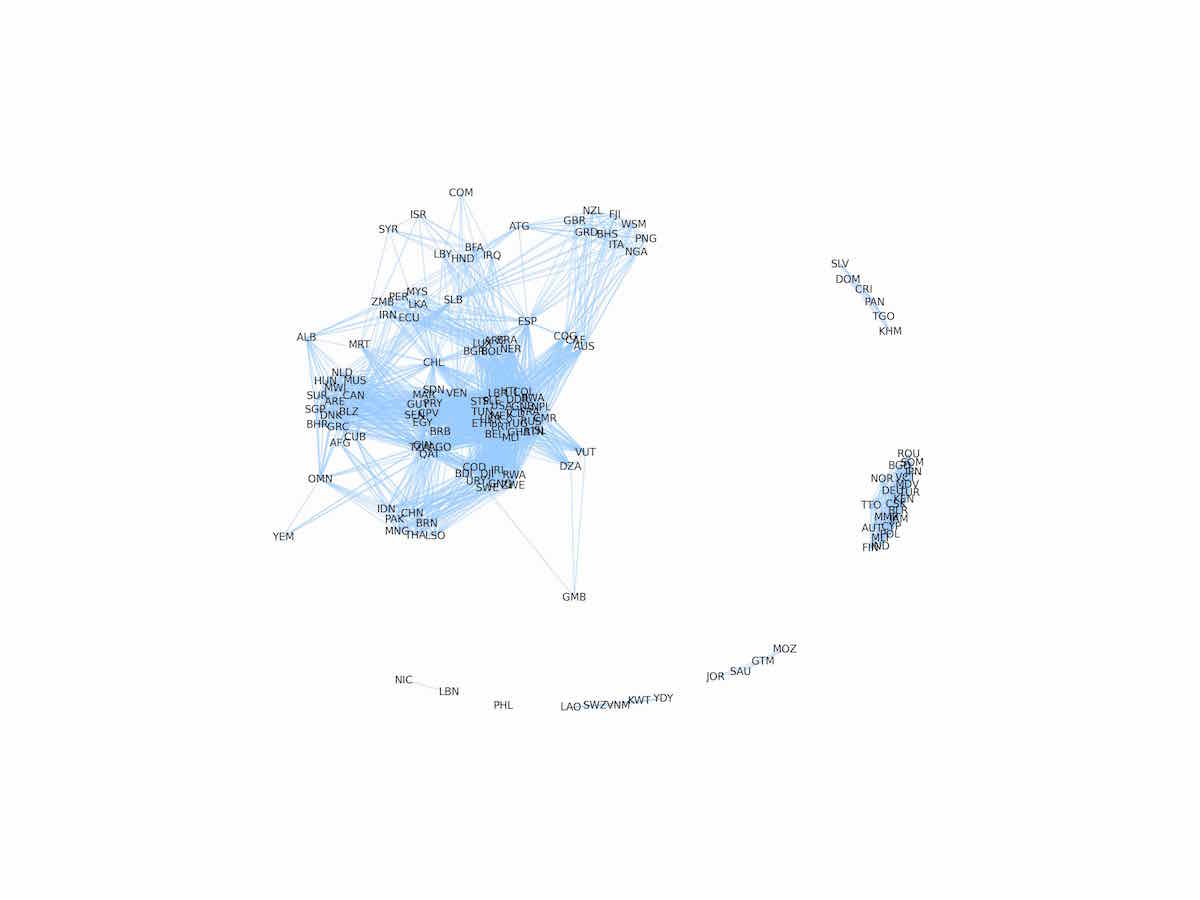}}
\subfloat[1987]{\includegraphics[width=.5\textwidth]{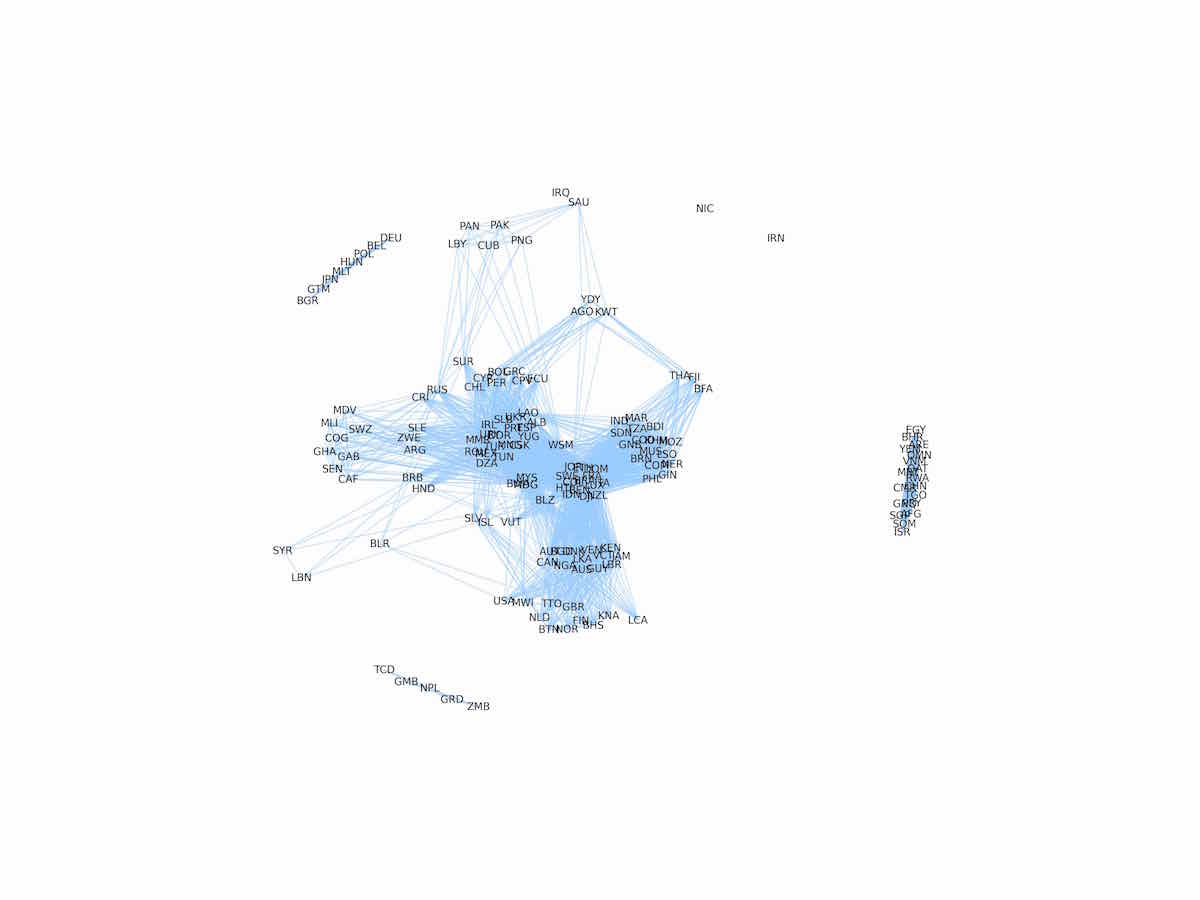}}

\caption{\emph{Networks from speeches, 1982-1987}. Networks built based on topics discussed in annual UNGA General Debate speeches. Links between countries is established using normalized mutual information criteria.
\label{fig:networks3}}

\end{figure}
%%%%%%%%%%%%%%%%%%%%%%%%%%%%%%%%%%%%%%%%%

%%%%%%%%%%%%%%%%%%%%%%%%%%%%%%%%%%%%%%%%%
%FIGURE: Networks based on MI measure for distance
\begin{figure}
\centering
\subfloat[1988]{\includegraphics[width=.5\textwidth]{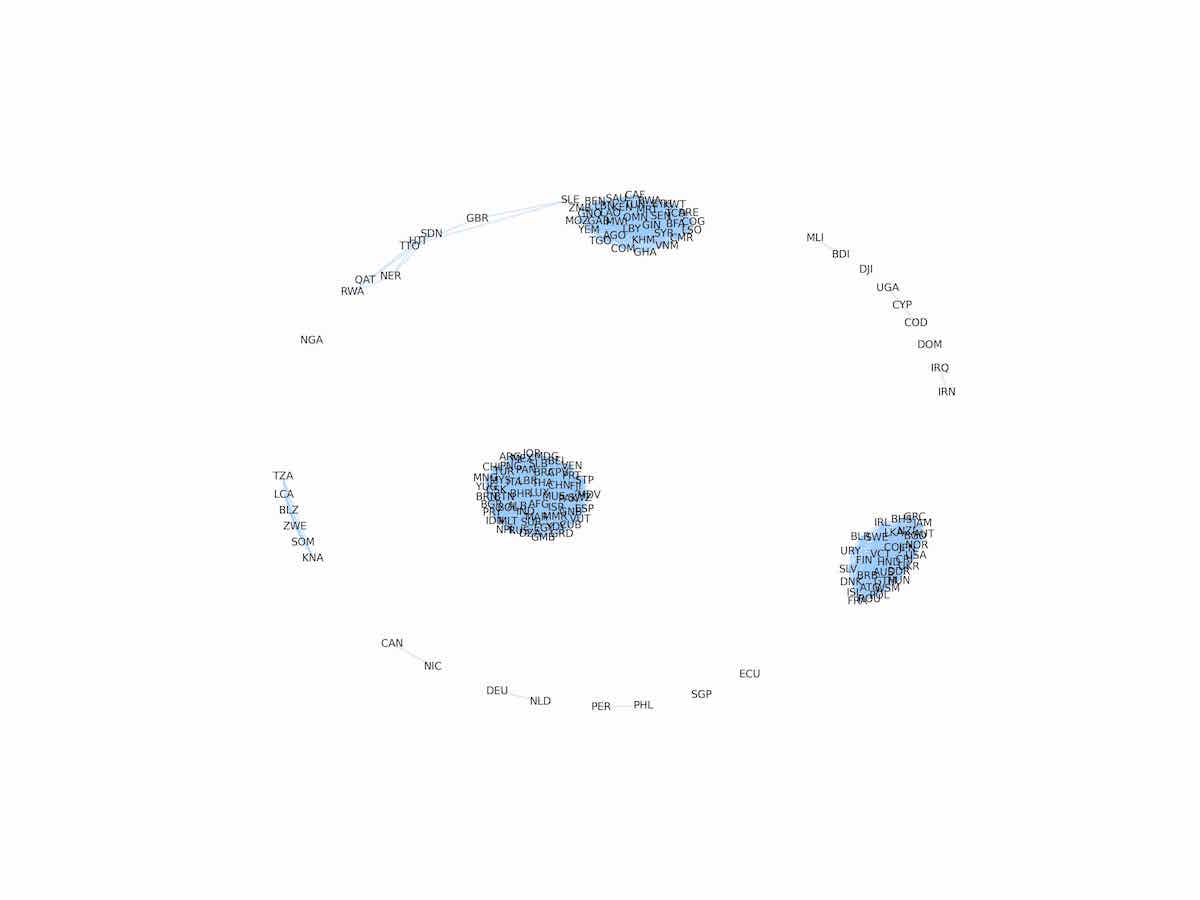}}
\subfloat[1989]{\includegraphics[width=.5\textwidth]{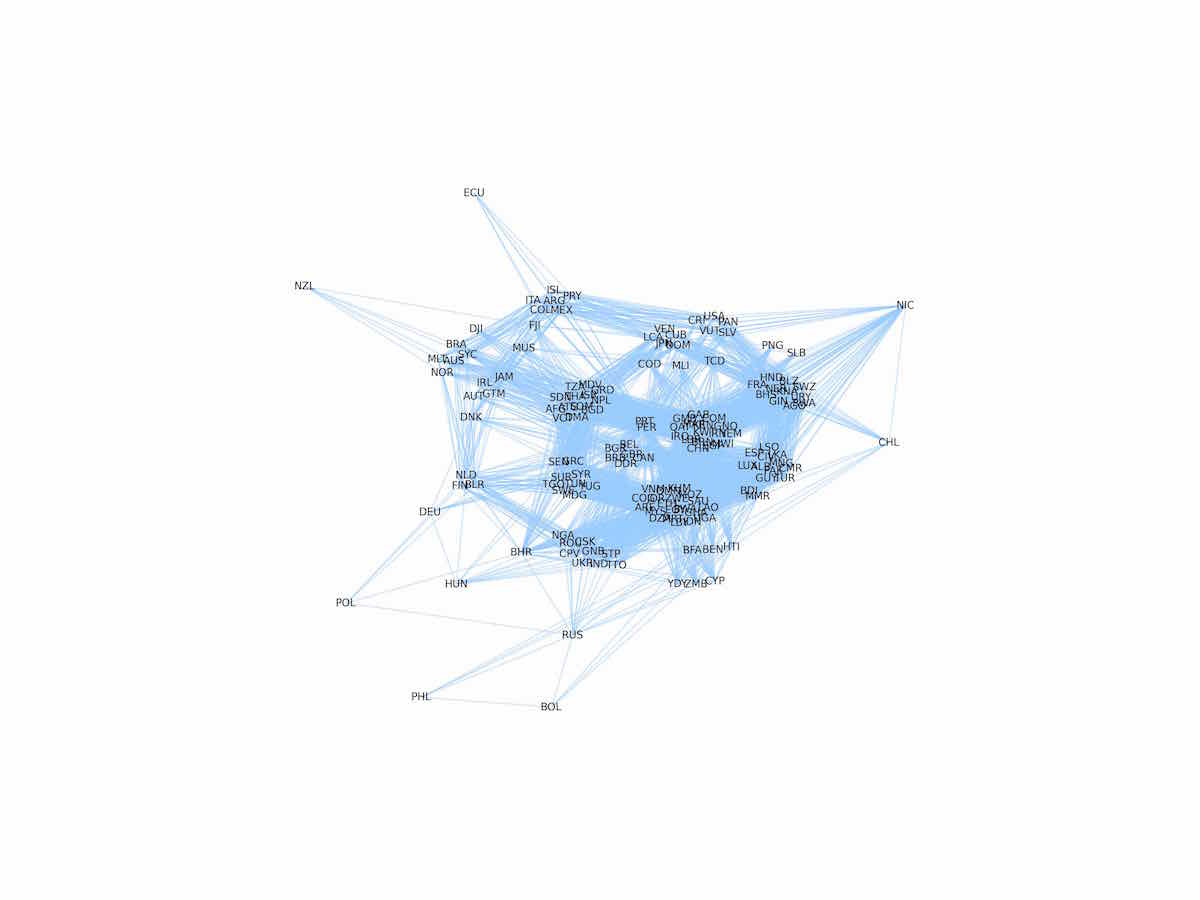}}\\
\subfloat[1990]{\includegraphics[width=.5\textwidth]{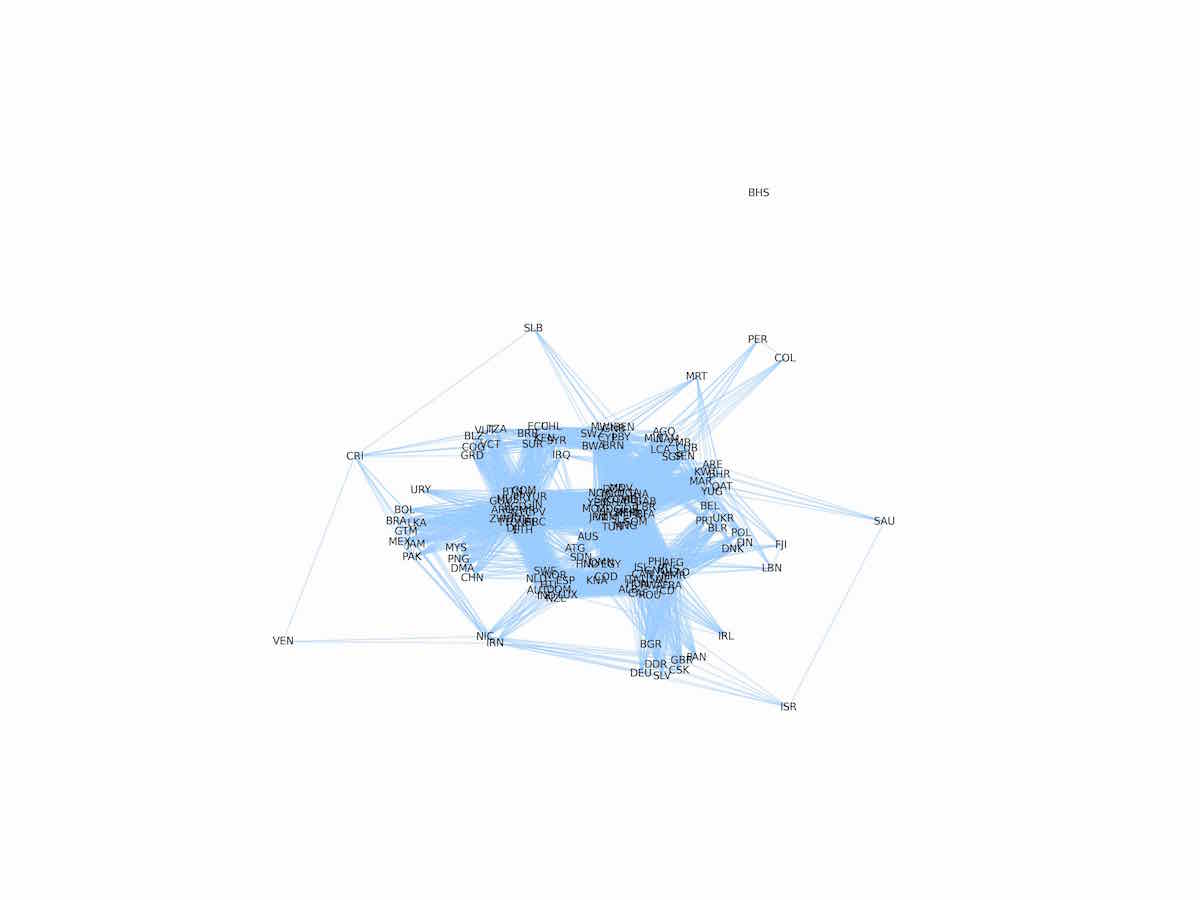}}
\subfloat[1991]{\includegraphics[width=.5\textwidth]{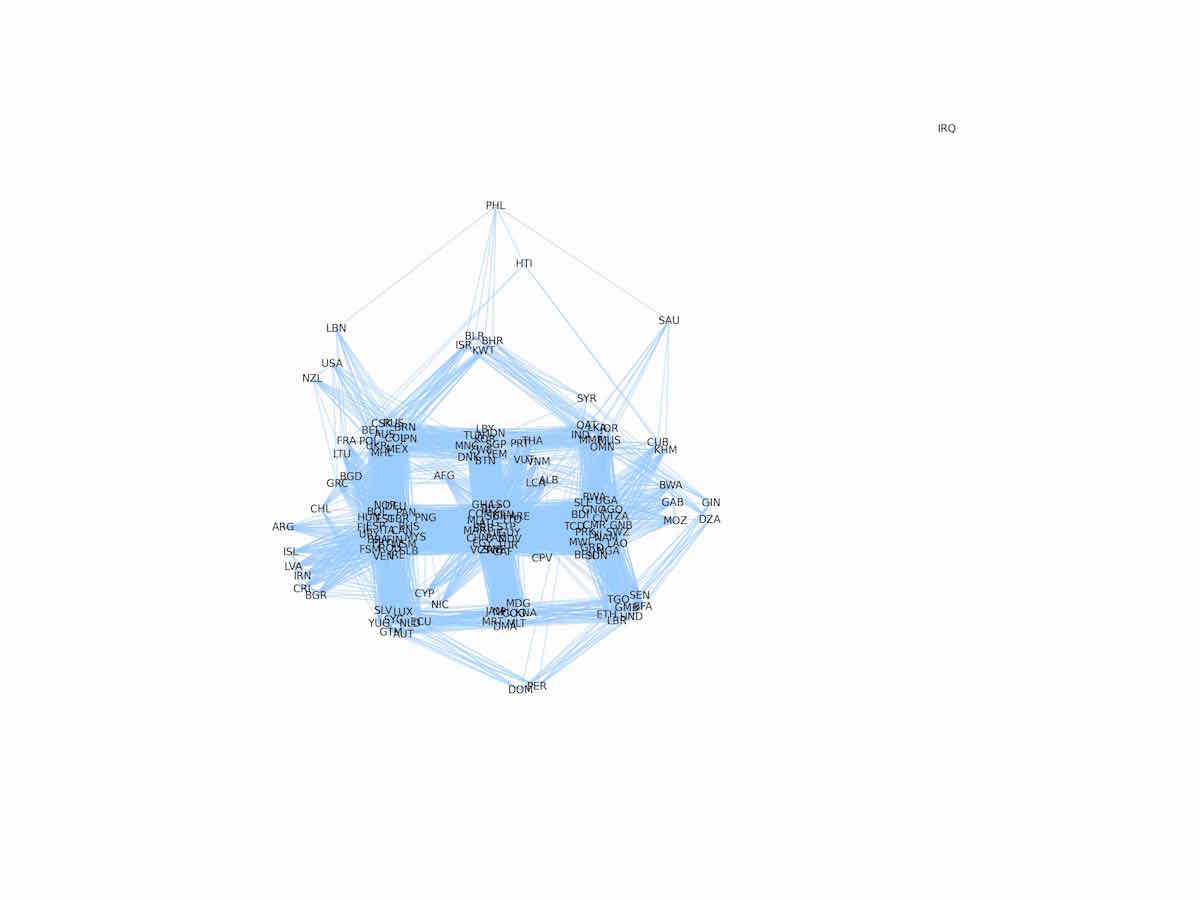}}\\
\subfloat[1992]{\includegraphics[width=.5\textwidth]{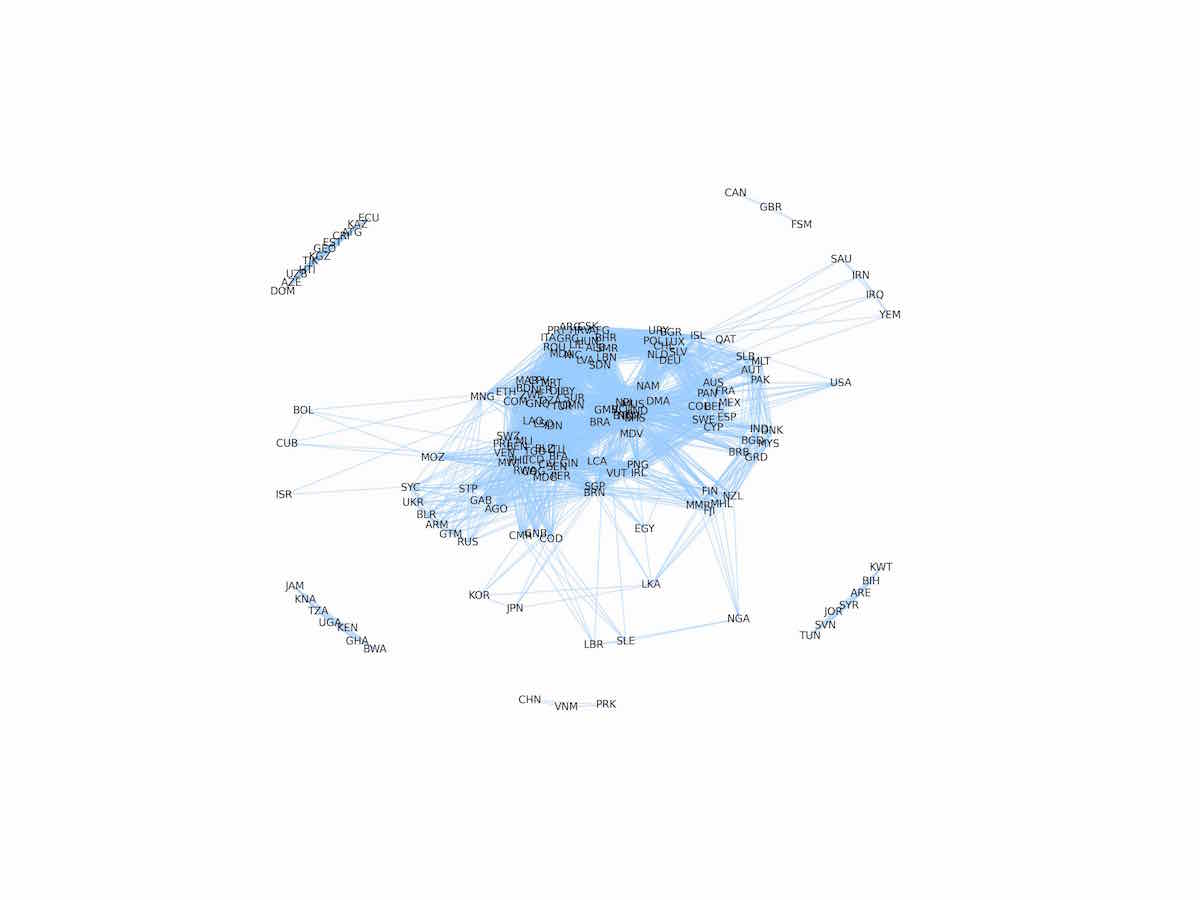}}
\subfloat[1993]{\includegraphics[width=.5\textwidth]{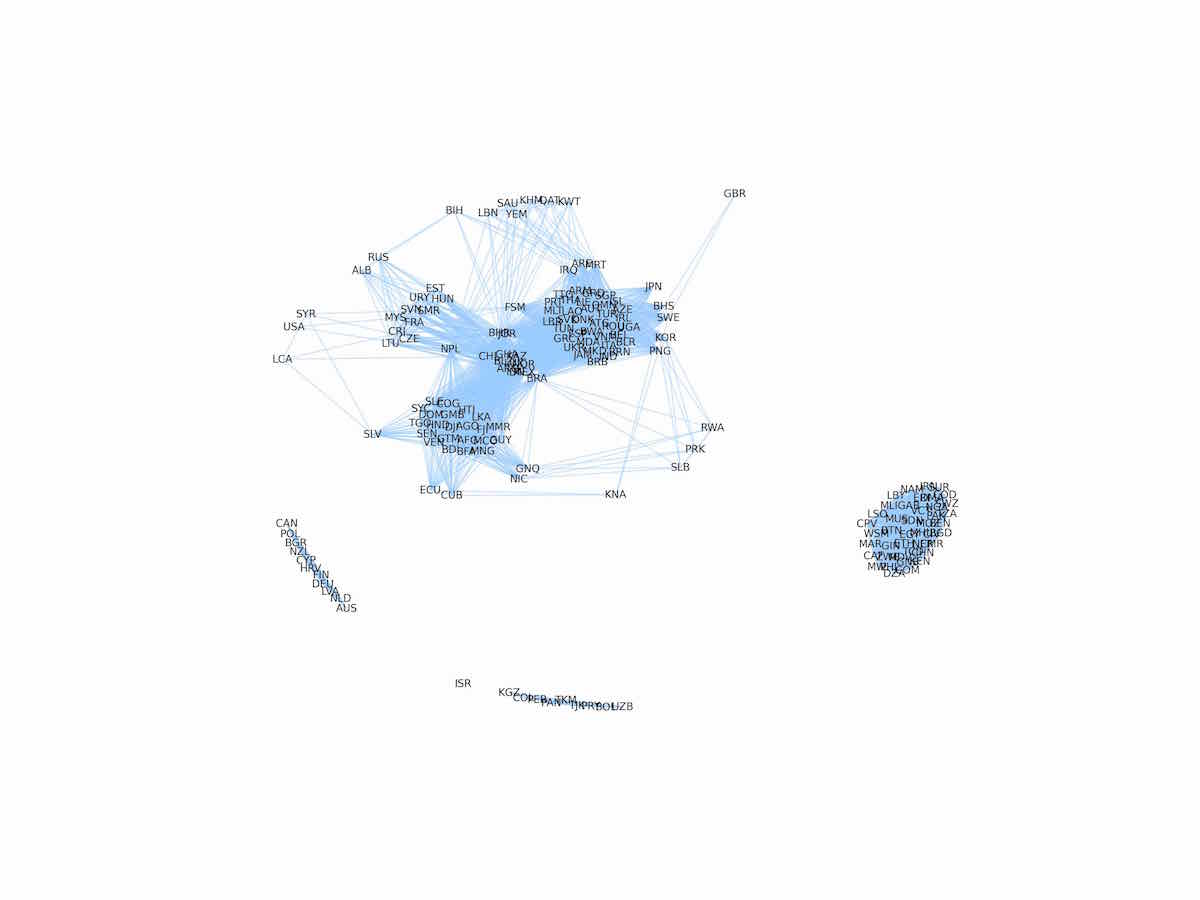}}

\caption{\emph{Networks from speeches, 1988-1993}. Networks built based on topics discussed in annual UNGA General Debate speeches. Links between countries is established using normalized mutual information criteria.
\label{fig:networks4}}

\end{figure}
%%%%%%%%%%%%%%%%%%%%%%%%%%%%%%%%%%%%%%%%%

%%%%%%%%%%%%%%%%%%%%%%%%%%%%%%%%%%%%%%%%%
%FIGURE: Networks based on MI measure for distance
\begin{figure}
\centering
\subfloat[1994]{\includegraphics[width=.5\textwidth]{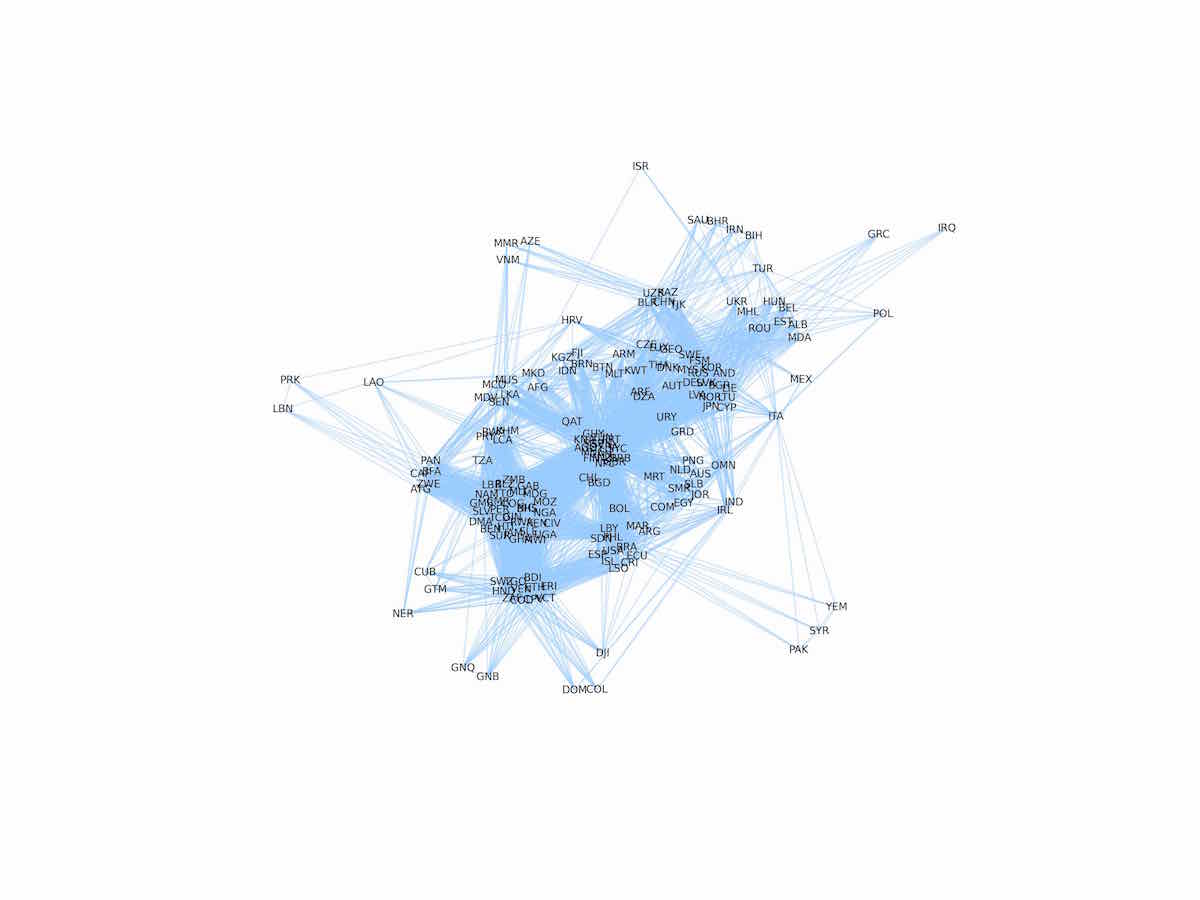}}
\subfloat[1995]{\includegraphics[width=.5\textwidth]{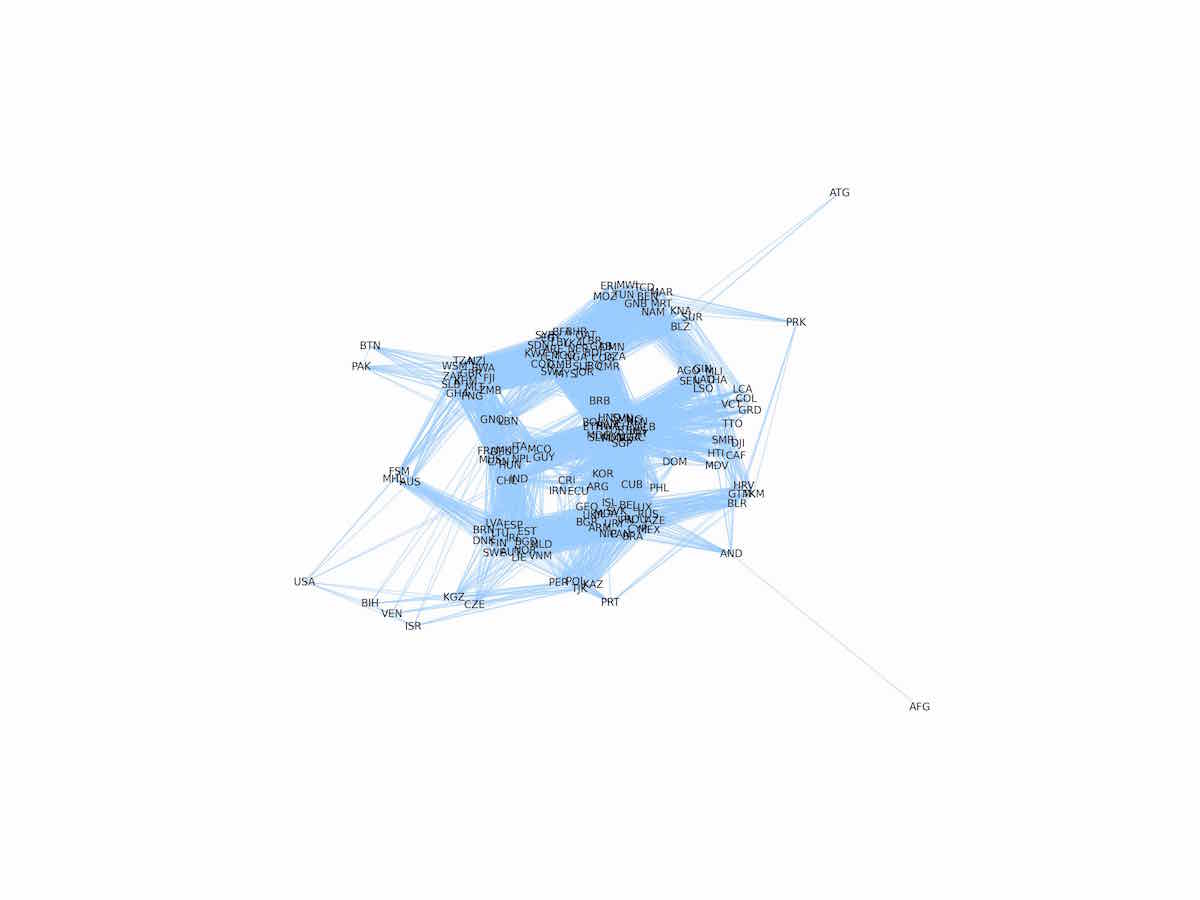}}\\
\subfloat[1996]{\includegraphics[width=.5\textwidth]{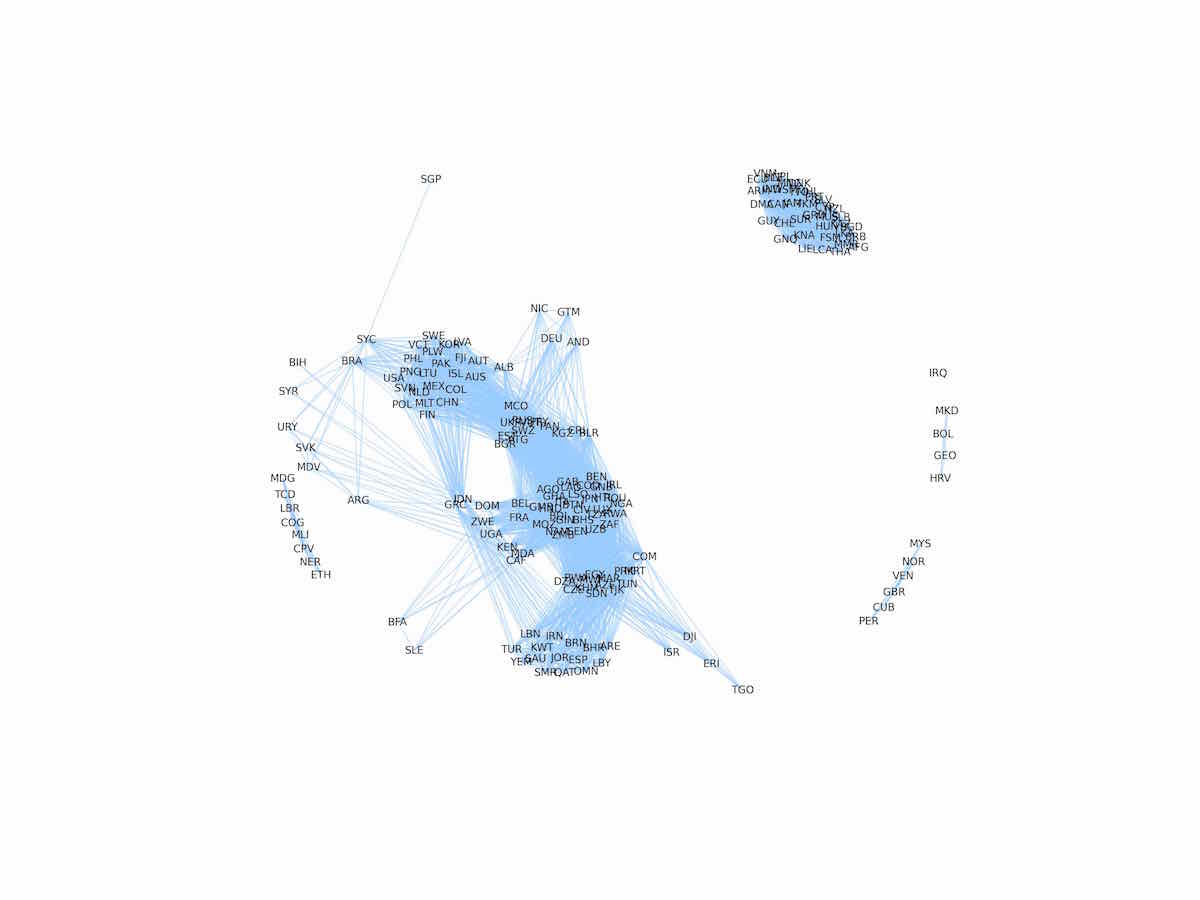}}
\subfloat[1997]{\includegraphics[width=.5\textwidth]{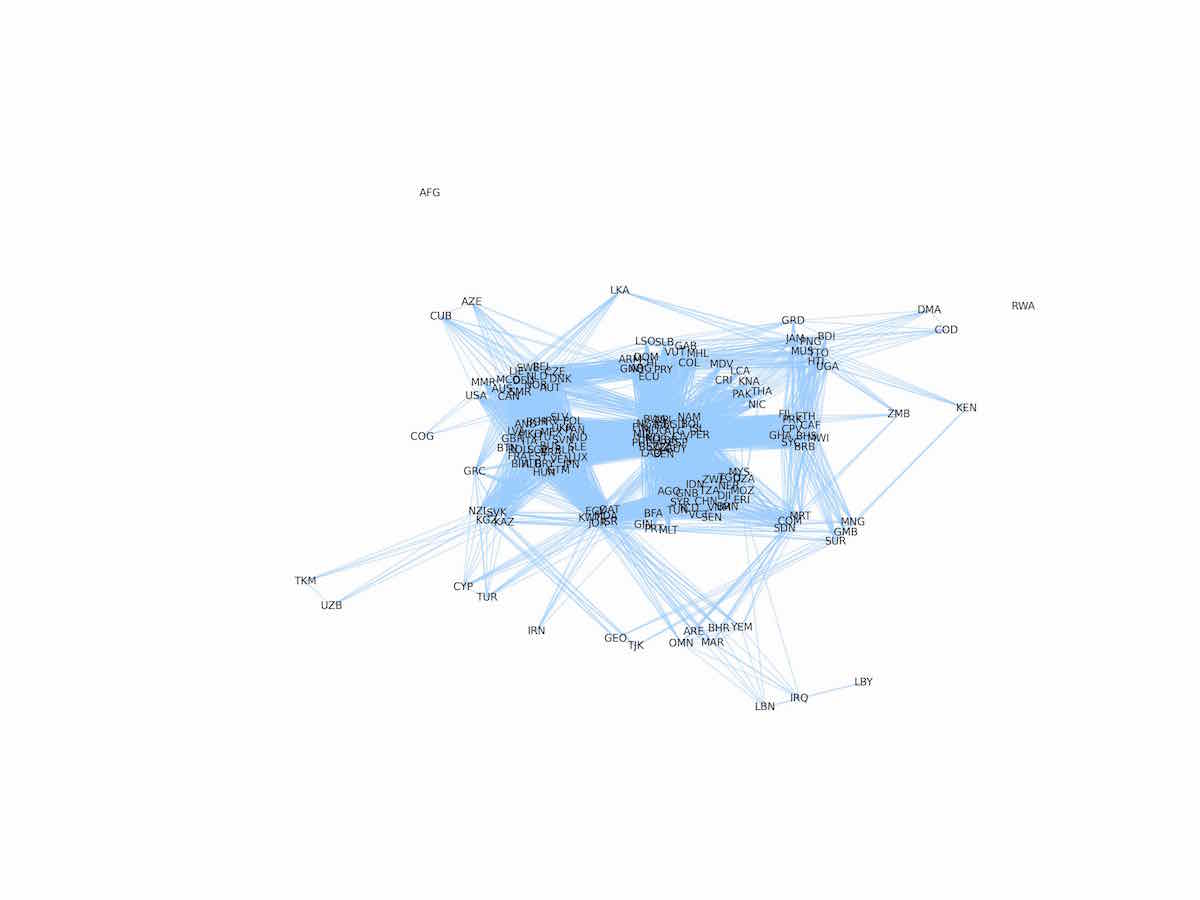}}\\
\subfloat[1998]{\includegraphics[width=.5\textwidth]{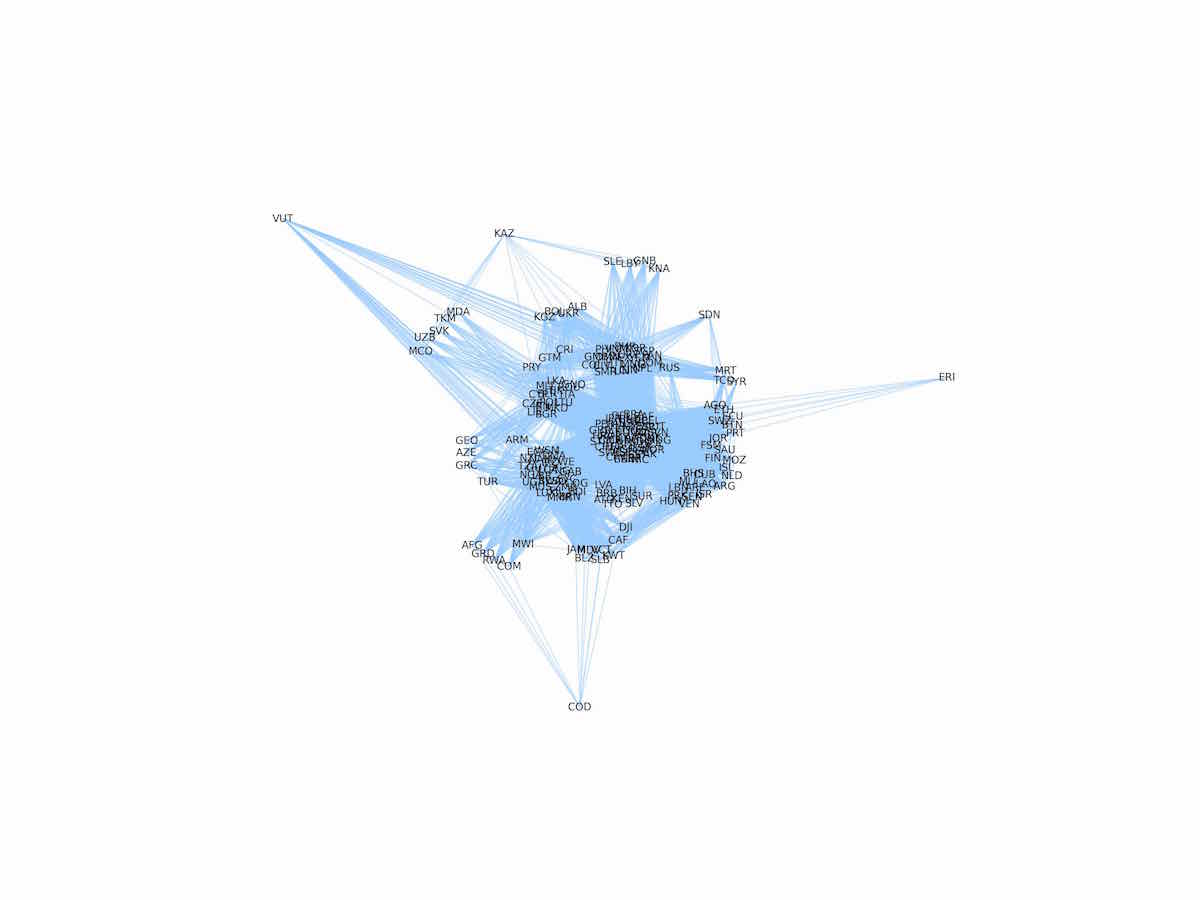}}
\subfloat[1999]{\includegraphics[width=.5\textwidth]{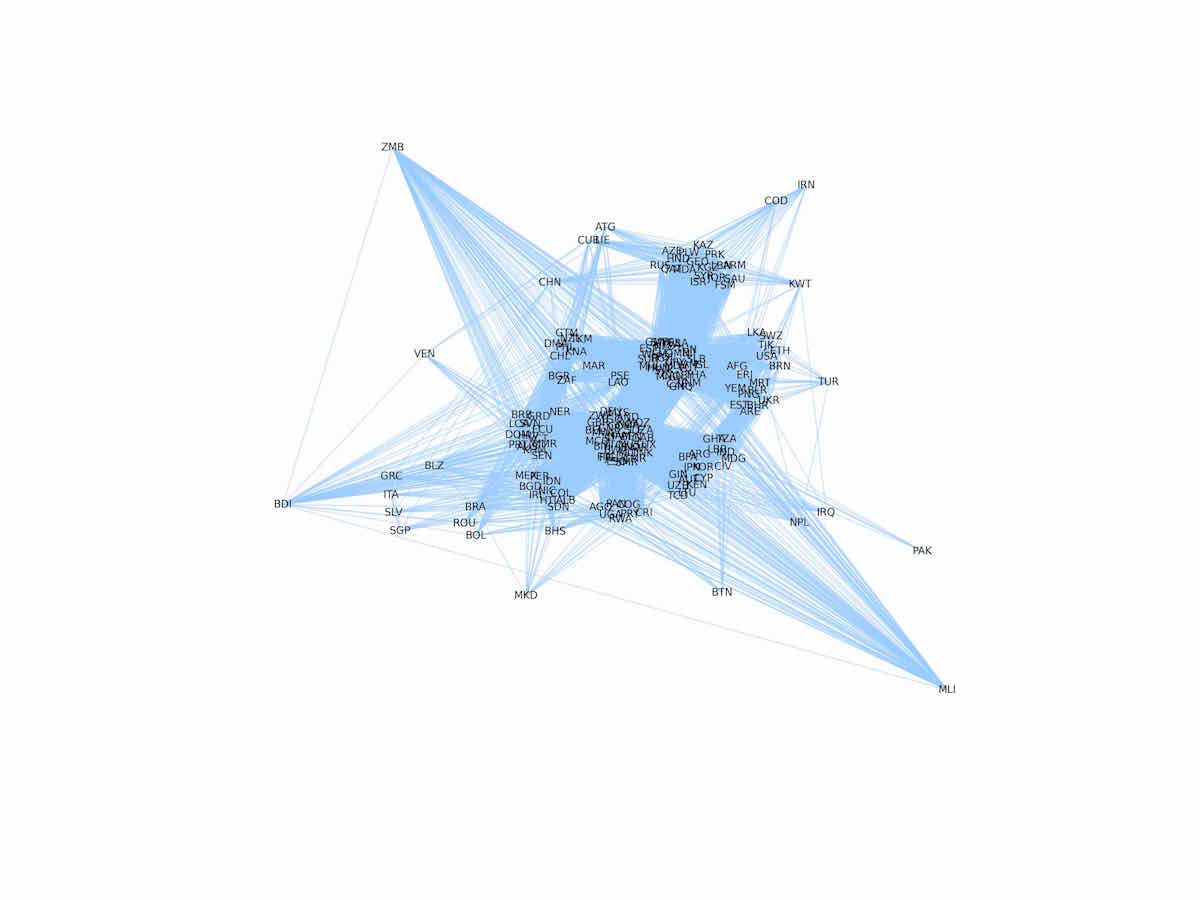}}

\caption{\emph{Networks from speeches, 1994-1999}. Networks built based on topics discussed in annual UNGA General Debate speeches. Links between countries is established using normalized mutual information criteria.
\label{fig:networks5}}

\end{figure}
%%%%%%%%%%%%%%%%%%%%%%%%%%%%%%%%%%%%%%%%%

%%%%%%%%%%%%%%%%%%%%%%%%%%%%%%%%%%%%%%%%%
%FIGURE: Networks based on MI measure for distance
\begin{figure}
\centering
\subfloat[2000]{\includegraphics[width=.5\textwidth]{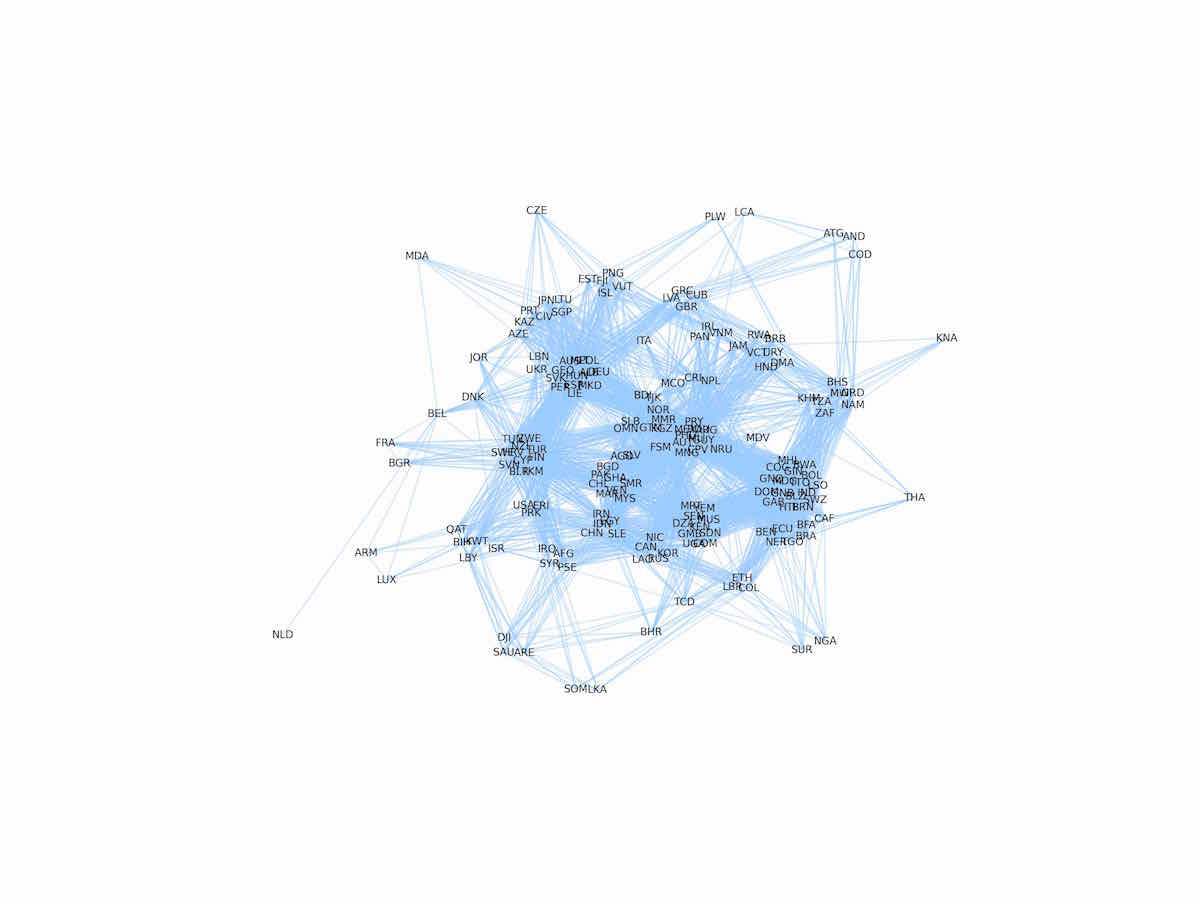}}
\subfloat[2001]{\includegraphics[width=.5\textwidth]{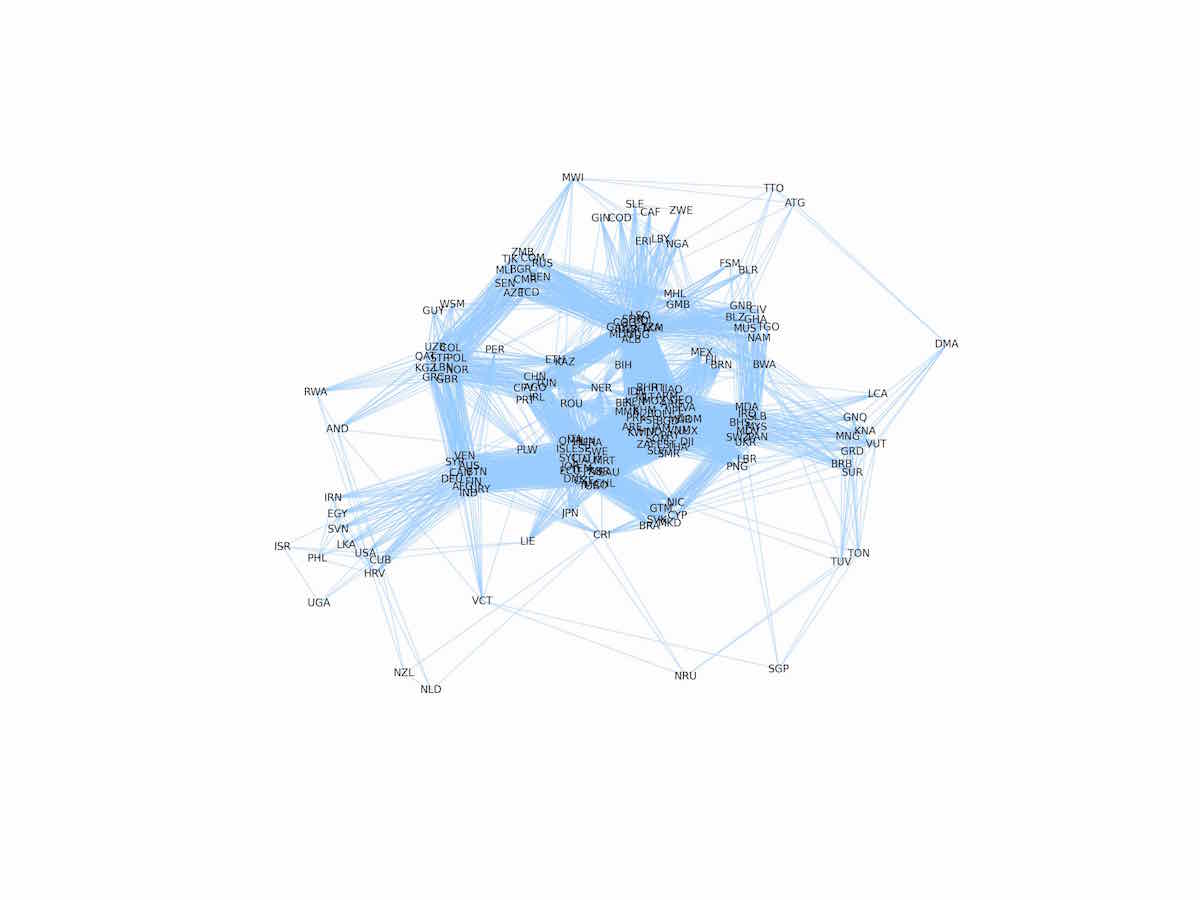}}\\
\subfloat[2002]{\includegraphics[width=.5\textwidth]{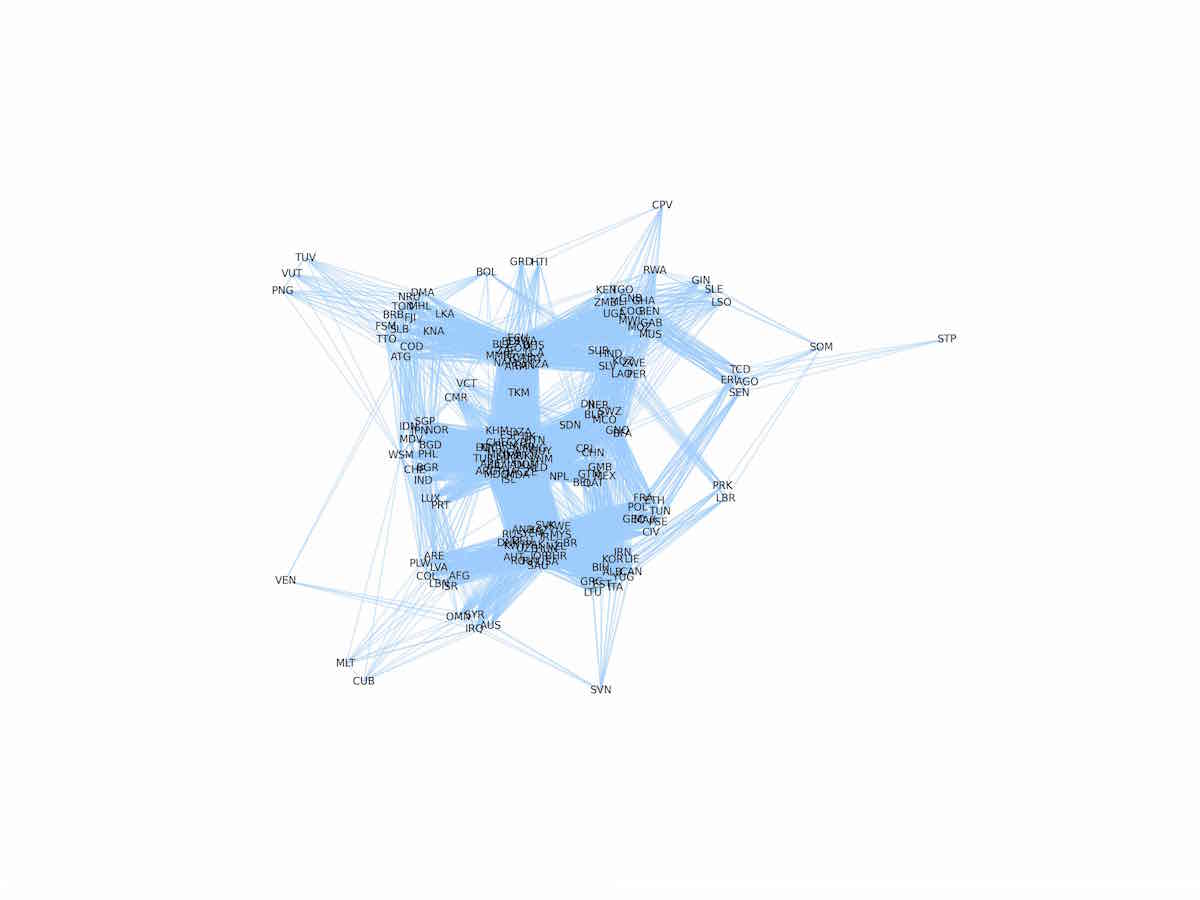}}
\subfloat[2003]{\includegraphics[width=.5\textwidth]{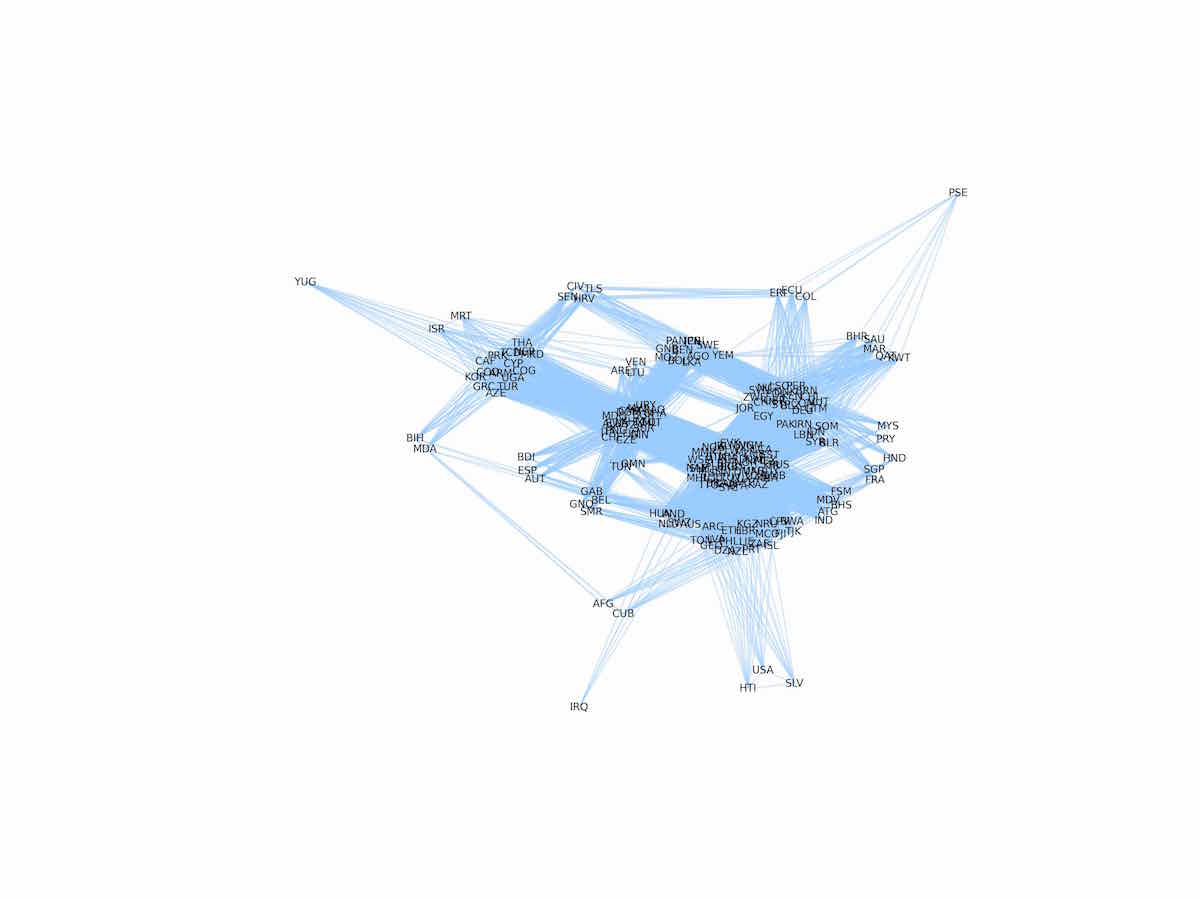}}\\
\subfloat[2004]{\includegraphics[width=.5\textwidth]{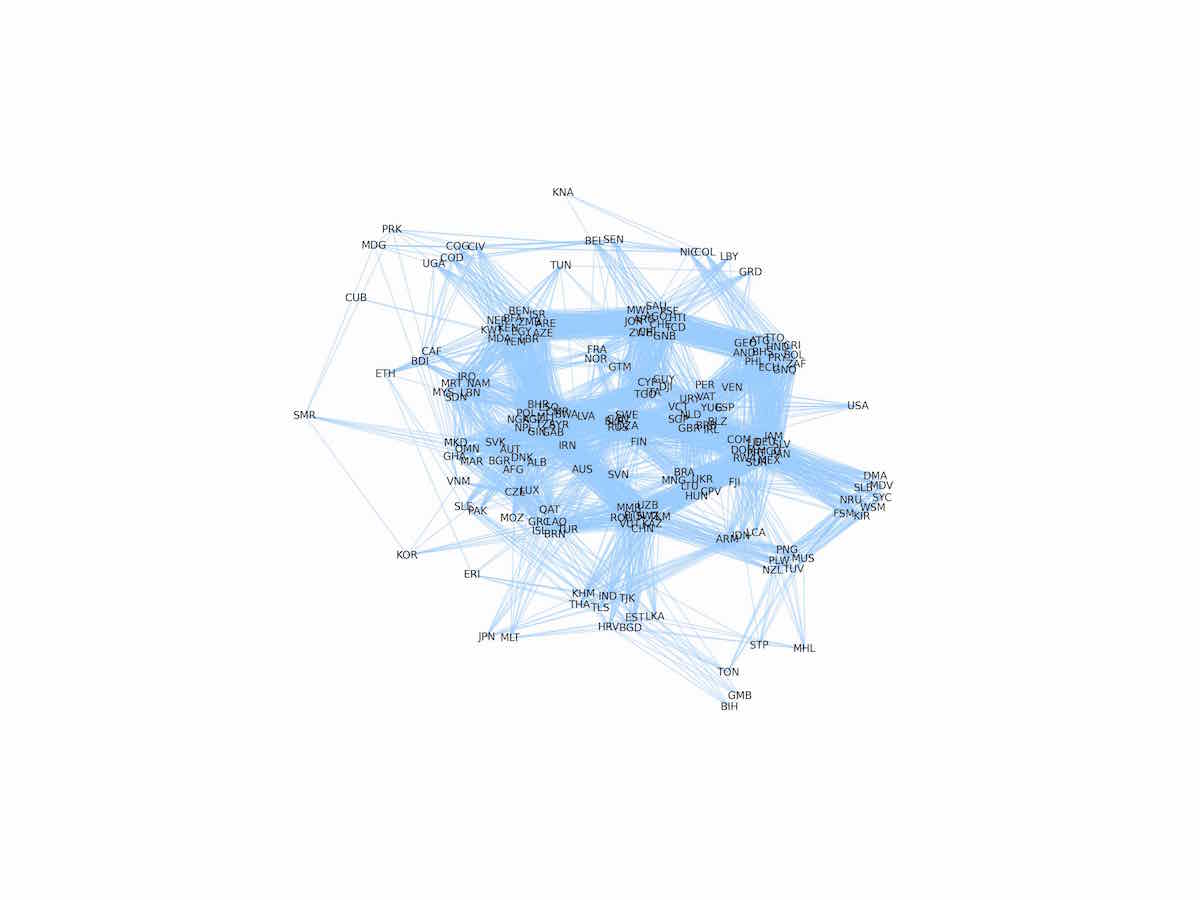}}
\subfloat[2005]{\includegraphics[width=.5\textwidth]{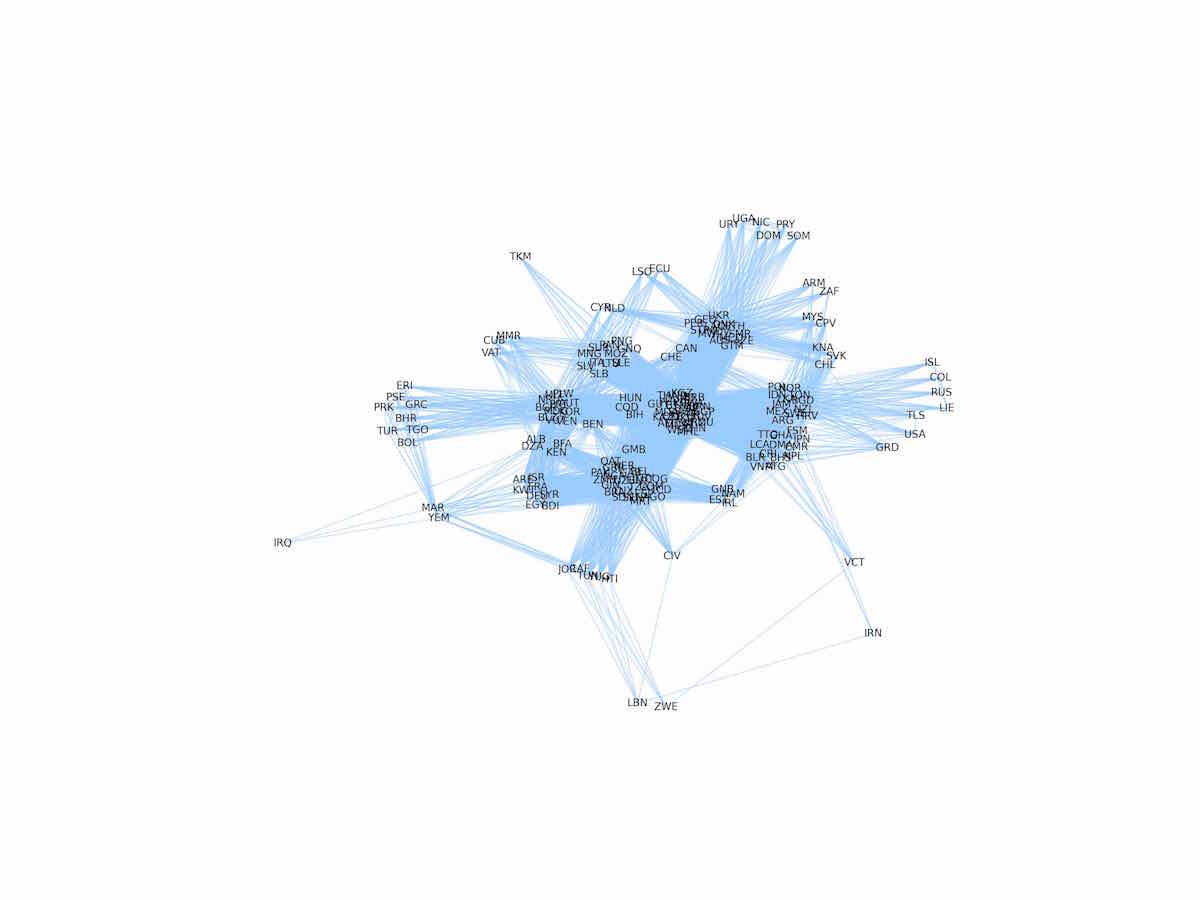}}

\caption{\emph{Networks from speeches, 2000-2005}. Networks built based on topics discussed in annual UNGA General Debate speeches. Links between countries is established using normalized mutual information criteria.
\label{fig:networks6}}

\end{figure}
%%%%%%%%%%%%%%%%%%%%%%%%%%%%%%%%%%%%%%%%%

%%%%%%%%%%%%%%%%%%%%%%%%%%%%%%%%%%%%%%%%%
%FIGURE: Networks based on MI measure for distance
\begin{figure}
\centering
\subfloat[2006]{\includegraphics[width=.5\textwidth]{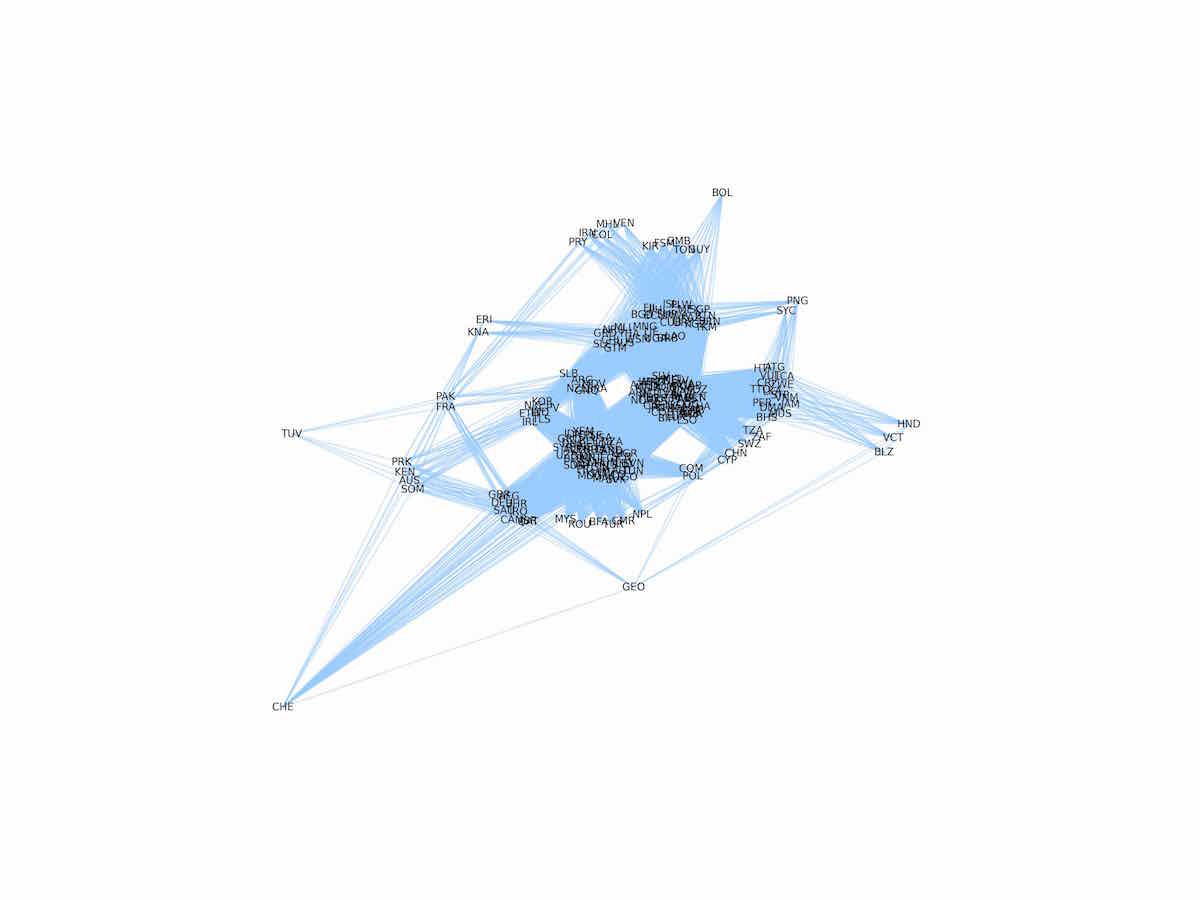}}
\subfloat[2007]{\includegraphics[width=.5\textwidth]{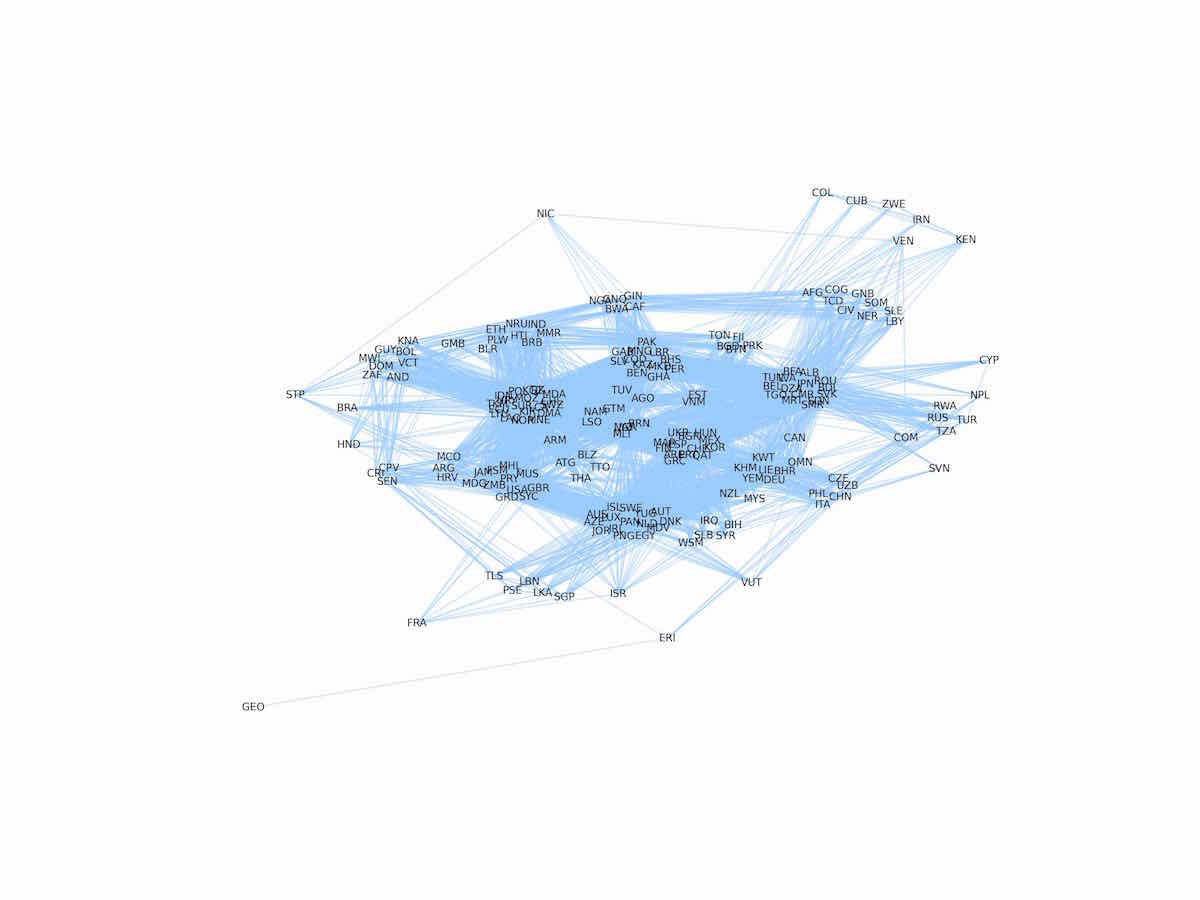}}\\
\subfloat[2008]{\includegraphics[width=.5\textwidth]{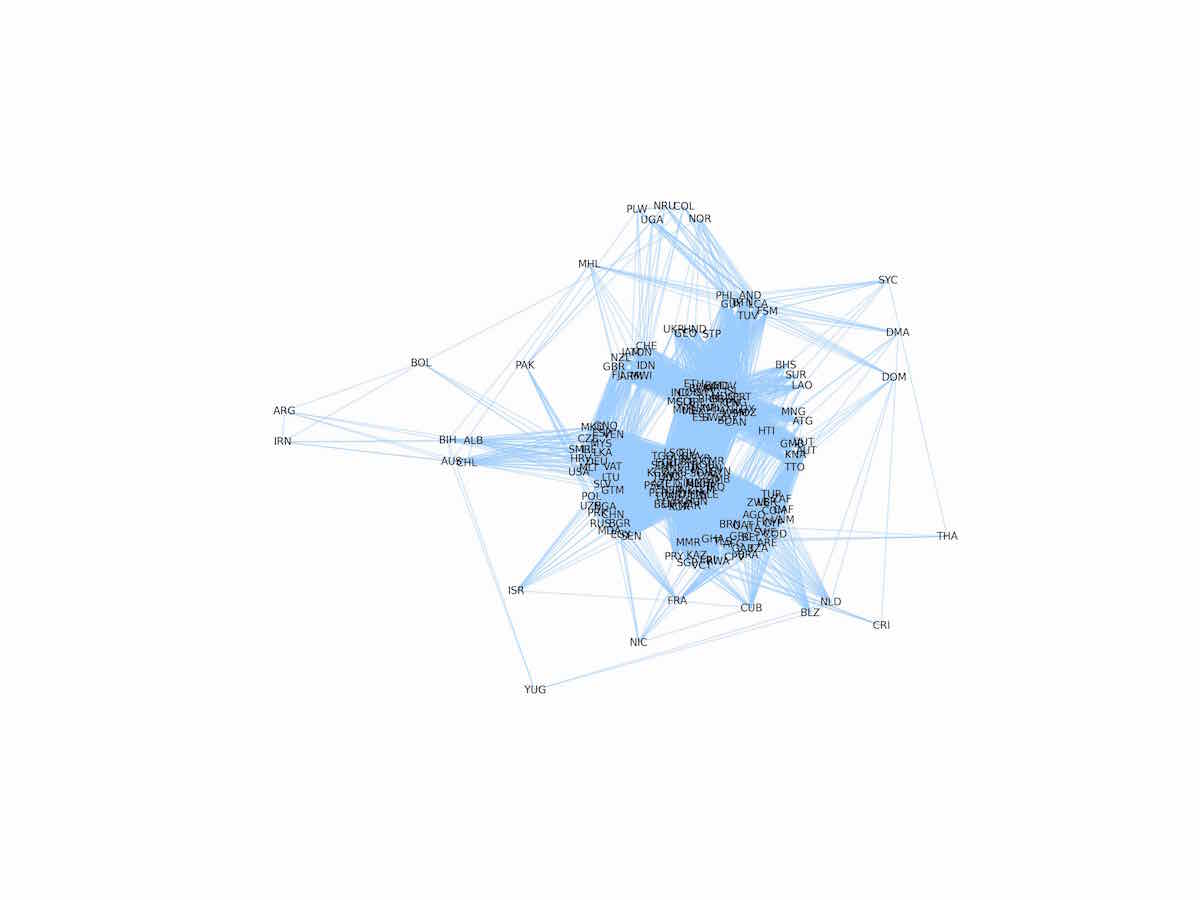}}
\subfloat[2009]{\includegraphics[width=.5\textwidth]{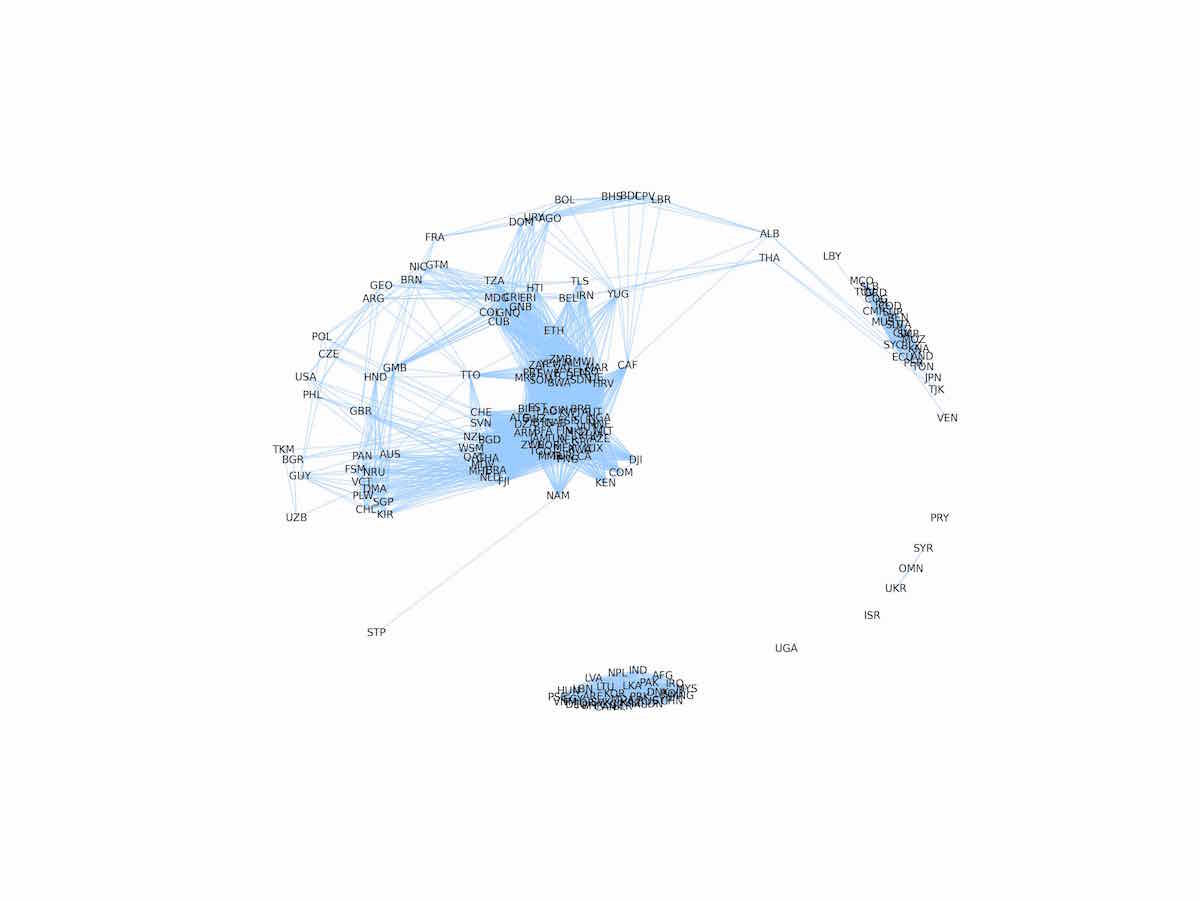}}\\
\subfloat[2010]{\includegraphics[width=.5\textwidth]{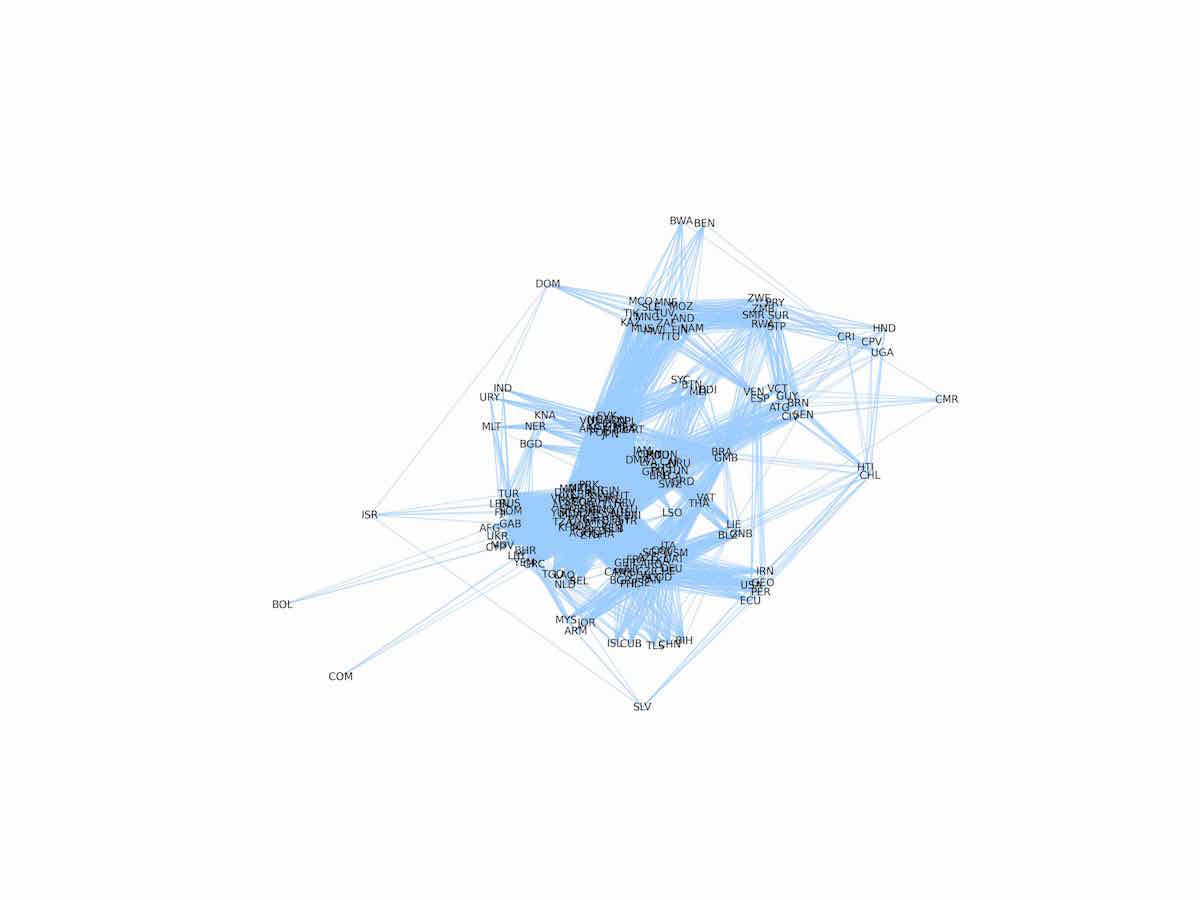}}
\subfloat[2011]{\includegraphics[width=.5\textwidth]{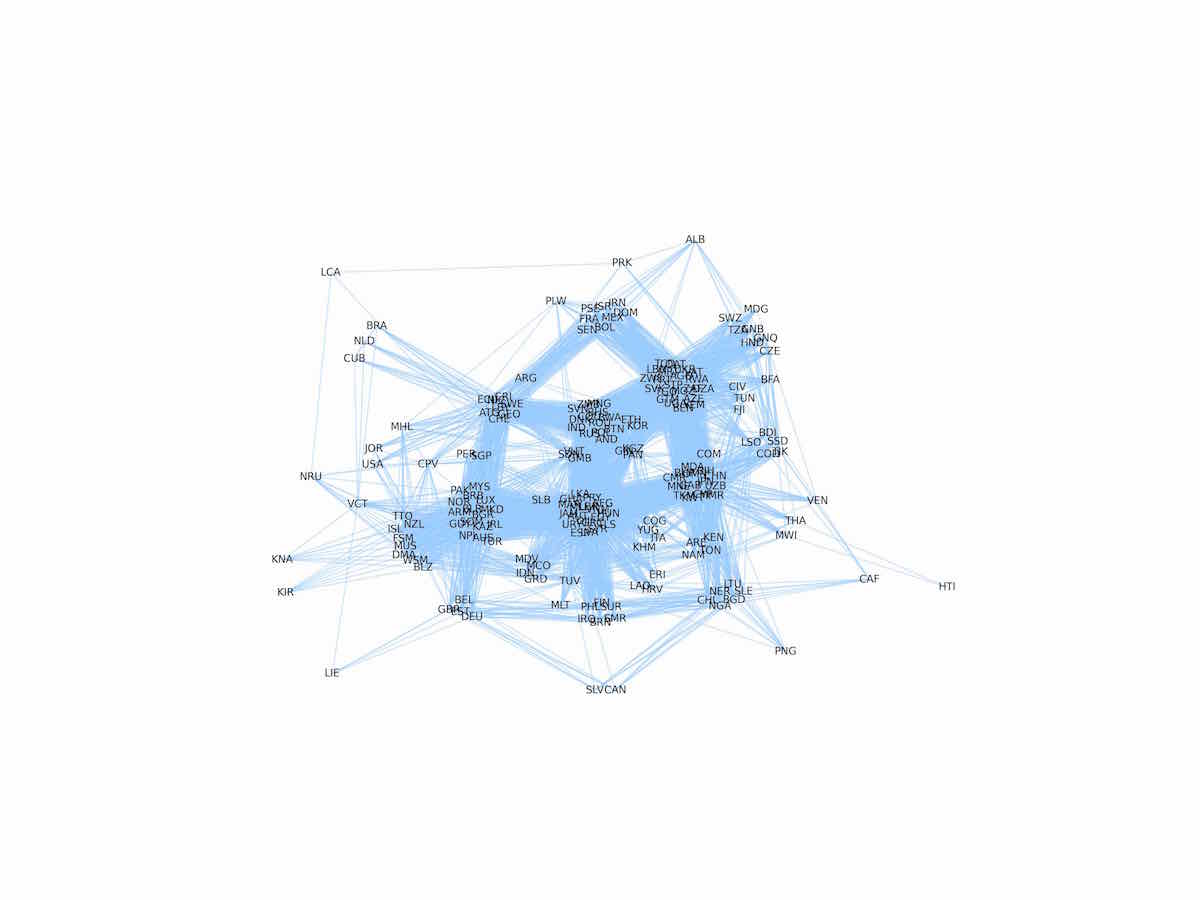}}

\caption{\emph{Networks from speeches, 2006-2011}. Networks built based on topics discussed in annual UNGA General Debate speeches. Links between countries is established using normalized mutual information criteria.
\label{fig:networks7}}

\end{figure}
%%%%%%%%%%%%%%%%%%%%%%%%%%%%%%%%%%%%%%%%%

%%%%%%%%%%%%%%%%%%%%%%%%%%%%%%%%%%%%%%%%%
%FIGURE: Networks based on MI measure for distance
\begin{figure}
\centering
\subfloat[2012]{\includegraphics[width=.5\textwidth]{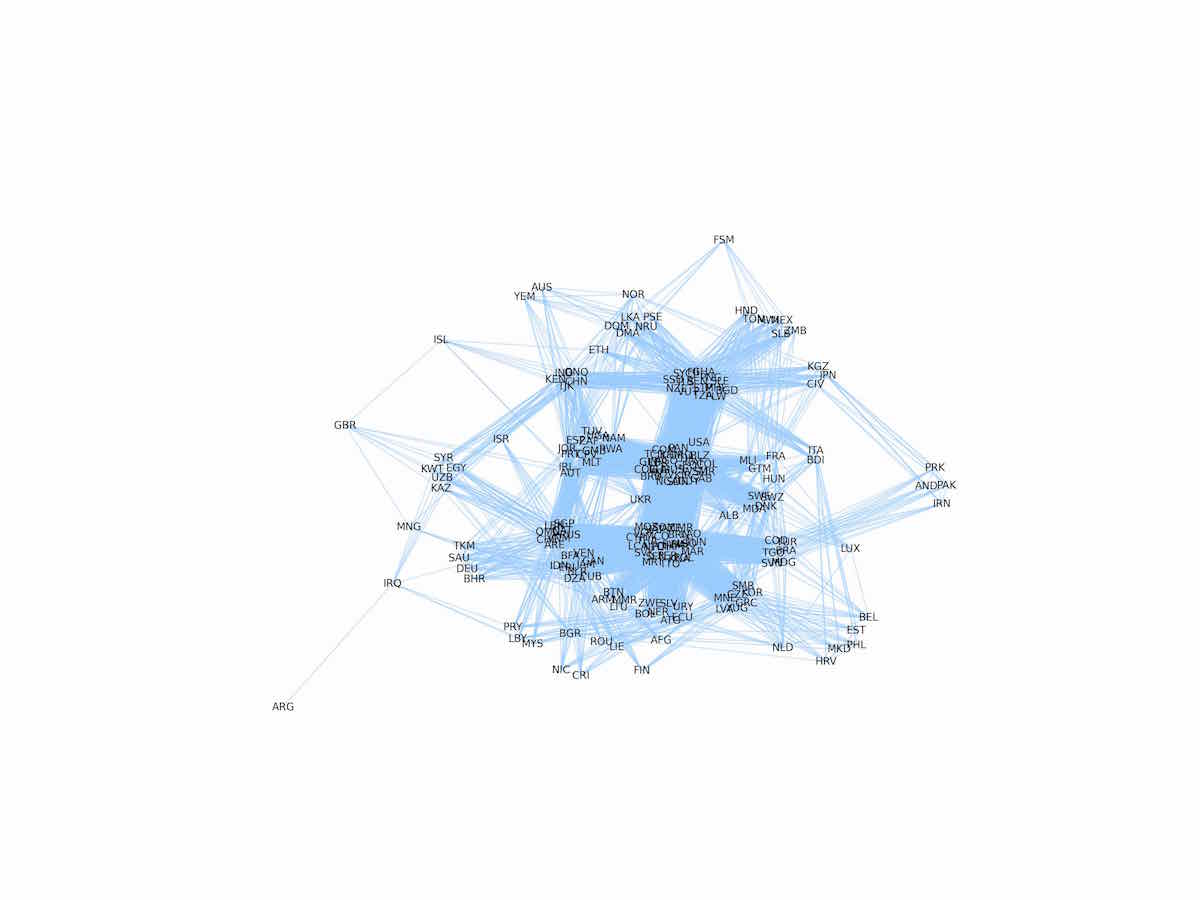}}
\subfloat[2013]{\includegraphics[width=.5\textwidth]{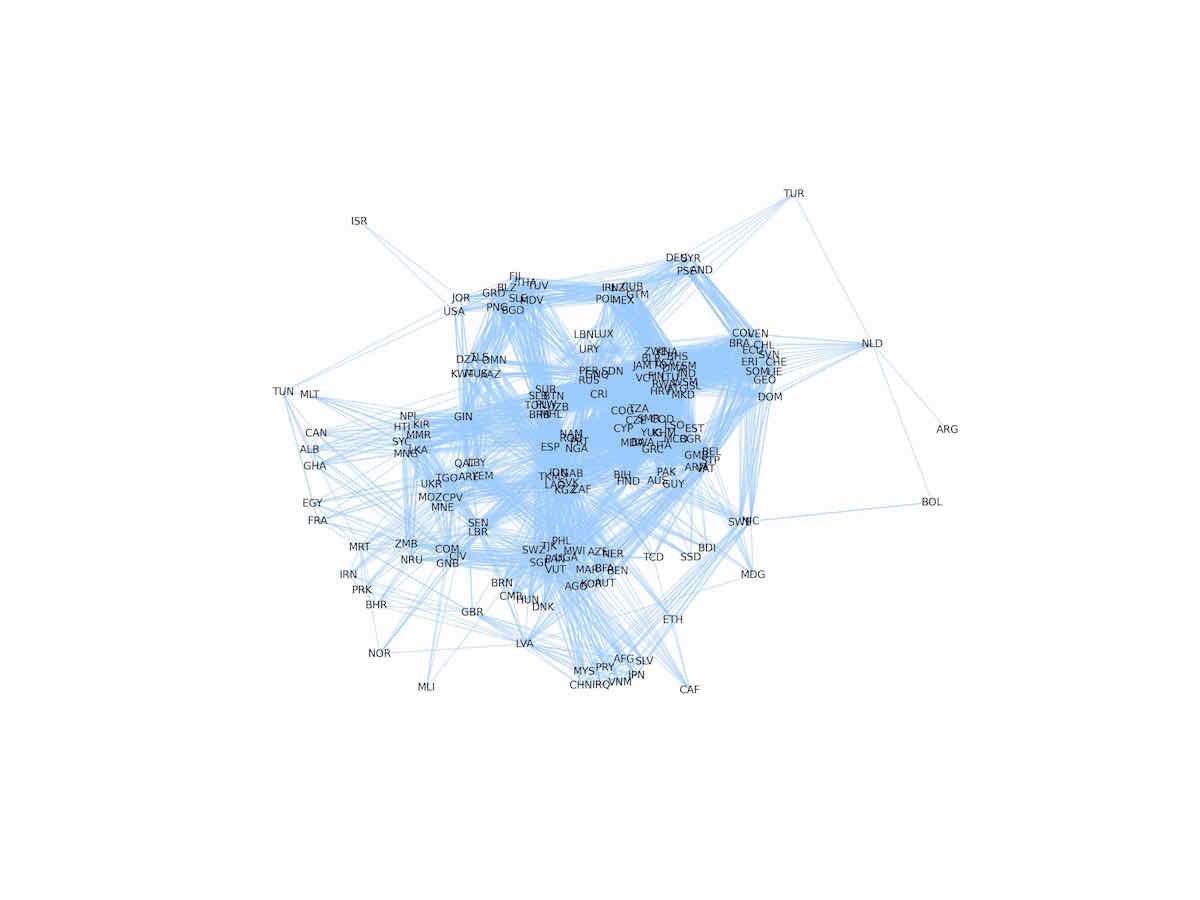}}\\
\subfloat[2014]{\includegraphics[width=.5\textwidth]{2014_net.jpg}}

\caption{\emph{Networks from speeches, 2012-2014}. Networks built based on topics discussed in annual UNGA General Debate speeches. Links between countries is established using normalized mutual information criteria.
\label{fig:networks8}}

\end{figure}
%%%%%%%%%%%%%%%%%%%%%%%%%%%%%%%%%%%%%%%%%

\subsection{Community detection}

%%%%%%%%%%%%%%%%%%%%%%%%%%%%%%%%%%%%%%%
%TABLE: Community detection
\begin{table}
\centering
\tiny
\begin{tabular}{c c p{14cm}}
\toprule
Year & Community & ISO Country Code \\
\midrule
1970	&	0	&	ALB, AUS, BLR, BRA, CMR, COG, CUB, DOM, DZA, ECU, FRA, GIN, GMB, IDN, IRQ, ISR, LBR, LBY, MMR, PRY, RWA, SDN, SOM, SYR, TTO, TUR, UKR, USA, YUG, ZAF	\\
1970	&	1	&	GHA, KWT, LBN, LKA, MAR, PAK, PHL, SLE, TGO, THA, TUN, ZMB	\\
1970	&	2	&	AUT, CAN, GTM, HND, HTI, IND, JPN, NOR, NZL, PER, URY	\\
1970	&	3	&	BEL, BOL, IRN, ITA, MDG, MEX, NLD, SLV, VEN	\\
1970	&	4	&	CRI, GBR, SGP	\\
1970	&	5	&	ARG, COL	\\
1970	&	6	&	ISL	\\
1970	&	7	&	KEN	\\
1970	&	8	&	KHM	\\
\hline
1971	&	0	&	AUT, BEL, CUB, CYP, GBR, GHA, GRC, HUN, IRL, ISR, ITA, LAO, LBR, LBY, MAR, MLT, MRT, RUS, SAU, SOM, SWE, SYR, TUR, UGA, USA, YDY, YEM	\\
1971	&	1	&	BEN, BFA, CHN, CIV, CMR, COD, COG, ETH, FJI, FRA, GAB, HTI, KEN, LUX, MDG, MLI, NER, NGA, NZL, RWA, SDN, SEN, SLE, TGO	\\
1971	&	2	&	AFG, BGR, BLR, CSK, FIN, IDN, IRN, IRQ, JOR, KHM, KWT, MMR, MNG, NOR, PAK, POL, QAT, ROU, THA, UKR, YUG	\\
1971	&	3	&	ARG, AUS, BOL, CAF, CAN, CHL, COL, ESP, GTM, JAM, LBN, MEX, MUS, NIC, NLD, PAN, PER, TTO, TUN, URY, VEN	\\
1971	&	4	&	BRA, GUY, IND, ISL, JPN, LKA, MYS, NPL, PHL	\\
1971	&	5	&	BDI, DZA, SGP, TCD, TZA, ZMB	\\
1971	&	6	&	DOM, ECU, PRY, SLV, ZAF	\\
1971	&	7	&	ALB, GIN	\\
1971	&	8	&	EGY	\\
\hline
1972	&	0	&	ARE, BGR, BHR, CHN, COL, CSK, CUB, DOM, DZA, FIN, GBR, GIN, IRQ, LAO, LBN, MAR, MLI, MNG, MRT, MUS, NIC, PAK, POL, PRT, QAT, ROU, RUS, SLE, SOM, SWE, SYR, TZA, UKR, YEM, YUG, ZMB	\\
1972	&	1	&	AUS, BEL, BEN, BRB, CAN, CHL, COG, CRI, FRA, GAB, GRC, LUX, MDG, MMR, NZL, PER, RWA, SAU, SDN, SEN, SWZ, TCD, TUN, TUR, UGA, VEN	\\
1972	&	2	&	ARG, AUT, BFA, BOL, BRA, BTN, CIV, CMR, ECU, ESP, HND, HTI, IND, ITA, JPN, MEX, MWI, NER, NLD, NPL, SGP, SLV, TGO, URY, ZAF	\\
1972	&	3	&	CYP, FJI, GUY, IDN, IRN, LBR, MLT, MYS, OMN, PHL, THA	\\
1972	&	4	&	AFG, BDI, CAF, EGY, ETH, GMB, JAM, KEN, USA	\\
1972	&	5	&	DNK, GHA, GTM, LKA, NGA, PRY	\\
1972	&	6	&	ISL, ISR, JOR	\\
1972	&	7	&	ALB, BLR	\\
1972	&	8	&	KHM, KWT, LBY, YDY	\\
1972	&	9	&	COD	\\
1972	&	10	&	HUN	\\
1972	&	11	&	IRL	\\
\hline
1973	&	0	&	AFG, AUS, BEL, CAN, CYP, DNK, FIN, FJI, GUY, IDN, IND, ISL, ITA, LUX, MMR, MUS, NLD, NOR, NZL, PHL, ROU, SEN, THA, UKR	\\
1973	&	1	&	ARG, BEN, BOL, CAF, CHL, COL, ESP, GTM, HND, HTI, HUN, LBN, MAR, MDG, MEX, MRT, PER, PRY, RWA, TUR	\\
1973	&	2	&	ARE, BDI, BHR, DZA, EGY, GHA, IRN, IRQ, KWT, LBY, MLI, QAT, SAU, SDN, SOM, SYR, YDY, YEM	\\
1973	&	3	&	BFA, BTN, CHN, COD, COG, CSK, DDR, GIN, LAO, MYS, NER, OMN, PRT, YUG	\\
1973	&	4	&	AUT, BLR, BRA, DEU, GBR, GRC, MNG, POL, RUS, USA	\\
1973	&	5	&	ETH, LKA, NGA, NPL, SWE, TZA, ZMB	\\
1973	&	6	&	CRI, ECU, NIC, PAN, URY, VEN	\\
1973	&	7	&	GAB, JAM, JPN, KEN, LBR, LSO, PAK, SLE, ZAF	\\
1973	&	8	&	ISR, SGP, SLV, TUN	\\
1973	&	9	&	ALB, BGR	\\
1973	&	10	&	CUB, KHM	\\
1973	&	11	&	BHS, FRA	\\
1973	&	12	&	IRL	\\
1973	&	13	&	JOR	\\
\hline
1974	&	0	&	AFG, AUT, BDI, BEL, BEN, BFA, BGD, BLR, BRA, BTN, CAF, CHL, CHN, CIV, CMR, COD, COG, CSK, CUB, DDR, DEU, DNK, DZA, EGY, ESP, ETH, FIN, FRA, GAB, GHA, GIN, GNQ, GRC, GTM, HND, IND, IRL, IRN, IRQ, ISL, ISR, ITA, JOR, JPN, KHM, LAO, LBN, LBR, LBY, LKA, LUX, MEX, MLI, MLT, MRT, MUS, MYS, NER, NLD, NOR, NPL, PAK, PER, POL, PRT, PRY, ROU, RUS, RWA, SEN, SGP, SLE, SLV, SOM, SWE, SYR, TGO, THA, TUN, TUR, UGA, VEN, YDY, YEM, YUG, ZMB	\\
1974	&	1	&	ARG, AUS, CAN, COL, DOM, ECU, GBR, GRD, NIC, NZL, PHL, URY, USA	\\
1974	&	2	&	ARE, BHR, IDN, KEN, KWT, OMN, QAT, SAU, SWZ, TCD	\\
1974	&	3	&	BOL, BRB, BWA, GMB, JAM, LSO, MMR, NGA, SDN, TZA	\\
1974	&	4	&	ALB, BGR, HUN, MNG, UKR	\\
1974	&	5	&	CRI, MAR	\\
1974	&	6	&	CYP	\\
1974	&	7	&	GUY	\\
1974	&	8	&	HTI	\\
\bottomrule

\end{tabular}
\caption{Infomapped communities 1970-1974 \label{tab:communities1}}

\end{table}
%%%%%%%%%%%%%%%%%%%%%%%%%%%%%%%%%%%%%%%% 

\begin{table}
\centering
\tiny
\begin{tabular}{c c p{14cm}}
\toprule
Year & Community & ISO Country Code \\
\midrule
1975	&	0	&	AFG, ALB, ARG, AUS, AUT, BEL, BFA, BGD, BGR, BHR, CAN, CHL, CHN, CMR, CRI, CUB, DDR, DEU, DNK, DOM, DZA, ECU, ETH, FJI, FRA, GBR, GIN, GRC, GUY, HUN, IDN, IND, IRL, IRN, IRQ, ISR, ITA, JPN, KWT, LAO, LBR, LBY, LKA, MAR, MDG, MMR, MUS, MYS, NER, NLD, NOR, NPL, NZL, PAK, PER, POL, PRY, RWA, SGP, SOM, SWE, THA, TUR, VEN, YDY, YUG	\\
1975	&	1	&	ARE, BEN, COL, GAB, GHA, GNQ, JAM, KEN, MLI, MOZ, MRT, MWI, SAU, SDN, SYR, TCD, TGO, YEM, ZMB	\\
1975	&	2	&	BDI, BOL, BRA, BRB, BTN, COD, HND, LSO, MEX, NGA, OMN, PHL, PRT, SEN, SLV, TUN, TZA, URY	\\
1975	&	3	&	BLR, CSK, MNG, ROU, RUS, UKR, USA	\\
1975	&	4	&	CAF, FIN, ISL, NIC, PAN	\\
1975	&	5	&	COG, JOR, KHM, QAT, SLE, UGA	\\
1975	&	6	&	CYP, ESP	\\
1975	&	7	&	GRD, GTM	\\
1975	&	8	&	BWA	\\
\hline
1976	&	0	&	AFG, ARE, BDI, BEN, BFA, BHR, BWA, CAF, CMR, COD, COG, COM, CRI, CUB, DDR, DZA, ECU, EGY, FRA, GAB, GHA, GIN, GUY, HUN, IDN, KEN, KWT, LBR, LBY, LKA, LSO, MAR, MDG, MDV, MLI, MMR, MOZ, MRT, MUS, MYS, NER, PER, PRT, QAT, RWA, SAU, SDN, SEN, SLE, SOM, STP, SUR, SWE, SWZ, SYR, TCD, TGO, TUN, TZA, UGA, YEM, YUG, ZMB	\\
1976	&	1	&	BTN, CAN, DNK, GBR, GTM, IRL, ISL, MLT, NIC, NZL, OMN, PAK, PHL, PRY	\\
1976	&	2	&	BGD, BRB, ETH, GRD, IND, JAM, JOR, LBN, NGA, PAN, PNG, SGP, TUR	\\
1976	&	3	&	ARG, AUS, AUT, BEL, BRA, CYP, FIN, GRC, IRN, ISR, ITA, JPN, NLD, NOR, NPL, SLV	\\
1976	&	4	&	BOL, CHL, COL, DOM, FJI, HND, MEX, URY, VEN	\\
1976	&	5	&	CPV, GNB, KHM, LAO, YDY	\\
1976	&	6	&	DEU, POL, ROU, RUS, UKR, USA	\\
1976	&	7	&	BGR, BLR, CSK, MNG	\\
1976	&	8	&	ALB, CHN, IRQ	\\
1976	&	9	&	ESP	\\
\hline
1977	&	0	&	AUS, BGD, BOL, COL, CRI, DNK, FIN, IDN, IND, IRN, ISR, ITA, JAM, JPN, LSO, MEX, MMR, MUS, MYS, NLD, NOR, NZL, PAK, PHL, POL, PRY, SUR, SWE, THA, URY, VEN	\\
1977	&	1	&	ARE, BDI, CMR, COD, COM, ETH, GHA, GIN, GMB, IRQ, KWT, LBY, MDG, MDV, MLI, MRT, MWI, NGA, NPL, SDN, SEN, SOM, STP, SWZ, SYC, TUN, TZA, UGA, WSM, YDY, YEM, YUG	\\
1977	&	2	&	AGO, ALB, BEN, CAF, COG, CPV, DZA, EGY, GNB, GRC, JOR, LBN, MAR, MOZ, OMN, QAT, RWA, SAU, SYR, TCD	\\
1977	&	3	&	ARG, CHL, ECU, GTM, HND, ISL, NIC, PAN, PER, SLV, USA	\\
1977	&	4	&	AFG, BRB, BWA, FJI, KEN, LBR, SLE, TGO, TTO, ZMB	\\
1977	&	5	&	AUT, BGR, BLR, CSK, CUB, DDR, GRD, HUN, LAO, MNG, ROU, RUS, UKR	\\
1977	&	6	&	BEL, BHR, BRA, CAN, ESP, FRA, HTI, LKA, LUX, PRT	\\
1977	&	7	&	BFA, BTN, CIV, CYP, GBR, IRL, NER, TUR	\\
1977	&	8	&	DEU, SGP	\\
1977	&	9	&	CHN, KHM	\\
1977	&	10	&	PNG	\\
\hline
1978	&	0	&	BDI, BFA, BGD, BHR, BOL, BRA, CAF, CAN, CHL, CIV, CMR, COM, CPV, DNK, DZA, FIN, FJI, FRA, GAB, GNB, GNQ, GRC, GRD, GTM, GUY, HTI, IDN, IND, IRL, ISL, ITA, JOR, KEN, KWT, LBR, LKA, LUX, MAR, MDG, MDV, MLI, MRT, NLD, NOR, NPL, POL, RWA, SEN, SUR, SWE, TCD, TGO, TUN, TUR, URY, USA, YUG	\\
1978	&	1	&	ARE, AUS, BRB, BTN, BWA, CYP, DEU, GBR, GMB, JAM, JPN, MMR, MYS, NGA, NZL, SLE, SYC, TTO, TZA, WSM	\\
1978	&	2	&	ARG, AUT, BEL, COL, CRI, DOM, ECU, ESP, MEX, MLT, NIC, PAN, PHL, PRT, SLV, VEN	\\
1978	&	3	&	AGO, ALB, BGR, BLR, CHN, COD, COG, CSK, CUB, DDR, HND, HUN, ISR, MNG, NER, ROU, RUS, STP, UKR	\\
1978	&	4	&	AFG, EGY, ETH, IRN, KHM, LBY, MOZ, OMN, PAK, QAT, SDN, SOM, SYR, THA, UGA, VNM, YDY, YEM, ZMB	\\
1978	&	5	&	BEN, LAO	\\
1978	&	6	&	IRQ, SGP	\\
1978	&	7	&	GHA, MUS, PNG	\\
1978	&	8	&	LBN, PER	\\
1978	&	9	&	LSO	\\
\hline
1979	&	0	&	AFG, ARE, ARG, AUT, BFA, BHR, BRA, BRB, BWA, CAN, CHL, CIV, CMR, COD, COG, COL, COM, CPV, CRI, CYP, ECU, EGY, ESP, ETH, FJI, FRA, GHA, GRC, GRD, HUN, IRQ, ISR, JOR, JPN, KEN, KWT, LBN, LBR, LKA, LSO, LUX, MDG, MDV, MEX, MLT, MRT, MUS, MWI, NER, NGA, NPL, OMN, PNG, PRT, QAT, SDN, SLE, SLV, SOM, STP, SYR, THA, TUN, TUR, TZA, URY, USA, YUG	\\
1979	&	1	&	BOL, DJI, GNB, GNQ, LBY, MAR, SEN, TCD, UGA, YEM, ZMB	\\
1979	&	2	&	BGD, BTN, FIN, IND, IRL, JAM, MMR, MYS, SUR, WSM	\\
1979	&	3	&	BGR, BLR, CHN, CSK, DDR, IDN, MNG, NOR, POL, ROU, RUS, UKR	\\
1979	&	4	&	ALB, BEL, DEU, DOM, GBR, GUY, IRN, ITA, PAK, PER, VEN	\\
1979	&	5	&	BEN, GAB, GIN, LAO, MOZ, NIC, SAU, SGP, SYC, YDY	\\
1979	&	6	&	BDI, CAF, CUB, GTM, MLI, PAN, RWA, TGO	\\
1979	&	7	&	DMA, ISL, NZL, PHL, TTO	\\
1979	&	8	&	AUS, NLD	\\
1979	&	9	&	DZA, PRY	\\
1979	&	10	&	KHM, VNM	\\
1979	&	11	&	HND, HTI	\\
1979	&	12	&	VAT	\\
\bottomrule

\end{tabular}
\caption{Infomapped communities 1975-1979 \label{tab:communities2}}

\end{table}
%%%%%%%%%%%%%%%%%%%%%%%%%%%%%%%%%%%%%%%% 

\begin{table}
\centering
\tiny
\begin{tabular}{c c p{14cm}}
\toprule
Year & Community & ISO Country Code \\
\midrule
1980	&	0	&	AFG, AGO, ALB, ARG, AUT, BEL, BEN, BFA, BGD, BHR, BHS, BLR, BOL, BRA, BRB, BTN, BWA, CAN, CHL, CHN, CMR, COD, COL, COM, CPV, CRI, CSK, CUB, CYP, ECU, EGY, ESP, ETH, FJI, FRA, GBR, GHA, GNQ, GRC, GRD, GTM, GUY, HND, HTI, IDN, IND, IRL, ITA, JAM, JOR, KEN, KHM, LAO, LBR, LBY, LCA, LKA, LSO, LUX, MAR, MDG, MDV, MEX, MLI, MLT, MMR, MNG, MOZ, MUS, MWI, MYS, NGA, NIC, NPL, NZL, OMN, PAK, PAN, PER, PHL, PNG, PRT, PRY, QAT, RUS, RWA, SAU, SDN, SEN, SLV, SOM, SUR, SYC, SYR, THA, TUN, TUR, TZA, URY, VCT, VEN, VNM, YDY, YUG, ZMB	\\
1980	&	1	&	AUS, BGR, DDR, DEU, DNK, FIN, HUN, ISL, ISR, JPN, NLD, NOR, POL, ROU, SWE, UKR, USA	\\
1980	&	2	&	ARE, CAF, COG, DJI, DZA, GAB, GMB, GNB, MRT, STP, SWZ, TTO, UGA, YEM, ZWE	\\
1980	&	3	&	CIV, GIN, LBN, NER, SLE, TCD, TGO	\\
1980	&	4	&	BDI, IRQ	\\
1980	&	5	&	DOM	\\
1980	&	6	&	IRN	\\
1980	&	7	&	SGP	\\
\hline
1981	&	0	&	ARG, AUS, BDI, BFA, BOL, BRB, CAF, CHL, CIV, CMR, COD, COG, COL, COM, CPV, CRI, CUB, DEU, DOM, DZA, FRA, GAB, GBR, GIN, GNB, GNQ, GRD, GTM, GUY, HND, HTI, IND, IRL, ISR, ITA, JAM, KEN, MLI, MRT, MUS, NER, NLD, PAN, PER, PRY, ROU, RWA, SEN, SLE, SLV, STP, TCD, TGO, URY, USA, VCT, VEN	\\
1981	&	1	&	AGO, ARE, BHR, BWA, CYP, ECU, EGY, ESP, ETH, FJI, JOR, KHM, KWT, LBR, MDV, MYS, NGA, OMN, PRT, QAT, SAU, SDN, SGP, SOM, SUR, THA, TZA, UGA, WSM, ZMB	\\
1981	&	2	&	AFG, AUT, BEL, BGR, BLR, CSK, DDR, DJI, DNK, FIN, HUN, IDN, ISL, JPN, LAO, MDG, MEX, MLT, MNG, PAK, PHL, PNG, POL, RUS, SWE, TUN, UKR, VNM, YUG	\\
1981	&	3	&	BGD, BHS, BRA, BTN, CAN, GHA, GRC, LCA, LKA, LUX, NPL, NZL, TUR	\\
1981	&	4	&	BEN, CHN, MOZ, YDY	\\
1981	&	5	&	IRN, LBY, MAR, NIC, SYC, SYR, YEM, ZWE	\\
1981	&	6	&	ALB	\\
1981	&	7	&	DMA	\\
1981	&	8	&	IRQ	\\
1981	&	9	&	LBN	\\
\hline
1982	&	0	&	AUS, AUT, BGD, BTN, COD, CYP, DNK, EGY, FIN, FJI, GBR, GRC, IDN, IND, IRL, ITA, JPN, LKA, LSO, LUX, MDV, MMR, NLD, NOR, NPL, NZL, PAK, PHL, PNG, POL, SWE, TUR, TZA, YUG	\\
1982	&	1	&	AGO, BEL, BLZ, BOL, CHL, COG, COL, CRI, DZA, GAB, GIN, GNB, HND, MDG, PER, PRY, SGP, SUR, TGO, TUN, USA, VEN	\\
1982	&	2	&	AFG, BEN, BHR, DDR, DJI, ETH, GRD, GUY, IRN, JAM, LBY, MAR, MOZ, NGA, SLE, SOM, STP, VCT, VNM, YDY, ZWE	\\
1982	&	3	&	BDI, BFA, BHS, COM, CPV, HTI, IRQ, KHM, KWT, MLI, MRT, MUS, NER, SEN, TCD	\\
1982	&	4	&	ARE, ATG, BRB, CAN, GMB, ISR, KEN, LBR, RWA, SDN, TTO, UGA, ZMB	\\
1982	&	5	&	ALB, BGR, BLR, CSK, CUB, HUN, LAO, MNG, NIC, RUS, UKR	\\
1982	&	6	&	BRA, ESP, FRA, MEX, SLV	\\
1982	&	7	&	ARG, ECU, MLT, PAN	\\
1982	&	8	&	BWA, JOR, QAT, THA	\\
1982	&	9	&	CAF, DEU, DOM, GNQ, GTM, PRT, URY	\\
1982	&	10	&	ISL, ROU	\\
1982	&	11	&	CHN, YEM	\\
1982	&	12	&	CMR, GHA, OMN	\\
1982	&	13	&	SAU, SYR	\\
1982	&	14	&	MYS	\\
\hline
1983	&	0	&	AFG, AGO, ALB, BLR, BWA, CHN, COG, CSK, CUB, DDR, GHA, GNB, GRD, IRL, IRN, KHM, LAO, LBN, LBY, LSO, MLT, MOZ, MUS, NGA, NIC, NOR, POL, PRT, ROU, RUS, STP, SWE, SYC, SYR, UKR, URY, USA, YDY, ZWE	\\
1983	&	1	&	ARE, ATG, AUT, BLZ, BRA, BRB, CAF, CRI, CYP, ESP, GNQ, GRC, GTM, GUY, ISL, ITA, JAM, JPN, KWT, MDV, MEX, MMR, NER, NPL, RWA, SLB, SUR, THA, TUR, UGA, VEN, ZMB	\\
1983	&	2	&	BDI, BEN, BGD, BHR, BHS, CHL, CMR, COD, DEU, DJI, DNK, FIN, FJI, GBR, GIN, ISR, JOR, LUX, MDG, NLD, OMN, PAK, PER, PRY, QAT, SAU, SEN, TUN, TZA, WSM, YEM	\\
1983	&	3	&	COM, ETH, KEN, LBR, LKA, MAR, MRT, MYS, NZL, PNG, SGP, TTO, VCT, VUT	\\
1983	&	4	&	AUS, BEL, BOL, DZA, ECU, EGY, FRA, GAB, IND, MLI, SDN, SOM, YUG	\\
1983	&	5	&	BFA, BTN, CPV, IRQ, SLE, TCD	\\
1983	&	6	&	BGR, HND, HUN, MNG, SLV	\\
1983	&	7	&	ARG, DOM	\\
1983	&	8	&	HTI, TGO	\\
1983	&	9	&	CAN, IDN, PHL	\\
1983	&	10	&	COL	\\
1983	&	11	&	VNM	\\
\hline
1984	&	0	&	AGO, ARG, ATG, AUS, AUT, BDI, BEN, BFA, BGR, BHS, BLZ, BRB, BTN, BWA, CAF, CHN, CMR, COG, COM, CPV, CUB, DDR, DMA, DNK, DOM, DZA, ECU, ESP, ETH, FIN, FJI, FRA, GAB, GNB, GNQ, GRC, GUY, HND, HTI, HUN, IDN, IND, IRL, ISR, ITA, JAM, JPN, LBR, LCA, LSO, MAR, MDG, MDV, MLI, MLT, MMR, MUS, MYS, NER, NGA, NIC, NLD, NPL, NZL, PHL, PNG, PRT, RWA, SEN, SGP, SLB, SLE, SLV, STP, SUR, SWE, TCD, TGO, TTO, TUN, TUR, TZA, VUT, YDY, YUG, ZAF	\\
1984	&	1	&	BGD, BHR, CYP, DJI, EGY, GHA, GIN, GMB, KWT, LBN, LKA, OMN, QAT, SDN, SWZ, YEM, ZWE	\\
1984	&	2	&	BOL, BRA, CAN, CHL, COD, COL, GBR, GTM, ISL, LUX, MEX, PAN, PER, URY, VEN	\\
1984	&	3	&	AFG, ARE, IRQ, JOR, KEN, LAO, LBY, MNG, MOZ, MRT, PAK, SAU, SOM, SYR, THA, UGA, UKR, VNM	\\
1984	&	4	&	BEL, BLR, CSK, DEU, NOR, POL, PRY, ROU, RUS, USA	\\
1984	&	5	&	ALB, IRN, KHM	\\
1984	&	6	&	CRI	\\
\bottomrule

\end{tabular}
\caption{Infomapped communities 1980-1984 \label{tab:communities3}}

\end{table}
%%%%%%%%%%%%%%%%%%%%%%%%%%%%%%%%%%%%%%%% 

\begin{table}
\centering
\tiny
\begin{tabular}{c c p{14cm}}
\toprule
Year & Community & ISO Country Code \\
\midrule
1985	&	0	&	ARG, BDI, BEN, BGD, BHS, BRA, BRB, CAF, CMR, COG, COM, CRI, CYP, DOM, ECU, ESP, GUY, HND, HTI, IDN, IND, JOR, KEN, LKA, MAR, MEX, MLI, MMR, MOZ, MUS, NPL, OMN, PAK, PAN, PHL, SAU, SDN, SGP, SLE, SOM, SUR, SWZ, TGO, TTO, TUN, URY, VEN, YEM, YUG, ZWE	\\
1985	&	1	&	AFG, ALB, CHN, CSK, CUB, FIN, GIN, HUN, IRL, IRN, ISL, JPN, LAO, MDG, NOR, POL, ROU, RUS, SWE, SYR, USA, VNM, YDY	\\
1985	&	2	&	AGO, ARE, AUT, CAN, CHL, COD, DEU, DNK, DZA, EGY, ETH, FRA, GBR, GRC, ITA, LBR, MLT, MWI, NLD, PNG, PRY, RWA, STP, TUR	\\
1985	&	3	&	AUS, BFA, BLZ, BOL, BRN, DJI, GHA, JAM, KNA, MRT, NER, NGA, NZL, SLB, TCD, THA, TZA, UGA, VUT, WSM	\\
1985	&	4	&	BHR, COL, GTM, IRQ, KWT, LBY, MYS, QAT, SLV	\\
1985	&	5	&	BGR, BLR, DDR, KHM, MNG, UKR	\\
1985	&	6	&	BEL, LUX, PRT	\\
1985	&	7	&	PER	\\
1985	&	8	&	ISR	\\
\hline
1986	&	0	&	AGO, ARG, AUS, BDI, BEL, BGR, BOL, BRA, BRB, BTN, BWA, CAF, CHL, CMR, COD, COG, COL, CPV, DDR, DJI, DZA, EGY, ESP, ETH, FRA, GHA, GIN, GMB, GNB, GNQ, GUY, HTI, IRL, ISL, LBR, LUX, MAR, MEX, MLI, MRT, NER, NPL, PRT, PRY, QAT, RUS, RWA, SDN, SEN, SLE, STP, SWE, TCD, TUN, TZA, UKR, URY, USA, VEN, VUT, YUG, ZWE	\\
1986	&	1	&	AUT, BGD, BLR, CSK, CYP, DEU, FIN, IND, JAM, JPN, KEN, MDV, MLT, MMR, NOR, POL, ROU, SOM, TTO, TUR, VCT	\\
1986	&	2	&	ALB, ARE, BHR, BLZ, CAN, DNK, GRC, HUN, MUS, MWI, NLD, OMN, SGP, SLB, SUR, YEM	\\
1986	&	3	&	AFG, BRN, CHN, CUB, ECU, IDN, IRN, LKA, LSO, MNG, MYS, PAK, PER, THA, ZMB	\\
1986	&	4	&	ATG, BHS, FJI, GBR, GRD, ITA, NGA, NZL, PNG, WSM	\\
1986	&	5	&	CRI, DOM, KHM, PAN, SLV, TGO	\\
1986	&	6	&	KWT, LAO, SWZ, VNM, YDY	\\
1986	&	7	&	GTM, JOR, MOZ, SAU	\\
1986	&	8	&	BFA, COM, HND, IRQ, ISR, LBY, SYR	\\
1986	&	9	&	LBN, NIC	\\
1986	&	10	&	PHL	\\
\hline
1987	&	0	&	BDI, BEN, BFA, BLZ, BRA, BRN, BWA, COD, COL, COM, DJI, DOM, ETH, FJI, FRA, GIN, GNB, HTI, IDN, IND, ITA, JOR, KHM, LAO, LSO, LUX, MAR, MDG, MOZ, MUS, MYS, NER, NZL, PHL, SDN, SWE, THA, TZA, VUT, WSM	\\
1987	&	1	&	AGO, ALB, BOL, BRB, CHL, CPV, CRI, CSK, CYP, DDR, DZA, ECU, ESP, GRC, HND, IRL, ISL, KWT, LBN, MEX, MMR, MNG, PER, PRT, ROU, SLB, SLV, SUR, SYR, TUN, TUR, UKR, URY, YDY, YUG	\\
1987	&	2	&	AUS, AUT, BGD, BHS, BLR, BTN, CAN, DNK, FIN, GBR, GUY, JAM, KEN, KNA, LBR, LCA, LKA, MWI, NGA, NLD, NOR, TTO, USA, VCT, VEN	\\
1987	&	3	&	AFG, ARE, BHR, CHN, CMR, EGY, GNQ, ISR, MRT, OMN, PRY, QAT, RWA, SGP, SOM, TGO, VNM, YEM	\\
1987	&	4	&	ARG, CAF, COG, GAB, GHA, MDV, MLI, RUS, SEN, SLE, SWZ, ZWE	\\
1987	&	5	&	BEL, BGR, DEU, GTM, HUN, JPN, MLT, POL	\\
1987	&	6	&	GMB, GRD, NPL, TCD, ZMB	\\
1987	&	7	&	CUB, LBY, PAK, PAN, PNG, SAU	\\
1987	&	8	&	IRN	\\
1987	&	9	&	IRQ	\\
1987	&	10	&	NIC	\\
\hline
1988	&	0	&	AFG, ALB, ARG, BEL, BGR, BHR, BLZ, BOL, BRA, BRN, BTN, CAN, CHL, CHN, CPV, CSK, CUB, CYP, DEU, DOM, DZA, ECU, EGY, ESP, FJI, GBR, GMB, GNB, GRD, IDN, IND, ISR, ITA, JOR, KNA, LBR, LCA, LUX, MAR, MDG, MDV, MEX, MLT, MMR, MNG, MUS, MYS, NGA, NIC, NLD, NPL, PAK, PAN, PNG, PRT, PRY, RUS, SLB, SLE, SOM, STP, SUR, SWZ, THA, TTO, TUR, TZA, VEN, VUT, YDY, YUG, ZWE	\\
1988	&	1	&	AGO, ARE, BEN, BFA, BWA, CAF, CMR, COG, COM, ETH, GAB, GHA, GIN, GNQ, KEN, KHM, KWT, LAO, LBN, LBY, LSO, MOZ, MRT, MWI, OMN, SAU, SEN, SYR, TCD, TGO, TUN, VNM, YEM, ZMB	\\
1988	&	2	&	ATG, AUS, AUT, BGD, BHS, BLR, BRB, COL, CRI, DDR, DNK, FIN, FRA, GRC, GTM, HND, HUN, IRL, ISL, JAM, JPN, LKA, NOR, NZL, POL, ROU, SLV, SWE, UKR, URY, USA, VCT, WSM	\\
1988	&	3	&	HTI, NER, RWA, SDN	\\
1988	&	4	&	IRN, IRQ	\\
1988	&	5	&	COD, DJI, QAT, UGA	\\
1988	&	6	&	BDI, MLI	\\
1988	&	7	&	PER, PHL	\\
1988	&	8	&	SGP	\\
\hline
1989	&	0	&	ALB, ARE, BDI, BEL, BGR, BRB, BWA, CAN, CIV, CMR, COG, CYP, DDR, DZA, EGY, ESP, ETH, GBR, GHA, GUY, IDN, JOR, KHM, LAO, LBY, LKA, LSO, LUX, MMR, MNG, MOZ, MRT, MYS, OMN, PAK, RUS, SAU, TUR, UGA, VNM, YDY, ZMB, ZWE	\\
1989	&	1	&	AFG, ATG, BGD, BHR, BLR, DJI, DMA, FIN, FJI, GRC, GRD, ISR, MDG, MDV, MUS, NLD, NPL, NZL, POL, SDN, SEN, SOM, SUR, SWE, SYR, TGO, THA, TUN, TZA, VCT, YUG	\\
1989	&	2	&	AGO, BHS, BLZ, CHL, CUB, DOM, FRA, GIN, HND, JPN, KNA, LCA, NER, NIC, RWA, SWZ, URY, VEN	\\
1989	&	3	&	BRN, CHN, COM, GAB, GMB, GNQ, IRN, IRQ, KEN, KWT, LBR, MAR, MWI, QAT, SGP, YEM	\\
1989	&	4	&	ARG, AUT, COL, DNK, ECU, GTM, IRL, ISL, ITA, JAM, MEX, PER, PRT, PRY	\\
1989	&	5	&	CPV, CSK, GNB, IND, NGA, ROU, STP, TTO, UKR	\\
1989	&	6	&	CRI, PAN, SLV, USA, VUT	\\
1989	&	7	&	AUS, BRA, MLT, NOR, SYC	\\
1989	&	8	&	BOL, PHL	\\
1989	&	9	&	COD, MLI, TCD	\\
1989	&	10	&	BEN, BFA, HTI	\\
1989	&	11	&	DEU, HUN	\\
1989	&	12	&	PNG, SLB	\\
\bottomrule

\end{tabular}
\caption{Infomapped communities 1985-1989 \label{tab:communities4}}

\end{table}
%%%%%%%%%%%%%%%%%%%%%%%%%%%%%%%%%%%%%%%% 

\begin{table}
\centering
\tiny
\begin{tabular}{c c p{14cm}}
\toprule
Year & Community & ISO Country Code \\
\midrule
1990	&	0	&	ARG, BGD, BLZ, BRB, BTN, CHL, CHN, CMR, COG, COM, CPV, CRI, DJI, DMA, ECU, ETH, GIN, GRC, GRD, GUY, IDN, IRQ, KEN, LSO, MUS, MYS, NPL, PNG, PRY, SUR, SYR, TTO, TUR, TZA, URY, VCT, VUT, ZWE	\\
1990	&	1	&	AFG, ALB, AUS, AUT, BEL, BGR, BLR, CAF, CAN, DNK, DOM, ESP, FIN, FRA, GNQ, HND, HTI, HUN, IND, IRL, IRN, ISL, ITA, LAO, LIE, LUX, MMR, NIC, NLD, NOR, NZL, PHL, POL, PRT, ROU, RUS, RWA, SWE, TCD, UKR, USA, VEN	\\
1990	&	2	&	BDI, BFA, DZA, GAB, GHA, GMB, JOR, JPN, LBR, MDG, MDV, MLI, MNG, MOZ, NER, NGA, SLE, SOM, SYC, TGO, THA, TUN, UGA, VNM, WSM, YEM	\\
1990	&	3	&	ARE, ATG, BHR, COD, EGY, FJI, KNA, KWT, LBN, MAR, OMN, QAT, SDN, YUG	\\
1990	&	4	&	BOL, BRA, GTM, JAM, LKA, MEX, PAK	\\
1990	&	5	&	BEN, BRN, BWA, CYP, GNB, LBY, MRT, MWI, SWZ	\\
1990	&	6	&	CSK, DDR, DEU, GBR, PAN, SLV	\\
1990	&	7	&	AGO, CUB, LCA, MLT, NAM, SEN, SGP, ZMB	\\
1990	&	8	&	COL, PER	\\
1990	&	9	&	BHS	\\
1990	&	10	&	ISR	\\
1990	&	11	&	SAU	\\
1990	&	12	&	SLB	\\
\hline
1991	&	0	&	AFG, ALB, ARG, ATG, BLZ, BRB, BTN, CAF, CHN, COM, CPV, CUB, CYP, DNK, EGY, FRA, GHA, GUY, IDN, KEN, KHM, KOR, LBY, LCA, LIE, LSO, LTU, MAR, MDV, MLI, MNG, NIC, PAK, PNG, PRT, SGP, STP, SUR, SWE, THA, TTO, TUN, TUR, VCT, VNM, VUT, YEM, ZMB, ZWE	\\
1991	&	1	&	BHS, BOL, BRA, CAN, DEU, ESP, EST, FIN, FJI, FSM, GBR, HUN, IRL, ITA, MYS, NOR, PAN, PRY, ROU, SLB, URY, VEN, WSM	\\
1991	&	2	&	ARE, AUT, BFA, COG, DMA, ECU, ETH, GMB, GTM, HND, JAM, KNA, LBR, LUX, MDG, MLT, MRT, NLD, NPL, SEN, SLV, SYC, TGO, YUG	\\
1991	&	3	&	AGO, BDI, BEN, CIV, CMR, DJI, GNB, GNQ, GRD, LAO, MWI, NAM, NGA, PRK, RWA, SDN, SLE, SWZ, TCD, TZA, UGA	\\
1991	&	4	&	AUS, BEL, BGD, BRN, CHL, COL, CSK, GRC, JPN, LBN, MEX, MHL, NZL, POL, RUS, UKR, USA	\\
1991	&	5	&	DZA, GIN, IND, JOR, LKA, MMR, MUS, OMN, PHL, QAT, SAU, SYR	\\
1991	&	6	&	BHR, BLR, ISR, KWT	\\
1991	&	7	&	BGR, CRI, IRN, ISL, LVA	\\
1991	&	8	&	DOM, PER	\\
1991	&	9	&	BWA, GAB, MOZ	\\
1991	&	10	&	HTI	\\
1991	&	11	&	IRQ	\\
\hline
1992	&	0	&	AFG, ALB, ARG, BDI, BGR, BHR, BHS, BRA, BTN, CHL, COM, CPV, CSK, DEU, DJI, DMA, DZA, EGY, ETH, GMB, GNQ, GRC, HND, HRV, HUN, IDN, ISL, ITA, LAO, LBN, LBY, LIE, LSO, LUX, LVA, MAR, MDA, MDV, MNG, MRT, MUS, NAM, NER, NIC, NLD, NOR, NPL, OMN, POL, PRY, QAT, ROU, SDN, SLV, SMR, SUR, TUR, URY, VCT, ZWE	\\
1992	&	1	&	AGO, BEN, BFA, BLZ, BRN, CIV, CMR, COD, COG, GAB, GIN, GNB, IRL, LBR, LCA, LTU, MDG, MLI, MOZ, MWI, PER, PHL, PNG, PRT, RWA, SEN, SGP, SLE, STP, SWZ, TCD, TGO, VEN, VUT	\\
1992	&	2	&	AUT, BGD, BRB, DNK, FIN, FJI, GRD, IND, MHL, MLT, MMR, MYS, NGA, NZL, PAK, SLB	\\
1992	&	3	&	ATG, AZE, CRI, DOM, ECU, EST, GEO, HTI, KAZ, KGZ, TJK, UZB	\\
1992	&	4	&	AUS, BEL, COL, CYP, ESP, FRA, MEX, PAN, SWE, USA	\\
1992	&	5	&	BWA, GHA, JAM, KEN, KNA, TZA, UGA	\\
1992	&	6	&	ARE, BIH, JOR, KWT, SVN, SYR, TUN	\\
1992	&	7	&	IRN, IRQ, SAU, YEM	\\
1992	&	8	&	ARM, BLR, GTM, RUS, SYC, UKR	\\
1992	&	9	&	CAN, FSM, GBR	\\
1992	&	10	&	CHN, PRK, VNM	\\
1992	&	11	&	JPN, KOR, LKA	\\
1992	&	12	&	BOL, CUB	\\
1992	&	13	&	ISR	\\
\hline
1993	&	0	&	ARE, ARG, ARM, ATG, AUT, AZE, BEL, BHR, BHS, BLR, BLZ, BRA, BRB, BRN, BWA, CHL, DNK, ESP, FSM, GBR, GHA, GRC, GRD, IDN, IND, IRL, IRQ, ISL, ITA, JAM, JOR, JPN, KAZ, KOR, LAO, LBR, LIE, LUX, MDA, MEX, MKD, MLT, MRT, NOR, OMN, PNG, PRT, ROU, SGP, SVK, SWE, THA, TTO, TUN, TUR, UGA, UKR, USA, VNM	\\
1993	&	1	&	BEN, BGD, BTN, CAF, CHN, CIV, CMR, COD, COM, CPV, DMA, DZA, EGY, ERI, ETH, GAB, GIN, GNB, IRN, KEN, LBY, LSO, MAR, MDV, MHL, MLI, MOZ, MUS, MWI, NAM, NER, NGA, PAK, PHL, SDN, SUR, SWZ, TCD, TZA, VCT, WSM, ZWE	\\
1993	&	2	&	AFG, AGO, BDI, BFA, COG, CUB, DJI, DOM, ECU, FJI, GMB, GNQ, GTM, GUY, HND, HTI, KNA, LKA, MCO, MMR, MNG, NIC, NPL, SEN, SLE, SLV, SYC, TGO, VEN	\\
1993	&	3	&	AUS, BGR, CAN, CYP, DEU, FIN, HRV, LVA, NLD, NZL, POL	\\
1993	&	4	&	BOL, COL, KGZ, PAN, PER, PRY, TJK, TKM, UZB	\\
1993	&	5	&	ALB, BIH, CRI, CZE, EST, FRA, HUN, LTU, MYS, RUS, SMR, SVN, URY	\\
1993	&	6	&	KHM, KWT, LBN, QAT, SAU, YEM	\\
1993	&	7	&	PRK, RWA, SLB	\\
1993	&	8	&	LCA, SYR	\\
1993	&	9	&	ISR	\\
\hline
1994	&	0	&	AGO, ARE, ARM, AUS, AUT, BRB, BRN, BTN, CAN, CZE, DNK, DZA, EGY, FIN, FJI, FRA, GBR, GEO, GRD, GUY, HRV, IDN, JOR, KGZ, KNA, KWT, LKA, LUX, MCO, MDV, MHL, MKD, MLT, MNG, MUS, NLD, NPL, NZL, OMN, PNG, PRT, QAT, ROU, SEN, SGP, SLB, SMR, SVN, SYC, THA, TUN, TZA, UKR, URY	\\
1994	&	1	&	BEN, BHS, BLZ, CIV, CMR, COG, DMA, GAB, GHA, GIN, GMB, HTI, JAM, KEN, LBR, MDG, MLI, MOZ, MWI, NAM, NGA, NIC, PER, RWA, SLE, SLV, SUR, TCD, TTO, UGA, ZMB	\\
1994	&	2	&	ARG, BDI, BGD, BOL, BRA, CHL, COD, CPV, CRI, CUB, DJI, ECU, ERI, ESP, ETH, GNB, GNQ, GTM, HND, IND, IRL, ISL, LSO, MAR, NER, SWZ, TGO, USA, VCT, VEN, ZAF	\\
1994	&	3	&	AND, BGR, CYP, DEU, FSM, JPN, KOR, LIE, LTU, LVA, MYS, NOR, RUS, SVK, SWE	\\
1994	&	4	&	BHR, BIH, BLR, CHN, IRN, ITA, KAZ, MEX, SAU, TJK, TUR, UZB	\\
1994	&	5	&	AFG, BWA, COM, KHM, LBY, LCA, MRT, PHL, PRY, SDN	\\
1994	&	6	&	ALB, BEL, EST, GRC, HUN, MDA, POL	\\
1994	&	7	&	ATG, BFA, CAF, LAO, PAN, ZWE	\\
1994	&	8	&	AZE, MMR, VNM	\\
1994	&	9	&	PAK, SYR, YEM	\\
1994	&	10	&	COL, DOM	\\
1994	&	11	&	LBN, PRK	\\
1994	&	12	&	IRQ	\\
1994	&	13	&	ISR	\\
\bottomrule

\end{tabular}
\caption{Infomapped communities 1990-1994 \label{tab:communities5}}

\end{table}
%%%%%%%%%%%%%%%%%%%%%%%%%%%%%%%%%%%%%%%% 

\begin{table}
\centering
\tiny
\begin{tabular}{c c p{14cm}}
\toprule
Year & Community & ISO Country Code \\
\midrule
1995	&	0	&	ALB, AND, ARG, BHS, BLZ, BOL, CAF, CHN, COL, CRI, CUB, DJI, DOM, ECU, ETH, GNQ, GRC, GRD, HND, HTI, IDN, IRN, JAM, KAZ, KEN, KNA, LBN, LCA, MDG, MDV, MMR, MNG, PER, PHL, POL, PRY, RWA, SGP, SLV, SMR, SUR, SVN, TJK, TTO, TUR, UGA, VCT	\\
1995	&	1	&	AUT, BGD, BRB, BRN, BTN, BWA, CAN, CHL, CZE, DEU, DNK, ESP, EST, FIN, FJI, FRA, GBR, GHA, GUY, HUN, IND, IRL, ITA, KGZ, KHM, KOR, LIE, LTU, LVA, MCO, MKD, MLT, MUS, NLD, NOR, NPL, NZL, PAK, PNG, PRT, SLB, SWE, TZA, USA, VNM, WSM, ZAF, ZMB	\\
1995	&	2	&	ARE, BDI, BFA, BHR, CMR, COD, COG, DZA, EGY, GAB, GMB, IRQ, JOR, KWT, LBR, LBY, LKA, MYS, NER, NGA, OMN, QAT, SDN, SLE, SWZ, SYR, TGO, YEM	\\
1995	&	3	&	ARM, AZE, BEL, BGR, BRA, CYP, GEO, ISL, JPN, LUX, MDA, MEX, NIC, PAN, ROU, RUS, SVK, UKR, URY	\\
1995	&	4	&	AGO, BEN, BLR, ERI, GIN, GNB, GTM, HRV, LAO, LSO, MAR, MLI, MOZ, MRT, MWI, NAM, SEN, TCD, THA, TKM, TUN	\\
1995	&	5	&	AUS, FSM, MHL	\\
1995	&	6	&	BIH, ISR, VEN	\\
1995	&	7	&	PRK	\\
1995	&	8	&	AFG	\\
1995	&	9	&	ATG	\\
\hline
1996	&	0	&	AGO, ARE, AZE, BDI, BEN, BFA, BHR, BHS, BRN, BTN, BWA, CAF, CIV, COD, COM, CZE, DJI, DOM, DZA, EGY, ERI, ESP, GAB, GHA, GIN, GMB, GNB, HND, HTI, IRL, IRN, ISR, ITA, JOR, JPN, KEN, KHM, KWT, LAO, LBN, LBY, LSO, LUX, MAR, MDA, MOZ, MRT, MWI, NAM, NGA, OMN, PRK, QAT, ROU, RWA, SAU, SDN, SEN, SLE, SMR, TGO, TJK, TUN, TUR, TZA, UGA, UZB, YEM, ZAF, ZMB, ZWE	\\
1996	&	1	&	ALB, AND, ATG, AUS, AUT, BEL, BGR, BIH, BLR, BRA, CHN, COL, CRI, DEU, EST, FIN, FJI, FRA, GRC, GTM, IDN, ISL, KGZ, KOR, LTU, LVA, MCO, MEX, MLT, NIC, NLD, PAK, PAN, PHL, PLW, PNG, POL, PRY, RUS, SGP, SVN, SWE, SWZ, SYC, SYR, UKR, USA, VCT, VUT	\\
1996	&	2	&	AFG, ARM, BGD, BLZ, BRB, CAN, CHL, CYP, DMA, DNK, ECU, FSM, GNQ, GRD, GUY, HUN, IND, JAM, KAZ, KNA, LCA, LIE, LKA, MHL, MMR, MNG, MUS, NPL, NZL, PRT, SLB, SLV, SUR, THA, TKM, TTO, VNM, WSM	\\
1996	&	3	&	COG, CPV, ETH, LBR, MDG, MLI, NER, TCD	\\
1996	&	4	&	CUB, GBR, MYS, NOR, PER, VEN	\\
1996	&	5	&	BOL, GEO, HRV, MKD	\\
1996	&	6	&	ARG, MDV, SVK, URY	\\
1996	&	7	&	IRQ	\\
\hline
1997	&	0	&	AGO, ATG, BDI, BEN, BFA, BGD, BLZ, BOL, BRN, BWA, CHN, CIV, COL, COM, CRI, DJI, DZA, EGY, ERI, ESP, FIN, GAB, GIN, GNB, GRD, GUY, HND, IDN, IRL, IRN, ISL, ISR, JOR, KNA, KOR, KWT, LAO, LBR, LCA, LSO, MDA, MDG, MDV, MHL, MLI, MLT, MMR, MOZ, MRT, MYS, NAM, NER, NGA, NIC, NPL, PAK, PER, PHL, PRT, QAT, SDN, SEN, SLB, SWZ, SYR, TCD, TGO, THA, TUN, TZA, USA, VCT, VNM, VUT, ZAF, ZWE	\\
1997	&	1	&	ALB, AND, BGR, BIH, BLR, BRA, BTN, EST, FRA, GBR, GTM, HRV, HUN, IND, ITA, JPN, LTU, LUX, LVA, MEX, MKD, PAN, POL, ROU, RUS, SGP, SLE, SLV, SVN, UKR, URY, VEN	\\
1997	&	2	&	ARG, ARM, AUS, AUT, AZE, BEL, CAN, CHL, CUB, CZE, DEU, DNK, DOM, ECU, GNQ, LIE, MCO, NLD, NOR, PRY, SMR, SWE	\\
1997	&	3	&	BHS, BRB, CAF, CPV, ETH, FJI, GHA, MWI, PRK, SYC	\\
1997	&	4	&	COD, DMA, HTI, JAM, KEN, MUS, PNG, TTO, UGA, ZMB	\\
1997	&	5	&	CYP, GRC, KAZ, KGZ, NZL, SVK, TKM, TUR, UZB	\\
1997	&	6	&	ARE, BHR, MAR, OMN, YEM	\\
1997	&	7	&	IRQ, LBN	\\
1997	&	8	&	GEO, TJK	\\
1997	&	9	&	GMB, MNG, SUR	\\
1997	&	10	&	AFG	\\
1997	&	11	&	COG	\\
1997	&	12	&	LBY	\\
1997	&	13	&	LKA	\\
1997	&	14	&	RWA	\\
\hline
1998	&	0	&	BDI, BLZ, BRN, BWA, CAF, COD, COG, DJI, DMA, EGY, FJI, GAB, GNQ, GRC, GUY, IRQ, JAM, KWT, LBR, LCA, LSO, LUX, MDV, MMR, MUS, MWI, NAM, NGA, NZL, PNG, SLB, SYC, TZA, UGA, VCT, VUT, WSM, ZMB, ZWE	\\
1998	&	1	&	AND, ARG, AUS, AUT, BEL, BEN, BFA, BGD, BRA, CAN, CHL, CHN, CMR, CPV, DEU, DNK, DZA, ESP, EST, FRA, GBR, GHA, HND, HRV, IDN, IND, IRL, JPN, MAR, MDG, MYS, NER, NIC, NOR, PAK, PER, PSE, QAT, STP, SVN, SWE, TGO, THA, USA, ZAF	\\
1998	&	2	&	ALB, BHR, BOL, CIV, COL, CRI, DOM, GIN, GMB, GTM, HTI, KAZ, KGZ, KOR, MEX, MNG, MRT, NPL, OMN, PAN, PHL, PRY, RUS, SDN, SGP, SMR, SYR, TCD, TUN, UKR, URY, VNM, YEM	\\
1998	&	3	&	ARM, AZE, BGR, BLR, CYP, CZE, GEO, IRN, ITA, LIE, LKA, LTU, MKD, MLT, POL, ROU, TJK, TUR	\\
1998	&	4	&	ARE, BHS, CUB, HUN, ISR, LAO, LBN, MLI, PRK, SEN, VEN	\\
1998	&	5	&	ATG, BIH, BRB, KEN, LVA, SLV, SUR, TTO	\\
1998	&	6	&	FIN, FSM, ISL, JOR, MOZ, NLD, SAU	\\
1998	&	7	&	MCO, MDA, SVK, TKM, UZB	\\
1998	&	8	&	GNB, KNA, LBY, SLE	\\
1998	&	9	&	AFG, COM, GRD, RWA	\\
1998	&	10	&	AGO, BTN, ECU, ETH, PRT, SWZ	\\
1998	&	11	&	ERI	\\
\hline
1999	&	0	&	AFG, ARE, ATG, BGR, BHR, BLR, BRN, CAN, CHL, CHN, COD, CUB, CZE, DJI, DMA, EGY, ERI, ESP, EST, ETH, FRA, GMB, GNQ, GTM, HUN, IRN, ISL, ISR, JAM, KNA, KWT, LAO, LBY, LIE, LKA, LVA, MAR, MDV, MHL, MNG, MRT, NLD, NZL, OMN, PHL, PNG, POL, PSE, SLB, SUR, SVK, SWE, SWZ, THA, TJK, TKM, TUN, TUR, UKR, URY, USA, VNM, VUT, WSM, YEM, ZAF	\\
1999	&	1	&	AND, BEL, BEN, BIH, BWA, CMR, COM, CPV, DEU, DNK, DZA, FIN, FJI, GAB, GBR, GNB, GUY, LSO, LUX, MCO, MLI, MLT, MOZ, MUS, MWI, MYS, NAM, NGA, NOR, SLE, SMR, TGO, TTO, ZMB, ZWE	\\
1999	&	2	&	AUS, BDI, BHS, BLZ, BOL, BRA, BRB, DOM, ECU, GRD, HRV, KHM, LCA, MMR, PRT, ROU, SEN, SVN, VCT	\\
1999	&	3	&	ARM, AZE, FSM, GEO, HND, JOR, KAZ, KGZ, LBN, MDA, PLW, PRK, QAT, RUS, SAU, SYR	\\
1999	&	4	&	ALB, BGD, COL, HTI, IDN, IRL, MEX, MKD, NIC, PER, SDN	\\
1999	&	5	&	ARG, AUT, BFA, CYP, GIN, JPN, KEN, KOR, LTU, NPL, TCD, UZB	\\
1999	&	6	&	AGO, BTN, COG, CRI, PAN, PRY, RWA, UGA	\\
1999	&	7	&	GRC, ITA, SGP, SLV	\\
1999	&	8	&	CIV, GHA, IND, LBR, MDG, TZA	\\
1999	&	9	&	IRQ	\\
1999	&	10	&	NER	\\
1999	&	11	&	PAK	\\
1999	&	12	&	VEN	\\
\bottomrule

\end{tabular}
\caption{Infomapped communities 1995-1999 \label{tab:communities6}}

\end{table}
%%%%%%%%%%%%%%%%%%%%%%%%%%%%%%%%%%%%%%%% 

\begin{table}
\centering
\tiny
\begin{tabular}{c c p{14cm}}
\toprule
Year & Community & ISO Country Code \\
\midrule
2000	&	0	&	AGO, ARE, BGD, BGR, BHR, BIH, BLR, CHL, COM, CYP, DJI, DZA, EGY, FIN, FRA, GHA, GMB, HRV, IDN, IRN, JOR, KEN, KWT, LBN, LBY, MAR, MDV, MRT, MUS, MYS, NZL, OMN, PAK, QAT, SAU, SDN, SEN, SLB, SLE, SLV, SMR, SVN, SWE, TKM, TUN, TUR, UGA, VEN, YEM, ZWE	\\
2000	&	1	&	ARG, AUT, CAN, CHN, COL, CPV, ETH, FSM, GUY, ITA, KOR, LAO, LBR, MEX, MLI, MNG, NIC, PHL, PRY, ROU, RUS	\\
2000	&	2	&	BHS, BRB, CUB, DMA, GBR, GRC, GRD, HND, JAM, KGZ, KNA, LVA, MMR, MWI, NAM, NPL, NRU, RWA, TJK, URY, VCT	\\
2000	&	3	&	BLZ, BOL, BRN, BWA, COG, DOM, GAB, GIN, GNB, GNQ, HTI, IND, LSO, MDG, MHL, SWZ, TTO	\\
2000	&	4	&	BEN, BFA, BRA, CAF, ECU, GTM, NER, NGA, NOR, SUR, TCD, TGO, THA	\\
2000	&	5	&	ALB, AUS, DEU, ESP, GEO, HUN, LIE, MKD, MLT, PER, POL, SVK	\\
2000	&	6	&	CRI, EST, FJI, IRL, ISL, KHM, PAN, PNG, TZA, VNM, VUT, ZAF	\\
2000	&	7	&	AZE, CIV, CZE, DNK, ISR, JPN, KAZ, LTU, MDA, PRT, SGP, UKR	\\
2000	&	8	&	AFG, IRQ, PSE, SYR	\\
2000	&	9	&	ARM, BEL, ERI, LUX, PRK, USA	\\
2000	&	10	&	AND, ATG, COD	\\
2000	&	11	&	LKA, SOM	\\
2000	&	12	&	BDI, MCO	\\
2000	&	13	&	LCA	\\
2000	&	14	&	NLD	\\
2000	&	15	&	PLW	\\
\hline
2001	&	0	&	AGO, AUT, BRA, CHL, CHN, CPV, CYP, CZE, DNK, ECU, ESP, ETH, FRA, GTM, HUN, IRL, ISL, ITA, JOR, JPN, KAZ, KOR, LIE, LTU, MCO, MKD, MRT, NER, NIC, OMN, PAK, PER, PRT, SAU, SVK, SWE, SYC, TUN, TUR, YEM	\\
2001	&	1	&	ARE, ARG, ARM, BEL, BGD, BHR, BOL, DJI, DOM, EST, GEO, HND, HTI, IDN, JAM, KEN, KHM, KWT, LAO, LUX, LVA, MAR, MLT, MMR, MOZ, NPL, PRK, PRY, PSE, SLV, SMR, SOM, THA, VNM, ZAF	\\
2001	&	2	&	ALB, BDI, BFA, BIH, BLR, BRN, CAF, COD, COG, DZA, ERI, FJI, FSM, GAB, GIN, LBR, LBY, LSO, MDG, MEX, MWI, NGA, PNG, ROU, SDN, SLE, TKM, TZA, YUG, ZWE	\\
2001	&	3	&	BHS, BLZ, BWA, CIV, GHA, GMB, GNB, IRQ, LCA, MDA, MDV, MHL, MUS, MYS, NAM, PAN, SLB, SWZ, TGO, UKR	\\
2001	&	4	&	AND, COL, GBR, GRC, GUY, KGZ, LBN, NOR, POL, QAT, RWA, STP, UZB, WSM	\\
2001	&	5	&	AZE, BEN, BGR, CMR, COM, MLI, RUS, SEN, TCD, TJK, ZMB	\\
2001	&	6	&	AFG, AUS, BTN, CAN, DEU, FIN, IND, SYR, URY, VEN	\\
2001	&	7	&	BRB, GNQ, GRD, KNA, MNG, SUR	\\
2001	&	8	&	EGY, HRV, IRN, LKA, SVN	\\
2001	&	9	&	NRU, PLW, TON, TUV, VCT	\\
2001	&	10	&	CRI, CUB, NLD, NZL, USA	\\
2001	&	11	&	ISR, UGA	\\
2001	&	12	&	ATG, TTO	\\
2001	&	13	&	DMA	\\
2001	&	14	&	PHL	\\
2001	&	15	&	SGP	\\
2001	&	16	&	VUT	\\
\hline
2002	&	0	&	AFG, ARE, ARG, AZE, BEL, BFA, BGD, BGR, BLR, BRA, BRN, BTN, CHE, CHL, CHN, CMR, CRI, CYP, CZE, DJI, DOM, DZA, EGY, ESP, GMB, GNQ, GTM, GUY, HRV, IDN, IND, ISL, JAM, JPN, KHM, KNA, LKA, LUX, MCO, MDA, MDG, MDV, MEX, MKD, MNG, MRT, NER, NGA, NIC, NLD, NOR, NPL, PHL, PRT, QAT, SDN, SGP, SMR, SWZ, THA, TJK, TKM, TUR, UKR, VCT, VNM	\\
2002	&	1	&	AND, AUT, BHR, DEU, DNK, FIN, GBR, HUN, IRL, JOR, KAZ, KWT, MYS, NZL, PAK, ROU, RUS, SAU, SVK, SWE, USA, UZB, YEM	\\
2002	&	2	&	BEN, COG, CPV, GAB, GHA, GIN, GNB, KEN, LSO, MLI, MOZ, MUS, MWI, RWA, SLE, SOM, TGO, UGA, ZMB	\\
2002	&	3	&	ARM, BDI, BHS, BLZ, BWA, CAF, COM, ECU, LCA, MMR, NAM, PAN, PRY, TZA, URY, ZAF	\\
2002	&	4	&	ATG, BOL, BRB, COD, DMA, FJI, FSM, MHL, NRU, PNG, SLB, TON, TTO, TUV, VUT, WSM	\\
2002	&	5	&	ALB, BIH, CAN, EST, GRC, IRN, ITA, KOR, LIE, LTU, SVN, YUG	\\
2002	&	6	&	AUS, COL, CUB, IRQ, ISR, LBN, LVA, MLT, OMN, PLW, SYR	\\
2002	&	7	&	HND, KGZ, LAO, PER, SLV, SUR, ZWE	\\
2002	&	8	&	CIV, ETH, FRA, GEO, MAR, POL, PSE, TUN	\\
2002	&	9	&	AGO, ERI, SEN, TCD	\\
2002	&	10	&	LBR, PRK	\\
2002	&	11	&	GRD, HTI	\\
2002	&	12	&	STP	\\
2002	&	13	&	VEN	\\
\hline
2003	&	0	&	AGO, ALB, ARE, AUT, BDI, BEN, BGR, BOL, CHE, CIV, CMR, CRI, CZE, DOM, ESP, FIN, GHA, GIN, GNB, GNQ, HRV, ISR, ITA, JPN, KHM, LAO, LKA, LTU, MDG, MLT, MOZ, MRT, MWI, OMN, PAN, PNG, RWA, SEN, SMR, SUR, SWE, TGO, TLS, TUN, TZA, URY, VEN, YEM	\\
2003	&	1	&	BFA, BGD, BRA, BRB, BTN, CAN, DMA, EST, GMB, GRD, GUY, IRL, JAM, KAZ, KIR, KNA, LCA, LUX, MEX, MHL, MLI, MMR, MNG, MUS, NAM, NGA, NOR, NPL, PLW, ROU, RUS, SDN, SLB, SLE, SVK, SYC, TTO, TUV, UKR, UZB, VCT, VNM, WSM, ZMB	\\
2003	&	2	&	BHR, BLZ, BRN, CHL, CHN, COM, DEU, DNK, FRA, GBR, GTM, KEN, KWT, LSO, MAR, MYS, NIC, PER, POL, PRY, PSE, QAT, SAU, SGP, STP, SVN, VUT, ZWE	\\
2003	&	3	&	ARM, AZE, CAF, COD, COG, CYP, GRC, KOR, MKD, NER, PRK, TCD, THA, TUR, UGA	\\
2003	&	4	&	ARG, BEL, BIH, BWA, CPV, ETH, FJI, GAB, HND, KGZ, LBR, MCO, MDA, NRU, TJK	\\
2003	&	5	&	DZA, GEO, ISL, LIE, LVA, NZL, PHL, PRT, TON, ZAF	\\
2003	&	6	&	BLR, EGY, IDN, IRN, JOR, LBN, PAK, SOM, SYR	\\
2003	&	7	&	ATG, BHS, FSM, IND, MDV	\\
2003	&	8	&	HTI, SLV, USA	\\
2003	&	9	&	COL, ECU, ERI	\\
2003	&	10	&	AND, AUS, HUN, NLD, SWZ	\\
2003	&	11	&	AFG, CUB	\\
2003	&	12	&	YUG	\\
2003	&	13	&	IRQ	\\
\bottomrule

\end{tabular}
\caption{Infomapped communities 2000-2003 \label{tab:communities7}}

\end{table}
%%%%%%%%%%%%%%%%%%%%%%%%%%%%%%%%%%%%%%%% 

\begin{table}
\centering
\tiny
\begin{tabular}{c c p{14cm}}
\toprule
Year & Community & ISO Country Code \\
\midrule
2004	&	0	&	AUS, BDI, BHR, BLR, BWA, CAF, CAN, CMR, CYP, CZE, DJI, DZA, ETH, FIN, GAB, GHA, GIN, GTM, GUY, IRN, ITA, KGZ, LSO, LUX, LVA, MAR, MKD, MLI, MOZ, NGA, NPL, OMN, PAK, POL, QAT, RUS, SLE, SVN, SWE, SYR, TGO, TZA	\\
2004	&	1	&	BLZ, BRB, COM, DEU, DMA, DOM, FJI, FSM, JAM, KIR, LIE, MCO, MDV, MEX, MUS, NRU, NZL, PAN, PLW, PNG, PRT, RWA, SLB, SLV, STP, SUR, SYC, TUV, WSM	\\
2004	&	2	&	BRA, BRN, CPV, ESP, GBR, GRC, HUN, IRL, ISL, LAO, LTU, MNG, NLD, SGP, TUR, UKR, URY, USA, VAT, VCT, YUG	\\
2004	&	3	&	AGO, ARG, BEL, CHE, CHL, FRA, GNB, GRD, HTI, JOR, LBY, MWI, NOR, PSE, SAU, SEN, TCD, TUN, ZWE	\\
2004	&	4	&	ARE, AZE, BEN, BFA, EGY, ISR, KEN, KWT, LBR, MDA, NER, YEM, ZMB	\\
2004	&	5	&	AND, ATG, BHS, BOL, CRI, ECU, GEO, GNQ, HND, PHL, PRY, TTO, ZAF	\\
2004	&	6	&	ARM, CUB, IDN, IND, JPN, KHM, LCA, PER, THA, TJK, TLS, TON, VEN	\\
2004	&	7	&	BTN, CHN, KAZ, MMR, ROU, SWZ, TKM, UZB, VUT	\\
2004	&	8	&	AFG, ALB, AUT, BGR, DNK, SVK	\\
2004	&	9	&	IRQ, LBN, MRT, MYS, NAM, SDN	\\
2004	&	10	&	BGD, EST, HRV, LKA, MHL, MLT	\\
2004	&	11	&	CIV, COD, COG, UGA	\\
2004	&	12	&	BIH, GMB	\\
2004	&	13	&	MDG, PRK	\\
2004	&	14	&	COL, KNA, NIC	\\
2004	&	15	&	KOR, SMR, VNM	\\
2004	&	16	&	ERI	\\
\hline
2005	&	0	&	AGO, ALB, ARE, BDI, BEL, BEN, BFA, BIH, BRA, CIV, COD, COG, COM, DEU, DZA, EGY, ESP, EST, FIN, FRA, GAB, GBR, GIN, GMB, GNB, HUN, IRL, IRQ, ISR, KEN, KWT, LBN, LBY, LUX, MAR, MRT, NAM, NER, NGA, PAK, QAT, SDN, SEN, SYR, TCD, TZA, UZB, YEM, ZMB	\\
2005	&	1	&	AFG, AND, BRB, BRN, BTN, CHN, CZE, FJI, GUY, IND, KAZ, KGZ, KHM, KIR, LVA, MDA, MHL, MKD, MUS, PHL, ROU, RWA, SGP, SVN, SWZ, THA, TJK, TUV, WSM	\\
2005	&	2	&	ARM, AUS, AZE, CAN, CPV, CYP, DNK, ECU, ETH, GEO, GTM, LAO, LSO, MCO, MDV, MWI, MYS, NLD, PER, SMR, STP, UKR, ZAF	\\
2005	&	3	&	ARG, BGD, CHE, GHA, HRV, IDN, JAM, JPN, LKA, MEX, NOR, NPL, NZL, POL, SWE, TLS, TON, USA	\\
2005	&	4	&	ATG, BHS, BLR, CHL, CMR, CRI, DMA, FSM, GRD, KNA, LCA, SVK, TTO, VNM	\\
2005	&	5	&	CUB, GNQ, ITA, LTU, MMR, MNG, MOZ, PAN, PNG, SLB, SLE, SLV, SUR, VAT	\\
2005	&	6	&	AUT, BGR, BLZ, KOR, MDG, MLT, NRU, PLW, PRT, VEN, VUT	\\
2005	&	7	&	BHR, BOL, ERI, GRC, PRK, PSE, TGO, TUR	\\
2005	&	8	&	DOM, NIC, PRY, SOM, UGA, URY	\\
2005	&	9	&	CAF, HTI, JOR, TUN, YUG	\\
2005	&	10	&	COL, ISL, LIE, RUS	\\
2005	&	11	&	IRN	\\
2005	&	12	&	TKM	\\
2005	&	13	&	VCT	\\
2005	&	14	&	ZWE	\\
\hline
2006	&	0	&	ALB, AND, ARE, ARG, AUS, AUT, BEL, BFA, BGR, BLR, BRA, CHN, CMR, COM, CPV, CYP, DNK, DZA, EGY, ETH, FRA, GEO, GIN, GNQ, GRC, HUN, IDN, IND, IRL, ITA, JOR, KEN, KOR, KWT, LBN, LBR, LBY, LTU, MAR, MCO, MDG, MDV, MRT, MYS, NIC, NLD, NPL, NZL, OMN, PAK, POL, PRK, PRT, PSE, ROU, SDN, SEN, SLB, SMR, SOM, SVK, SVN, SWE, SYR, TCD, TGO, TLS, TUN, TUR, TUV, USA, UZB, VAT, YEM	\\
2006	&	1	&	AGO, ARM, AZE, BDI, BEN, BIH, BWA, CAF, CIV, COD, COG, CZE, DOM, ESP, EST, FIN, GAB, GHA, GNB, HRV, JPN, KHM, LSO, LUX, LVA, MDA, MKD, MLT, MMR, MNE, MOZ, MWI, NAM, NER, NGA, NOR, PAN, SLV, TJK, UKR, YUG, ZMB	\\
2006	&	2	&	BGD, BOL, BRN, BTN, CUB, ECU, FJI, FSM, GMB, GUY, ISL, KAZ, KGZ, KIR, MEX, PHL, PLW, RWA, SGP, SUR, TKM, TON, URY	\\
2006	&	3	&	ATG, BHS, CRI, DMA, HTI, JAM, LCA, LKA, MUS, PER, PNG, STP, SYC, TTO, VNM, VUT, ZWE	\\
2006	&	4	&	BRB, CHL, GRD, GTM, LAO, LIE, MLI, MNG, NRU, RUS, SLE, THA, UGA, WSM	\\
2006	&	5	&	AFG, BHR, CAN, CHE, DEU, GBR, IRQ, ISR, QAT, SAU	\\
2006	&	6	&	COL, IRN, MHL, PRY, VEN	\\
2006	&	7	&	BLZ, HND, VCT	\\
2006	&	8	&	ERI, KNA	\\
2006	&	9	&	SWZ, TZA, ZAF	\\
\hline
2007	&	0	&	AGO, ALB, BDI, BEL, BEN, BFA, BHR, BHS, BRN, BWA, CAF, CAN, CMR, COD, COM, CYP, DEU, DZA, EST, GAB, GHA, GIN, GMB, GNQ, GTM, JPN, KAZ, KHM, KWT, LBR, LIE, LSO, LVA, MKD, MLT, MNG, MRT, NAM, NGA, NIC, OMN, PAK, PER, ROU, RUS, RWA, SDN, SLV, SMR, SVK, TGO, TUN, TUR, TZA, UGA, VAT, VEN, VNM, YEM	\\
2007	&	1	&	ARG, ATG, BGD, BLR, BLZ, BRA, BRB, BTN, CHL, DMA, ECU, ETH, FJI, HND, HRV, HTI, IDN, IND, KGZ, KIR, LAO, LCA, LTU, MCO, MDA, MMR, MNE, MOZ, NOR, NRU, PLW, POL, PRK, STP, SUR, SWZ, TJK, TKM, TON, TTO, TUV, URY	\\
2007	&	2	&	ARM, AUS, AUT, AZE, CPV, CRI, DNK, EGY, FRA, FSM, GBR, GRD, IRL, ISL, ISR, JAM, JOR, LUX, MDG, MDV, MHL, MUS, MYS, NLD, NZL, PAN, PNG, PRY, SEN, SWE, SYC, THA, USA, VUT, YUG, ZMB	\\
2007	&	3	&	ARE, BGR, CHE, ESP, FIN, GRC, HUN, KOR, MAR, MEX, PRT, QAT, UKR	\\
2007	&	4	&	AFG, CIV, COG, GNB, LBY, NER, SLE, SOM, TCD	\\
2007	&	5	&	AND, BOL, DOM, GUY, KNA, MWI, VCT, ZAF	\\
2007	&	6	&	LBN, LKA, PSE, SGP, TLS	\\
2007	&	7	&	COL, CUB, IRN, ZWE	\\
2007	&	8	&	CHN, CZE, ITA, NPL, PHL, UZB	\\
2007	&	9	&	BIH, IRQ, SLB, SVN, SYR, WSM	\\
2007	&	10	&	ERI	\\
2007	&	11	&	GEO	\\
2007	&	12	&	KEN	\\
\bottomrule

\end{tabular}
\caption{Infomapped communities 2004-2007 \label{tab:communities8}}

\end{table}
%%%%%%%%%%%%%%%%%%%%%%%%%%%%%%%%%%%%%%%% 

\begin{table}
\centering
\tiny
\begin{tabular}{c c p{14cm}}
\toprule
Year & Community & ISO Country Code \\
\midrule
2008	&	0	&	ARM, ATG, BDI, BGD, BHS, BRB, BWA, CAN, CHE, COG, DMA, EST, ETH, FJI, GBR, GEO, GRD, HND, HTI, IDN, IND, IRL, ISL, JAM, KEN, KIR, KWT, LAO, LUX, MCO, MDG, MDV, MEX, MHL, MNG, MOZ, MUS, MWI, NAM, NOR, NPL, NZL, PAN, PNG, PRT, SLB, STP, SUR, SWZ, SYC, TON, UKR, WSM	\\
2008	&	1	&	AZE, BEN, BFA, BHR, BLR, CIV, CMR, DNK, DZA, ECU, GIN, GNB, HUN, IRQ, JOR, JPN, KGZ, KHM, KOR, LBN, LBY, LSO, LVA, MAR, MLI, MNE, MRT, NER, OMN, PER, PSE, ROU, SDN, SLE, SOM, SVK, SVN, SYR, TCD, TGO, TJK, TKM, TUN, URY, YEM, ZMB	\\
2008	&	2	&	AFG, AGO, ARE, AUT, BEL, BRA, CAF, COD, COM, CPV, CYP, DOM, FIN, GAB, GMB, GRC, ITA, KNA, LBR, QAT, SWE, TTO, TUR, TZA, VNM, VUT, ZAF, ZWE	\\
2008	&	3	&	ALB, ARG, CHL, CZE, DEU, ESP, GNQ, HRV, IRN, LIE, LKA, MKD, MLT, MYS, SMR, USA, VEN	\\
2008	&	4	&	BGR, CHN, EGY, FRA, ISR, MDA, NGA, POL, PRK, RUS, SEN, UZB	\\
2008	&	5	&	AND, BTN, FSM, GUY, LCA, PHL, TUV	\\
2008	&	6	&	CUB, ERI, KAZ, NIC, PRY, RWA, SGP, VCT	\\
2008	&	7	&	GTM, LTU, SLV, VAT	\\
2008	&	8	&	BLZ, NLD	\\
2008	&	9	&	AUS, BIH	\\
2008	&	10	&	COL, NRU, PLW, UGA	\\
2008	&	11	&	BRN, GHA, MMR, TLS	\\
2008	&	12	&	BOL	\\
2008	&	13	&	CRI	\\
2008	&	14	&	PAK	\\
2008	&	15	&	THA	\\
2008	&	16	&	YUG	\\
\hline
2009	&	0	&	ALB, ARG, ARM, ATG, AUS, AUT, AZE, BEL, BFA, BGD, BIH, BRA, BRB, BTN, BWA, CAF, CHE, COM, DJI, DZA, ESP, EST, ETH, FIN, FJI, GAB, GBR, GEO, GHA, GIN, GRC, HRV, IRN, JAM, KEN, KHM, KWT, LAO, LCA, LIE, LSO, LUX, MAR, MDV, MEX, MHL, MKD, MLI, MLT, MMR, MNE, MRT, MWI, NAM, NER, NGA, NLD, NOR, NZL, PHL, PNG, PRT, QAT, RWA, SDN, SEN, SLE, SOM, STP, SVN, SWE, SWZ, TCD, TGO, TLS, TTO, TUN, TUR, USA, VAT, VUT, WSM, YEM, YUG, ZAF, ZMB, ZWE	\\
2009	&	1	&	AFG, ARE, BHR, BLR, CAN, CHN, CYP, DEU, DNK, EGY, HUN, IDN, IND, IRL, IRQ, JOR, KAZ, KGZ, KOR, LBN, LKA, LTU, LVA, MDA, MNG, MYS, NPL, PAK, PRK, PSE, ROU, RUS, SVK, VNM	\\
2009	&	2	&	AND, BEN, BLZ, CIV, CMR, COD, COG, ECU, GRD, ISL, ITA, KNA, MCO, MOZ, MUS, PER, SLB, SLV, SMR, SUR, SYC, TON, TUV	\\
2009	&	3	&	AGO, BDI, BHS, BOL, COL, CPV, CRI, CUB, CZE, DOM, ERI, FRA, GNB, GNQ, GTM, HTI, LBR, MDG, NIC, POL, TZA, URY	\\
2009	&	4	&	BRN, CHL, DMA, FSM, GMB, GUY, HND, KIR, NRU, PAN, PLW, SGP, VCT	\\
2009	&	5	&	LBY, VEN	\\
2009	&	6	&	OMN, SYR, UKR	\\
2009	&	7	&	JPN, THA, TJK	\\
2009	&	8	&	BGR, TKM, UZB	\\
2009	&	9	&	ISR	\\
2009	&	10	&	PRY	\\
2009	&	11	&	UGA	\\
\hline
2010	&	0	&	ARG, ARM, AZE, BGR, BIH, BLZ, CAN, CHE, CHN, COD, COL, CUB, CZE, DEU, FRA, GBR, GNB, ISL, ITA, JOR, LIE, LKA, LSO, MYS, NIC, PAK, PAN, PHL, PLW, PSE, QAT, SGP, SLV, THA, TLS, VAT, WSM	\\
2010	&	1	&	AGO, ALB, AUS, AUT, BFA, BLR, COG, DNK, EGY, ERI, EST, ETH, FSM, GHA, GIN, GNQ, HRV, IRL, IRQ, KEN, KHM, KIR, KOR, KWT, LBR, LTU, LUX, MDA, MHL, MKD, MMR, MRT, NOR, NZL, PNG, PRK, SDN, SLB, SVN, SWE, SYR, TCD, TZA, VNM, YUG	\\
2010	&	2	&	BDI, BGD, BHS, BRA, BRB, BTN, CAF, DMA, GMB, GRD, GTM, JAM, KNA, LCA, LVA, MLI, NER, NRU, OMN, ROU, SWZ, SYC, TON, TUN	\\
2010	&	3	&	ARE, DOM, DZA, HUN, IDN, IND, JPN, KGZ, MAR, MEX, MLT, NGA, NPL, POL, PRT, SVK, URY, VUT	\\
2010	&	4	&	AND, FIN, KAZ, MCO, MNE, MNG, MOZ, MUS, MWI, NAM, SLE, TJK, TTO, TUV, ZAF	\\
2010	&	5	&	CHL, CPV, CRI, ESP, HND, HTI, PRY, RWA, SMR, STP, SUR, UGA, VEN, ZMB, ZWE	\\
2010	&	6	&	ATG, BRN, CIV, GUY, SEN, VCT	\\
2010	&	7	&	ECU, GEO, IRN, PER, USA	\\
2010	&	8	&	AFG, CYP, MDV, UKR	\\
2010	&	9	&	FJI, GAB, LBN, RUS, SOM, TUR	\\
2010	&	10	&	BHR, GRC, LBY, YEM	\\
2010	&	11	&	BEN, BWA	\\
2010	&	12	&	BEL, LAO, NLD, TGO	\\
2010	&	13	&	BOL	\\
2010	&	14	&	CMR	\\
2010	&	15	&	COM	\\
2010	&	16	&	ISR	\\
\hline
2011	&	0	&	AGO, ALB, AND, AZE, BEN, BFA, BHS, BTN, BWA, CIV, COL, COM, CZE, DNK, DZA, ETH, FJI, GIN, GMB, GNB, GNQ, GTM, HND, IND, KGZ, KOR, LBN, LSO, MDG, MNG, MOZ, MRT, PAN, PLW, POL, PRT, QAT, ROU, RUS, RWA, SDN, STP, SVK, SVN, SWZ, TCD, TGO, TUN, TZA, UGA, UKR, VAT, VUT, YEM, ZAF, ZMB, ZWE	\\
2011	&	1	&	AFG, AUT, COG, DJI, EGY, ERI, ESP, GHA, GRC, GRD, HRV, HUN, IDN, ITA, JAM, KHM, LAO, LBR, LKA, LVA, MAR, MCO, MDV, MLI, MLT, MWI, PER, PRY, SGP, SLB, SYR, THA, TLS, URY, VEN, VNM, YUG	\\
2011	&	2	&	ARM, AUS, BGR, BLR, BRB, GUY, IRL, KAZ, LUX, MKD, MYS, NOR, NPL, PAK, SOM, TUR	\\
2011	&	3	&	BEL, BGD, BRN, CAF, CHL, DEU, EST, FIN, GBR, IRQ, LIE, LTU, NER, NGA, PHL, PNG, SLE, SMR, SUR, TUV	\\
2011	&	4	&	BHR, BIH, CHN, CMR, CYP, GAB, JPN, KWT, MDA, MMR, MNE, OMN, TKM, UZB	\\
2011	&	5	&	ARG, ATG, BRA, CHE, CPV, CRI, CUB, ECU, GEO, JOR, LBY, NIC, NLD, SWE, USA, VCT	\\
2011	&	6	&	BLZ, DMA, FSM, ISL, KIR, KNA, LCA, MHL, MUS, NRU, NZL, TTO, WSM	\\
2011	&	7	&	BOL, DOM, FRA, IRN, ISR, MEX, PSE, SEN	\\
2011	&	8	&	BDI, COD, SSD, TJK	\\
2011	&	9	&	ARE, KEN, NAM, TON	\\
2011	&	10	&	CAN, SLV	\\
2011	&	11	&	HTI	\\
2011	&	12	&	PRK	\\
\bottomrule

\end{tabular}
\caption{Infomapped communities 2008-2011 \label{tab:communities9}}

\end{table}
%%%%%%%%%%%%%%%%%%%%%%%%%%%%%%%%%%%%%%%% 

\begin{table}
\centering
\tiny
\begin{tabular}{c c p{14cm}}
\toprule
Year & Community & ISO Country Code \\
\midrule
2012	&	0	&	AGO, ALB, AUT, BDI, BHS, BLZ, BRB, CAF, COG, COL, COM, CPV, DJI, DNK, ESP, FRA, GAB, GEO, GIN, GMB, GRD, GTM, GUY, IRL, ITA, JOR, KHM, KIR, LBR, LUX, MDA, MDV, MLI, MLT, MUS, NAM, NGA, PAN, PRT, RWA, SDN, SUR, SWE, SWZ, TCD, TUV, UGA, UKR, USA, WSM, ZAF	\\
2012	&	1	&	AZE, BIH, BRN, BWA, CHL, CMR, CYP, HTI, KNA, LAO, LCA, LSO, MAR, MCO, MOZ, MRT, NPL, PER, POL, SEN, SOM, SVK, THA, TTO, TUN, VAT, VCT	\\
2012	&	2	&	BFA, BHR, BLR, CAN, CRI, CUB, DEU, DZA, ERI, FIN, IDN, JAM, NIC, SAU, TKM, VEN	\\
2012	&	3	&	BEN, BGD, FJI, GHA, MHL, NZL, PLW, PNG, SLE, SSD, STP, SYC, TLS, TZA, VUT	\\
2012	&	4	&	ARG, ATG, BOL, ECU, IRQ, LBY, MYS, NER, PRY, SLV, URY, ZWE	\\
2012	&	5	&	ARE, CHE, LBN, OMN, QAT, RUS, SGP, VNM	\\
2012	&	6	&	BGR, CZE, GRC, KOR, LIE, LVA, MNE, ROU, SMR, YUG	\\
2012	&	7	&	AFG, BEL, BRA, COD, MDG, NLD, SVN, TGO, TUR	\\
2012	&	8	&	AUS, DMA, DOM, LKA, NOR, NRU, PSE, YEM	\\
2012	&	9	&	FSM, HND, MEX, MWI, SLB, TON, ZMB	\\
2012	&	10	&	CHN, ETH, GBR, GNQ, IND, ISL, KEN, TJK	\\
2012	&	11	&	EGY, KAZ, KWT, SYR, UZB	\\
2012	&	12	&	ARM, BTN, LTU, MMR	\\
2012	&	13	&	EST, HRV, MKD, PHL	\\
2012	&	14	&	AND, IRN, PAK, PRK	\\
2012	&	15	&	CIV, JPN, KGZ	\\
2012	&	16	&	HUN	\\
2012	&	17	&	ISR	\\
2012	&	18	&	MNG	\\
\hline
2013	&	0	&	AND, ARM, ATG, AUS, BEL, BGR, BHS, BIH, BLR, BWA, COD, COG, CRI, CUB, CYP, CZE, DEU, DMA, DOM, ESP, EST, FIN, FSM, GMB, GNQ, GRC, GTM, GUY, HND, HRV, IND, IRL, ISL, ITA, JAM, KHM, KNA, LBN, LCA, LSO, LTU, LUX, MCO, MDA, MDG, MEX, MKD, NAM, NGA, NIC, NLD, NZL, PAK, PER, POL, PRT, PSE, ROU, RUS, RWA, SDN, SMR, STP, SWE, SYR, TTO, TUR, TZA, VAT, VCT, WSM, YUG, ZAF, ZWE	\\
2013	&	1	&	CIV, COM, CPV, GNB, HTI, KIR, LBR, LKA, MLI, MLT, MMR, MNE, MNG, MOZ, NOR, NPL, NRU, SEN, SYC, TGO, UKR, ZMB	\\
2013	&	2	&	BGD, BLZ, DZA, FJI, GIN, GRD, KAZ, KWT, MDV, MUS, OMN, PNG, SLE, THA, TLS, TUV	\\
2013	&	3	&	BRA, CHE, CHL, COL, ECU, ERI, GEO, LIE, SOM, SVN, VEN	\\
2013	&	4	&	BRN, CMR, DNK, HUN, LVA, MWI, PAN, PHL, SGP, SWZ, TJK, UGA, VUT	\\
2013	&	5	&	BDI, CAF, ETH, GAB, IDN, KGZ, LAO, SSD, SVK, TKM	\\
2013	&	6	&	ARE, BHR, EGY, FRA, GBR, IRN, LBY, MRT, PRK, QAT, TUN, YEM	\\
2013	&	7	&	AFG, CHN, IRQ, JPN, MYS, PRY, SLV, VNM	\\
2013	&	8	&	AGO, AUT, AZE, BEN, BFA, KOR, MAR, NER	\\
2013	&	9	&	BRB, BTN, MHL, PLW, SLB, SUR, TON, UZB	\\
2013	&	10	&	ALB, CAN, GHA	\\
2013	&	11	&	JOR, USA	\\
2013	&	12	&	BOL	\\
2013	&	13	&	ISR	\\
2013	&	14	&	TCD	\\
2013	&	15	&	URY	\\
2013	&	16	&	ARG	\\
\hline
2014	&	0	&	AFG, AND, ARG, ATG, AUS, AZE, BEN, BGD, BIH, BRA, BRN, BTN, CHL, CIV, COL, COM, CPV, CRI, CUB, DOM, DZA, ECU, ERI, FIN, FJI, FSM, GAB, GHA, GNB, GUY, HND, HRV, HUN, IRL, ISL, JAM, KAZ, KGZ, KIR, KOR, KWT, LBN, LIE, LSO, LUX, LVA, MCO, MDV, MEX, MHL, MLI, MNE, MNG, MOZ, NAM, NER, NGA, NIC, NLD, NPL, NRU, NZL, PAN, PER, PHL, PLW, POL, PRK, PRT, PRY, RWA, SDN, SGP, SLE, SMR, SOM, SUR, SWE, SYR, THA, TJK, TON, TUN, TUV, UGA, URY, UZB, VCT, VEN, VNM, YEM, YUG, ZAF, ZMB	\\
2014	&	1	&	AGO, BFA, BHS, BRB, CAF, CMR, COD, COG, DMA, GRD, GTM, HTI, KHM, LAO, LCA, LKA, MDG, MMR, MRT, MUS, PNG, SLV, STP, SWZ, SYC, TCD, VUT, ZWE	\\
2014	&	2	&	ALB, AUT, BEL, BLR, BOL, CHN, CZE, DEU, DNK, EST, GEO, GMB, JPN, LBY, LTU, MKD, NOR, OMN, PAK, RUS, SVK, SVN, TLS, VAT	\\
2014	&	3	&	ARE, BHR, EGY, FRA, GBR, IRN, IRQ, ITA, JOR, MYS, PSE, QAT, TUR, UKR, USA	\\
2014	&	4	&	BDI, BWA, CAN, ETH, MAR, MWI, SEN, TGO, TTO, TZA, WSM	\\
2014	&	5	&	ESP, IDN, MDA, MLT	\\
2014	&	6	&	BGR, CHE, CYP, GRC, IND, ROU	\\
2014	&	7	&	ARM, TKM	\\
2014	&	8	&	BLZ, GNQ, KEN, SSD	\\
2014	&	9	&	GIN, KNA, SLB	\\
2014	&	10	&	ISR	\\
2014	&	11	&	LBR	\\
\bottomrule

\end{tabular}
\caption{Infomapped communities 2012-2014 \label{tab:communities10}}

\end{table}
%%%%%%%%%%%%%%%%%%%%%%%%%%%%%%%%%%%%%%%% 

\end{document}